\documentclass[12pt]{article}
\usepackage{chicagob}
\usepackage{times}

\input{amssym}
\ifx\pdfoutput\undefined
  \usepackage[dvips]{color}
  \usepackage{graphicx}
  \DeclareGraphicsExtensions{.eps}
\else
 \usepackage[pdftex]{color}
 \usepackage[pdftex]{graphicx}
 \DeclareGraphicsExtensions{.pdf}
\fi

\newcommand{\newcomment}{}

\renewcommand{\S}{{\cal S}}
\newcommand{\ML}{\mbox{{\it ML}}}
\newcommand{\FS}{\mbox{{\it FS}}}
\newcommand{\SE}{\mbox{{\it SE}}}
\newcommand{\Mer}{\mbox{{\it Mer}}}
\newcommand{\Mor}{\mbox{{\it Mor}}}
\newcommand{\MerE}{\mbox{{\it MerE}}}
\newcommand{\MorE}{\mbox{{\it MorE}}}
\newcommand{\FF}{\mbox{{\it FF}}}
\newcommand{\LL}{\mbox{{\it LL}}}
\newcommand{\LC}{\mbox{{\it LC}}}
\newcommand{\WB}{\mbox{{\it WB}}}
\newcommand{\BMC}{\mbox{{\it BMC}}}
\newcommand{\TT}{\mbox{{\it TT}}}
\newcommand{\TD}{\mbox{{\it TD}}}
\newcommand{\ST}{\mbox{{\it ST}}}
\newcommand{\SST}{\mbox{{\it SST}}}

\newcommand{\SPS}{\mbox{{\it SPS}}}

\newcommand{\LSS}{\mbox{{\it LSS}}}
\newcommand{\BPT}{\mbox{{\it BPT}}}
\newcommand{\HPT}{\mbox{{\it HPT}}}
\newcommand{\MT}{\mbox{{\it MT}}}
\newcommand{\LE}{\mbox{{\it LE}}}
\newcommand{\FB}{\mbox{{\it FB}}}
\newcommand{\AS}{\mbox{{\it AS}}}
\newcommand{\ES}{\mbox{{\it ES}}}
\newcommand{\BH}{\mbox{{\it BH}}}
\newcommand{\BT}{\mbox{{\it BT}}}
\newcommand{\BS}{\mbox{{\it BS}}}
\newcommand{\SH}{\mbox{{\it SH}}}
\newcommand{\LT}{\mbox{{\it LT}}}
\newcommand{\RT}{\mbox{{\it RT}}}
\newcommand{\arc}{-\!\!\!-\!\!\!\!\blacktriangleright}
\newtheorem{THEOREM}{Theorem}[section]
\newenvironment{theorem}{\begin{THEOREM} \hspace{-.85em} {\bf :} }%
                        {\end{THEOREM}}
\newtheorem{LEMMA}[THEOREM]{Lemma}
\newenvironment{lemma}{\begin{LEMMA} \hspace{-.85em} {\bf :} }%
                      {\end{LEMMA}}
\newtheorem{COROLLARY}[THEOREM]{Corollary}
\newenvironment{corollary}{\begin{COROLLARY} \hspace{-.85em} {\bf :} }%
                          {\end{COROLLARY}}
\newtheorem{PROPOSITION}[THEOREM]{Proposition}
\newenvironment{proposition}{\begin{PROPOSITION} \hspace{-.85em} {\bf :} }%
                            {\end{PROPOSITION}}
\newtheorem{DEFINITION}[THEOREM]{Definition}
\newenvironment{definition}{\begin{DEFINITION} \hspace{-.85em} {\bf :} \rm}%
                            {\end{DEFINITION}}
\newtheorem{CLAIM}[THEOREM]{Claim}
\newenvironment{claim}{\begin{CLAIM} \hspace{-.85em} {\bf :} \rm}%
                            {\end{CLAIM}}
\newtheorem{EXAMPLE}[THEOREM]{Example}
\newenvironment{example}{\begin{EXAMPLE} \hspace{-.85em} {\bf :} \rm}%
                            {\end{EXAMPLE}}
\newtheorem{REMARK}[THEOREM]{Remark}
\newenvironment{remark}{\begin{REMARK} \hspace{-.85em} {\bf :} \rm}%
                            {\end{REMARK}}
\newcommand{\thm}{\begin{theorem}}
\newcommand{\lem}{\begin{lemma}}
\newcommand{\pro}{\begin{proposition}}
\newcommand{\dfn}{\begin{definition}}
\newcommand{\rem}{\begin{remark}}
\newcommand{\xam}{\begin{example}}
\newcommand{\cor}{\begin{corollary}}
\newcommand{\prf}{\noindent{\bf Proof:} }
\newcommand{\ethm}{\end{theorem}}
\newcommand{\elem}{\end{lemma}}
\newcommand{\epro}{\end{proposition}}
\newcommand{\edfn}{\bbox\end{definition}}
\newcommand{\erem}{\bbox\end{remark}}
\newcommand{\exam}{\bbox\end{example}}
\newcommand{\ecor}{\end{corollary}}
\newcommand{\eprf}{\bbox\vspace{0.1in}}
\newcommand{\beqn}{\begin{equation}}
\newcommand{\eeqn}{\end{equation}}
\newcommand{\bbox}{\vrule height7pt width4pt depth1pt}

\newcommand{\clm}{\begin{claim}}
\newcommand{\eclm}{\end{claim}}
\newcommand{\sat}{\models}

\newcommand{\dimp}{\Leftrightarrow}
\newcommand{\union}{\cup}
\newcommand{\inter}{\cap}
\newcommand{\IN}{\mbox{$I\!\!N$}}

\renewcommand{\phi}{\varphi}
\newcommand{\E}{{\cal E}}
\newcommand{\F}{{\cal F}}

\newcommand{\R}{{\cal R}}

\newcommand{\U}{{\cal U}}
\newcommand{\V}{{\cal V}}

\newcommand{\<}{\langle}
\renewcommand{\>}{\rangle}

\newcommand{\respc}{resp.,\ }

\newcommand{\ol}{\setlength{\itemsep}{0pt}\begin{enumerate}}
\newcommand{\eol}{\end{enumerate}\setlength{\itemsep}{-\parsep}}
\newcommand{\ul}{\setlength{\itemsep}{0pt}\begin{itemize}}
\newcommand{\dl}{\setlength{\itemsep}{0pt}\begin{description}}
\newcommand{\edl}{\end{description}\setlength{\itemsep}{-\parsep}}
\newcommand{\eul}{\end{itemize}\setlength{\itemsep}{-\parsep}}

\newcommand{\commentout}[1]{}

\newcommand{\bi}{\begin{itemize}}
\newcommand{\ei}{\end{itemize}}
\newcommand{\be}{\begin{enumerate}}
\newcommand{\ee}{\end{enumerate}}

\begin{document}

%joe3: cut for UAI
\begin{titlepage}
%joe6*: I'd like to promote explanation more, since I now feel rather
%excited about it.
%\title{Actual Causality}>       

%jp7: me too, but the  title you suggested sounds like
%we are merely translating an existing definition to the language
%of structural equations. How about the following
%\title{Defining Actual Causality and Explanation Using Structural
%Equations}
%\title{Actual Causation and Causal Explanation}
%jp7: or even "Causes  and Explanations" ? simple  and devastating .
%joe7: I like the last version: how about
\title{Causes and Explanations: A
%joe8: what about the following?
%New
%jp9 fancier:
%joe9: OK
% Structural-Equations
Structural-Model
Approach.
%joe19
Part I: Causes}
\author{Joseph Y. Halpern%
\thanks{Supported in part by NSF under
grant IRI-96-25901}\\
Cornell University\\
Dept. of Computer Science\\
Ithaca, NY 14853\\
halpern@cs.cornell.edu\\
http://www.cs.cornell.edu/home/halpern\\
\and
Judea Pearl%
\thanks{Supported in part by grants from NSF,
ONR, AFOSR, and MICRO.}\\
Dept. of Computer Science\\
University of California, Los Angeles\\
Los Angeles, CA 90095\\
judea@cs.ucla.edu\\
http://www.cs.ucla.edu/$\sim$judea
}

%jp8: adding
\date{\today}
%joe: cut for UAI
\setcounter{page}{0}
\thispagestyle{empty}
\maketitle
\thispagestyle{empty}

\begin{abstract}
%joe25: removed explanation from abstract
%We propose new definitions of {\em actual cause}
%and {\em (causal) explanation}, using
%{\em structural equations\/} to model counterfactuals.
We propose a new definition of {\em actual causes},
using {\em structural equations\/} to model counterfactuals.
We show that the definition yields a plausible and elegant account of
causation that
handles well examples which have caused problems for other definitions
and resolves major difficulties in the traditional account.
\end{abstract}
%joe3: cut for UAI
\end{titlepage}

\section{Introduction}
What does it mean that an event $C$ {\em actually caused\/} event $E$?
%joe25: moved this up from below
The problem of defining ``actual cause'' goes beyond mere
philosophical speculation.
As Good \citeyear{good:93} and Michie \citeyear{michie:97}
argue persuasively,
in many legal settings, what
needs to be
established (for determining responsibility)
%joe25: this is a counterfactual kind of causation, to some extent!
%is not a counterfactual kind of causation, but
is exactly such ``cause in fact''. A typical example
\cite{wright:88} considers two fires advancing toward a house. If
fire $A$ burned the house before fire $B$,
we (and many juries nationwide) would consider fire $A$ ``the
%joe+
%actual cause'' for the damage, even supposing the house would have
%definitely burned down by fire $B$, if it were not for $A$.
actual cause'' for the damage, even supposing that the house would 
definitely have been burned down by fire $B$, if it were not for $A$.
Actual causation is also important in artificial intelligence applications.
Whenever we undertake to {\em explain\/} a set of events that
unfold in a specific scenario, the explanation produced must
acknowledge the actual cause of those events.
The automatic generation of adequate explanations,
a task essential in planning, diagnosis, and natural language processing,
therefore requires
a formal analysis of the concept of actual cause.

The philosophy literature has been struggling with this problem
%joe25
of defining causality
%joe**
%since the days of Hume~\citeyear{Hume39} who was the first
since at least the days of Hume~\citeyear{Hume39}, who was the first
to identify causation with counterfactual dependence.
%joe24: Is Hume, 1748 different from Hume, 1739.  If so, could you send
%me the bib details?
To quote Hume \citeyear[Section {VIII}]{hume:1748}:
\begin{quote}
We may define a cause to  be an object followed
by another, ..., where, if the first object had not been, the second
never had existed.
\end{quote}
Among modern philosophers, the counterfactual interpretation
of causality continues to receive most attention,
%joe25
%primarily due to the works of David Lewis (\citeyear{Lewis86a}).
%joe**: used earlier citation
%primarily due to the work of David Lewis \citeyear{Lewis86a}.
primarily due to the work of David Lewis \citeyear{Lewis73a}.
Lewis has given counterfactual dependence
formal  underpinning in possible-world semantics and
has equated actual causation with
the transitive closure of counterfactual dependencies.
$C$ is classified as a cause of $E$ if $C$ is linked to $E$ by a
chain of events each directly depending on its predecessor.
%joe+: added period at end
However, Lewis's dependence theory has encountered many difficulties.
%joe**: changed ref
%(See \cite{sosa:too93}, \cite{Hall98}, and \cite{pearl:2k}
(See \cite{Collins03,HallP03,pearl:2k,sosa:too93}
%joe25
%for some recent discussions.)
for some recent discussion.)
The problem is that effects may not always counterfactually
depend on their causes, either directly or indirectly,
%joe26:  jp0 change
as the two-fire example illustrates.
In addition, causation is not always transitive, as implied
Lewis's chain-dependence account (see Example~\ref{xam4}).

%joe25: cut this; it has nothing to do with the problems in Lewis's
%account
%To borrow just one example from Hall \citeyear{Hall98}, suppose a
%bolt lightning hits a tree and starts a forest fire.  It seems
%reasonable to say that the lightning bolt is a cause of the fire.
%(Indeed, the description ``the lightning bolt \ldots starts a forest
%fire'' can be viewed as saying this.)  But what about the oxygen in the
%air and the fact that the wood was dry?  Presumably, if there has not
%been oxygen or the wood was wet there would not have been a fire.
%Carrying this perhaps to the point of absurdity, what about the Big
%Bang?
%This problem is relatively easy to deal with, but there are a host of
%other, far more subtle, difficulties that have been raised over the
%years.

Here we give a definition of actual causality cast in the language of
{\em structural equations\/}.
%joe25: we're not really ``replacing'' counterfactual dependency; we're
%refning it.  (As we say in the conclusion, we definitely use
%counterfactuals in our definition.)
The basic idea is to
%joe25
%replace counterfactual dependency with ``contingent dependency''.
extend the basic notion of counterfactual dependency to allow ``contingent
dependency''.
In other words, while effects
may not always counterfactually depend on their causes
%they always depend on them under certain contingencies that must
%be carefully defined
%joe25
in the actual situation,
they do depend on them under certain contingencies.
In the case of the two fires, for example,
%the house would not have burned down if it were not
%for fire $A$ and if, countrary to fact, fire fighters would have reached
%the house any time between the actual arrival of fire $A$ and that
%of fire $B$.
the house burning down does depend on fire $A$ under the contingency
%joe+
%that fire fighters reach
that firefighters reach
the house any time between the actual arrival of fire $A$ and that
of fire $B$.  Under that contingency, if fire $A$ had not been started,
the house would not have burned down.  The house burning down also depends
on fire $A$ under the contingency that fire $B$ was not started.
But this leads to an obvious concern: the house burning down also
depends on fire $B$ under the contingency that fire $A$ was not
started.
%joe2
%Part I of our paper gives a precise definition of this type of
We do not want to consider this latter contingency.
%(such as the arrival of fire fighters).
Roughly speaking,
%joe25
%it is a contingency  which does not itself
we want to allow only contingencies
%joe**
%which do not interfere with active causal process.
that do not interfere with active causal processes.
%joe25
%in the scenario
%but which, nevertheless, render the effect dependent on
%the earlier fire and not the later fire.
Our formal definition of actual causality tries to make this precise.

%joe**: added ref
%joe+
%In Part II of the paper \cite{HP01b},
In Part II of the paper \cite{HP01a},
we  give a definition of {\em (causal) explanation} using the
definition of causality.
An explanation adds information to an agent's knowledge;
very roughly, an explanation of $\phi$ is a minimal elaboration
of events that suffice to  cause $\phi$ even in the face of
uncertainty about the actual situation.

The use of structural
equations as a model for causal relationships
is standard in the social sciences,
and seems to go back to the work of Sewall Wright in the 1920s (see
\cite{Goldberger72} for a discussion);
the particular framework that we use here
is due to Pearl \citeyear{Pearl.Biometrika}, and is further developed in
\cite{GallesPearl97,Hal20,pearl:2k}.
While it is hard to argue
that our definition (or any other definition, for that matter) is the
``right'' definition,
we show that it deals well with the difficulties
that have plagued other approaches in the past,
especially those exemplified by the rather extensive
%joe29
%compendium of Hall \citeyear{Hall98}
%joe**
%compendia of Hall \citeyear{Hall98} and Hall and Paul
%\citeyear{HallP03}
compendium of Hall and Paul \citeyear{HallP03}.
%joe**
%and Lewis's recent paper \citeyear{Lewis00}.

%joe25
%Formally,
According to our definition,
the truth of every claim must be
evaluated relative to a particular model of the world;
%joe25
%and, naturally, our definition will only allow us to claim
that is, our definition allows us to claim only
that $C$ causes $E$ in a
(particular context in a)
particular structural model.
It is possible to construct two closely related
structural models such that $C$ causes $E$ in one and 
%joe**
%some other event $C'$ causes $E$ in another.  Among other things, the
%joe*
%and 
$C$ does not cause $E$ in the other.  Among other things, the
modeler must decide 
which variables (events) to reason about and which to leave in the
background.
%joe25: this example doesn't really illustrate the point
%The lightning and the forest fire example already shows the
%impact of such decisions: if we include ``oxygen'' in the model, then it
%becomes a cause; if we leave it in the background, it does not.
We view this as a feature of our model, not a bug.  It moves the question
of actual causality to the right arena---debating which of
two (or more) models of the world is a better representation
%joe26: jp0 change
of those aspects of the world that one wishes to capture and
reason about.
This, indeed, is the type of debate that goes on in informal (and
legal) arguments all the time.

%joe25
%There has been extensive discussion about causality in the
%literature, particularly in the philosophy literature.
There has been extensive discussion about causality in the
philosophy literature.
To keep this paper to manageable length,
we spend only minimal time describing other approaches and comparing
%joe+
%ours to them.
ours with them.
We refer the reader to
%joe**
%\cite{Hall98,pearl:2k,sosa:too93,SpirtesSG} for details and criticism of
\cite{HallP03,pearl:2k,sosa:too93,SpirtesSG} for details and criticism of
the probabilistic and logical approaches to causality
in the philosophy literature.
(We do try to point out where our definition does better than perhaps
%joe25
%the best known approach, due to Lewis \citeyear{Lewis86a,Lewis00}, in
%joe**
%the best-known approach, due to Lewis \citeyear{Lewis86a,Lewis00}, as
the best-known approach, due to Lewis \citeyear{Lewis73a,Lewis00}, as
%joe**: changed to Yablo02 here and one other time
%well as some other recent approaches \cite{Hall98,paul:98,Yablo00}, in
well as some other recent approaches
\cite{Hall01,paul:98,Yablo02}, in 
the course of discussing the examples.)

There has also been work in the AI literature on causality.
Perhaps the closest to this are papers by Pearl and his
colleagues that use the structural-model approach.  The definition of
causality in this paper was inspired by an
earlier paper of Pearl's \citeyear{pearl:98c} that defined actual
causality in terms of a construction called a {\em causal beam}.
The definition was later modified somewhat (see \cite[Chapter
10]{pearl:2k}). The modifications were in fact largely due to the
considerations addressed in this paper.
The definition given here
is more transparent and handles a number of cases better (see
Example~\ref{voting} in the appendix).

Tian and Pearl \citeyear{TP00} give results on estimating
(from empirical data) the
probability that $C$ is a {\em necessary\/} cause of $E$---that is, the
probability that $E$ would not have occurred if $C$ had not occurred.
Necessary causality is related to but different from actual causality,
as the definitions should make clear.
Other work (for example, \cite{HeckShac}) focuses on when a
random variable $X$ is the cause of a random variable $Y$; by way of
contrast, we focus on when an {\em event\/} such as $X=x$ causes an
event such as $Y=y$.
Considering when a random variable is the cause of another is perhaps
more appropriate as a {\em prospective\/} notion of causality: could
$X$ potentially be a cause of changes in $Y$.  Our notion is
more appropriate for a
{\em retrospective\/} notion of causality: given all the
information relevant to a given scenario, was $X=x$ the actual
cause of $Y=y$ in that scenario?
Many of the subtleties that arise when dealing with
events simply disappear if we look at causality at the level of
random variables.
Finally, there is also a great deal of work in AI on
formal action theory (see, for example, \cite{lin:95,sandewall:94,reiter:01}),
%add reference to Reiter's new book "Knowledge in Action"
%MIT Press, 2001.
which is concerned with
the proper way of incorporating causal relationships
into a knowledge base so as to guide actions.
The focus of our work is quite different; we are concerned
with extracting the actual causality relation
from such a knowledge base, coupled with a specific scenario.

\commentout{
The literature on explanation is almost as vast as the literature on
causality; see
\cite{CH97,Gardenfors1,Hempel65,Pearl,Salmon89} for some background on
explanation.   Again, we do not pretend to do a serious comparison of our
approach to the many others; in Section~\ref{expl:compare}, there is
some comparison of our work to that of G\"{a}rdenfors
\citeyear{Gardenfors1} and the MAP (maximum a posteriori) approach
commonly used in the AI literature \cite{Pearl}.
}

%joe+
%The best judge of the
The best ways to judge  the
adequacy of an approach are the intuitive appeal of the definitions and
how well it deals with examples; we
believe that this paper shows that our
approach fares well on both counts.

The remainder of the paper is organized as follows.  In the next
section, we review structural models.
In Section~\ref{sec:actcaus}
we give
a preliminary
definition of actual causality
and show
in Section~\ref{sec:examples}
how it deals with
some examples of causality that have been problematic for
other accounts.
We refine the definition slightly in Section~\ref{sec:refined}, and show
how the refinement handles further examples.
We conclude in Section~\ref{sec:discussion} with some discussion.

%\section{The definitions}\label{sec:definitions}
%joe2
%\section{Causal Models}\label{sec:models}
\section{Causal Models: A Review}\label{sec:models}

In this section we review the basic definitions of causal models,
as defined in terms of structural equations, and
%and show how they can be used
%to define a reasonable notion of actual causality.
the syntax and semantics of a language for reasoning about
causality.
%joe**
\newcomment
%joe+
%We also briefly compare our approach to the more standard approaches to
We also briefly compare our approach with the more standard approaches to
modeling causality used in the literature.

%joe4
\paragraph{Causal Models:}
The description of causal models given here is taken
from \cite{Hal20}; the reader is referred to
%joe19
%\cite{GallesPearl97,galles:pea98,Hal20} for more details, motivation,
%joe+
%See 
\cite{GallesPearl97,Hal20,pearl:2k} for more details, motivation,
and intuition.

The basic picture here is that we are interested in the values of random
variables.  If $X$ is a
random variable, a typical event has the form $X=x$.  (In terms of
possible worlds, this just represents the set of possible worlds where
$X$ takes on value $x$, although the model does not describe the set of
%%possible worlds.)  Some random variables may have a causal effect
%%on others. This effect
%%is modeled by a set of {\em structural equations}.
%%judea 1 revise sentence after "possible worlds." 9.13
possible worlds.)  Some random variables may have a causal
influence on others. This influence
is modeled by a set of {\em structural equations}.
%joe20: note also period at end of last line
Each equation represents a distinct mechanism (or law) in the
world, one that may be modified (by external actions) without
altering the others.
%continue with: "In practice, ..."
In practice, it
seems useful to split the random variables into two sets, the {\em
exogenous\/} variables, whose values are determined by factors outside
the model, and the {\em endogenous\/} variables,
%joe20
whose values are ultimately determined by the exogenous variables.
It is these endogenous
variables whose values are described by the structural equations.

%joe25:
%More formally, a {\em signature\/} $\S$ is a tuple $(\U,\V,\R\}$,
Formally, a {\em signature\/} $\S$ is a tuple $(\U,\V,\R)$,
where $\U$ is a set of exogenous variables, $\V$ is a set of endogenous
variables,
%joe1: done; changed old \X to \V and old \V to \R
%[[Joe, how about using $V$, like the rest of the world?]],
and $\R$ associates with every variable $Y \in
\U \union \V$ a nonempty set $\R(Y)$ of possible values for $Y$
%joe1
(that is, the set of values over which $Y$ {\em ranges}).
%joe16
In most of this paper (except the appendix) we assume that $\V$ is
finite.
A {\em causal model\/}
%joe9: to connect to title
(or {\em structural model\/})
over signature $\S$ is a tuple $M=(\S,\F)$,
%joe1: done (I hope I caught all occurrences)
%[[Joe, how about replacing $T$ with $M$?]]
where $\F$ associates with each variable $X \in \V$ a function denoted
$F_X$ such that $F_X: (\times_{U \in \U} \R(U))
\times (\times_{Y \in \V - \{X\}} \R(Y)) \rightarrow \R(X)$.
%joe26; inserted insert-1 here, with minor changes marked below
$F_X$ determines the value of $X$
given the values of all the other variables in $\U \union \V$.
%joe28: added another variable to make the point you wanted to make
%For example, if $F_X(Y,U) = Y+U$ (which we usually write
For example, if $F_X(Y,Z,U) = Y+U$ (which we usually write
%joe27
as
$X=Y+U$), then
if $Y = 3$ and $U = 2$, then
%joe26: next line is unnecessary (I think it reads better without it)
%$F_X$ determines the value
$X=5$,
%joe28: added
regardless of how $Z$ is set.
%joe26: there's a problem here.  If there were other variables in the
%system, then these should have been arguments of F_X, according to our
%definition.  (We said we're not leaving out any variables.  Now we
%could just make them part of the input to F_X, leaving the Y+U
%unchanges.  That would certainly emphasize the point that the result
%does not depend on their values.  But I think this would just clutter
%things up.  I cut the next two lines.
%regardless of the values that other variables in the system (i.e., variables
%in $(\V \union \U) - \{X,Y,U\}$) might take.
%joe26: Why is it just a physical process?
%Additionally, these equations represent
These equations can be thought of as representing
%joe28
%physical 
processes (or mechanisms) by which values are assigned to
%joe27
%variables, hence, like physical laws, they support
variables.  Hence, like physical laws, they support a
counterfactual interpretation.  For example, the equation above
claims that, in the context $U = u$,
if $Y$ were $4$, then $X$ would
be $u+4$ (which we write as $(M,u) \sat
%joe+
%[Y \gets 4](X=u+4)$), regardless of what value 
[Y \gets 4](X=u+4)$), regardless of what values 
%joe28
%Y and X
$X$, $Y$, and $Z$
actually take in the real world.

%joe28: is this what you intended by the jp00 below?  The change is fine
%with me (although I think it was OK before too).
%The functions $F_X$ define a set of {\em (modifiable)
The function $\F$ defines a set of {\em (modifiable)
%jp00
%should'nt it be where $\F$, instead of  $F_X$ ??
structural equations}, relating the values of the variables.  Because
$F_X$ is a function, there is a unique value of $X$ once we have set all
the other variables.
Notice that we have such functions only for the endogenous variables.
The exogenous variables
are taken as given; it is their effect on the endogenous
variables (and the effect of the endogenous variables on each other)
that we are modeling with the structural equations.

%joe8: added paragraph below; I think we need to slow down here
%joe10
%An equation such as $X = F_X(\vec{u},y)$ should be that of as saying
%joe26: jp0 insert 2 (with minor changes, as marked)
The counterfactual interpretation and the causal
asymmetry associated with the structural equations are best
seen when we consider external interventions (or spontaneous
changes), under which some equations in $F$ are modified.
An equation such as
%joe26: I think this should be x, not X, right?
%$X = F_X(\vec{u},y)$ should be thought of as saying
$x = F_X(\vec{u},y)$ should be thought of as saying
that in a context where the exogenous variables have values $\vec{u}$,
if $Y$ were set to $y$
%joe26
%by some external intervention (not part of the model),
by some means (not specified in the model),
then $X$ would take on the value
$x$, as dictated by $F_X$. The same does not hold when we intervene
directly on $X$;
such an intervention amounts to assigning a value to $X$ by
external means, thus overruling the assignment specified
%joe26: minor English changes
%by $F_X$ and  hence, $Y$ is no longer committed to track $X$
by $F_X$. In this case, $Y$ is no longer committed to tracking $X$
according to $F_X$.
%joe26: more minor changes
%We see that variables on the left hand sides of the equations
%are treated differently that those in the right hand side.
Variables on the left-hand side of equations
are treated differently from ones on the right-hand side.

For those more comfortable with thinking of counterfactuals in terms of
possible worlds, this modification of equations
may be given a simple ``closest world'' interpretation:
%joe27: added next line
the solution of the equations obtained by
%jp00 Joe, I dont understand why you object to my emotional
%need to emphasize the invariance of the other equations .
%the replacement operator is simple, intuitive
%and matches intuitions of "minimal change" or "closest
%world". I thought you agreed to this.??
%joe28: OK; I'm comfortable with the current version
%equation for Y with the equation Y=y means formally writing the
%equation F_Y(y) = y.  However, F_Y takes as an argument every variable
%other than Y, so we can't have F_Y(y) = y.  I tried to rewrite it to
%make it technically correct while keeping your intuition.  
%setting $Y$ to $y$ by
%%joe26: this is not what we're doing
replacing the equation for $Y$ with the equation $Y = y$,
while leaving all other equations unaltered,
%by removing the equation for $Y$ and setting $Y$ to $y$ in all other
%equations
%$Y = y$, which represents the minimal change in mechanisms
%required for ensuring that the solution satisfies
%the specified condition %$Y = y$ for every $u$.
%joe27
%introduces a minimal change in the model that
%that guarantees that
%$Y=y$.
gives the closest ``world'' to the actual world where $Y=y$.
%joe27: now unnecessary
%Therefore, the solution of the modified
%%joe26
%%equations can be thought of as a
%%``possible world'' that is closest to ours and
%%still satisfies $Y=y$.
%system of equations can be viewed as the
%``possible world'' that satisfies $Y=y$ that is closest to the actual
%world.
In this possible-world interpretation, the asymmetry
embodied in the model says that if $X=x$ in the closest world to $w$
where $Y=y$, it does not follow that $Y=y$ in the closest worlds to $w$
where $X=x$.  In terms of structural equations, this just says that
if $X=x$ is the solution for $X$ under the intervention $Y=y$,
%joe+
%it does not follow that $Y=y$ in solution for Y under
it does not follow that $Y=y$ is the solution for $Y$ under
the intervention $X=x$. Each of two interventions modifies
the system of equations in a distinct way; the former
modifies the equation in which $Y$ stands on the left,
while the latter modifies the equation in which $X$ stands
on the left.

In summary, the equal sign in a structural equation differs
from algebraic equality; in addition to describing
%joe+
%an equality relationship between variables, it also acts as an
an equality relationship between variables, it acts as an
assignment statement in programming languages, since it
%joe26: I'm not sure that ``strategy'' is the right word here
%specifies the strategy by which variables values are determined.
%joe27
%specifies the means by which variables' values are determined.
specifies the way variables' values are determined.
This should become clearer in our examples.

%joe10: added example
\xam\label{xam:forestfire} Suppose that we want  to reason about a
forest fire that could be
caused by either lightning or a match lit by an arsonist.  Then the causal
model would have the following endogenous variables (and perhaps others):
\begin{itemize}
\item $F$ for fire ($F=1$ if there is one, $F=0$ otherwise);
\item $L$ for lightning ($L=1$ if lightning occurred, $L=0$ otherwise);
%joe20: check
\item $\ML$ for match lit ($\ML=1$ if the match was lit, $\ML = 0$ otherwise).
%\item \ldots
\end{itemize}
%joe13: corrected typo
%The set $\U$ of exogenous variables $\U$ would include things 
The set $\U$ of exogenous variables includes
%joe20
%things we need to assume so as
conditions that suffice
%jp11, clarifying next sentence
%want to take for granted (such as whether the wood is dry, there is
to render all relationships deterministic
%joe+
%(such as whether the wood is dry, there is
(such as whether the wood is dry, whether there is
enough oxygen in the air
%joe25
for the match to light,
etc.).
%If we want to reason explicitly about the dryness of the
%wood, it is more appropriate to make it an endogenous variable.
Suppose that $\vec{u}$ is a setting of the exogenous variables that
makes a forest fire possible (i.e., the wood is sufficiently dry, there
is oxygen in the air, and so on).  Then, for example,
$F_{F}(\vec{u},L,\ML)$ is such that $F = 1$ if either $L=1$ or $\ML=1$.
%joe25: added
Note that although the value of $F$ depends on the values of $L$ and
$\ML$, the value of $L$ does not depend on the values of
$F$ and $\ML$.
\exam

%joe25: added next line
%joe27
As we said,
a causal model has the resources to determine counterfactual effects.
Given a causal model $M = (\S,\F)$, a (possibly
empty)  vector
$\vec{X}$ of variables in $\V$, and vectors $\vec{x}$ and
$\vec{u}$ of values for the variables in
$\vec{X}$ and $\U$, respectively, we can define a new causal model
denoted
%$M_{\vec{X} \gets \vec{x}}(\vec{u})$ over the signature $\S_{\vec{X}}
$M_{\vec{X} \gets \vec{x}}$ over the signature $\S_{\vec{X}}
= (\U, \V - \vec{X}, \R|_{\V - \vec{X}})$.%
\footnote{We are implicitly identifying the vector $\vec{X}$ with the
subset of $\V$ consisting of the variables in $\vec{X}$.
$\R|_{\V - \vec{X}}$ is the restriction of $\R$
to the variables in $\V - \vec{X}$.}
%joe1: for what it's worth, I don't particularly like the word submodel.
%It's not a submodel in any obvious sense.  I can't think of a better
%name offhand though.
%$M_{\vec{X} \leftarrow \vec{x}}$ is named %``submodel''
$M_{\vec{X} \leftarrow \vec{x}}$ is called a {\em submodel\/} of $M$
by Pearl \citeyear{pearl:2k}.
%% judea1 rewrite sentence starting: "Intuitively". 9.13
Intuitively, this is the causal model that results when the variables in
$\vec{X}$ are set to $\vec{x}$ by
%joe18
%external action, the cause of which is not modeled explicitly.
%joe19
%an external action whose cause is not modeled explicitly.
by some external action that affects only the variables in $\vec{X}$;
we do not model the action or its causes explicitly.
%%Intuitively, this is the causal model that results when the variables in
%%$\vec{X}$ are set to $\vec{x}$.
%and the variables in $\U$ are set to $\vec{u}$.
%Formally, $M_{\vec{X} \gets \vec{x}}(\vec{u} = (\S_{\vec{X}},
%\F^{\vec{X} \gets \vec{x},\vec{u}}\})$,
%where $F_Y^{\vec{X} \gets \vec{x},\vec{u}}$ is obtained from $F_Y$
Formally, $M_{\vec{X} \gets \vec{x}} = (\S_{\vec{X}},
\F^{\vec{X} \gets \vec{x}})$,
where $F_Y^{\vec{X} \gets \vec{x}}$ is obtained from $F_Y$
by setting the values of the
variables in $\vec{X}$ to $\vec{x}$.
%joe25
For example, if $M$ is the structural model describing
Example~\ref{xam:forestfire}, then the model $M_{L \gets 0}$
has the equation $F = \ML$.  The equation for $F$ in $M_{L \gets 0}$ no
longer involves $L$; rather, it is determined by setting $L$ to 0 in the
equation for  $F$ in $M$.
%joe28
Moreover, there is no equation for $L$ in $M_{L \gets 0}$.

%joe10: added (this was a major point of discussion with the
%philosophers)
It may seem strange that we are trying to understand causality using
causal models, which clearly already encode causal relationships.
%jp11 deleting
%As we shall see,
%joe12:
%Our reasoning is in fact not circular.
Our reasoning is not circular.
%jp11, adding elaboration
Our  aim is not to reduce causation to noncausal concepts,
but to interpret questions about causes of specific events
in fully specified scenarios
in terms of generic causal knowledge such as
%joe12
%the one we
what we
obtain from the equations of physics.
The causal models encode background knowledge
%jp11 making  more specific
%that we take for granted (such as the fact
about the tendency of certain event types to cause  other
event types (such as the fact that lightning can cause forest fires).
%jp11,  questioning if the next line is needed.
%joe12:
%Whereas our targets are
We use the models to determine the causes
%joe19
%and explanations
of single (or token) events, such as whether it was arson
that caused the fire of June 10, 2000, given what is known or
assumed about that particular fire.
%joe12: cut
%This too will become clearer once we examine a few examples.

%joe2: cut for abstract
Notice that, in general, there may not be a unique vector of values that
simultaneously satisfies the equations in $M_{\vec{X} \gets
%\vec{x}}(\vec{u})$; indeed, there may not be a solution at all.  For
\vec{x}}$; indeed, there may not be a solution at all.  For
simplicity in this paper, we restrict attention to what are called {\em
recursive\/} (or {\em acyclic\/}) equations.  This is the special case
where there is some total ordering $\prec$ of the variables in $\V$
such
that if $X \prec Y$, then $F_X$ is independent of the value of $Y$;
%joe+
%i.e.,
that is,
$F_X(\ldots, y, \ldots) = F_X(\ldots, y', \ldots)$ for all $y, y' \in
\R(Y)$.  Intuitively, if a theory is recursive, there is no
%joe+
%feedback.  If $X \prec Y$, then the value of $X$ may affect the value of
feedback.  If $X \prec Y$, then the value of $X$ may effect the value of
$Y$, but the value of $Y$ has no effect on the value of $X$.
%joe2: cut for abstract
%joe6: also thought it wasn't worth it for the full paper.
%(This is made precise in Lemma~\ref{indep} below.)
%joe8: added
We do not lose much generality by restricting to recursive models (that
is, ones whose equations are recursive).  As suggested in the latter
half of Example~\ref{xam2}, it is always
possible to timestamp events to impose an ordering on variables and thus
construct a recursive model corresponding to a story.  In any case, in
the appendix, we sketch the necessary modifications of our definitions
to deal with nonrecursive models.

It should be clear that if $M$ is a recursive causal model,
then there is always a
unique solution to the equations in
$M_{\vec{X} \gets \vec{x}}$, given a setting $\vec{u}$ for the
variables in $\U$ (we call such a setting $\vec{u}$ a {\em context}).
We simply solve for the variables in the order given by $\prec$.

%joe2
%Given a causal model $M = (\S,\F)$ over signature $\S$,
%we can construct a corresponding {\em
%causal network}.
We can describe (some salient features of) a causal model $M$ using a
{\em causal network}.  This is a graph
%with nodes corresponding to the random variables in $\U \union \V$
with nodes corresponding to the random variables in $\V$ and an edge
from a node labeled $X$ to one labeled $Y$ if $F_Y$ depends on the value
of $X$.
%joe4: added
%In this paper, we restrict to causal models whose
%causal networks are dags; such models are called {\em recursive}.
%It is easy to check that, in recursive models, all equations have
%unique solutions.
%joe2: cut for abstract
This graph is a {\em dag\/}---a directed, acyclic graph
%joe27: added for philosophers
%(that is, a graph with no cycle of edges).
%jp00 suggesting
(that is, a graph with no cycle of directed edges).
The
acyclicity follows from the assumption that the equations are recursive.
%Note that the exogenous variables must always be roots of the
%dag---there are no edges coming into nodes labeled by random variables
%in $\U$.  [[EXAMPLE??]]
Intuitively, variables can have a causal effect only on their
descendants in the causal network; if $Y$ is not a descendant of $X$,
then a change in the value of $X$ has no affect on the value of $Y$.
For example, the causal network for Example~\ref{xam:forestfire} has the
following form:
%joe25: Another one for Kaoru
\begin{figure}[htb]
\input{psfig}
\centerline{\includegraphics{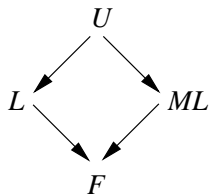}}
%\includegraphics{fig1-new}
%\begin{verbatim}
%
%   U
% /  \
%L    ML
% \  /
%  F
%
%\end{verbatim}
%\input{psfig}
\caption{A simple causal network.}
\end{figure}

\noindent We remark that we occasionally omit the exogenous variables $\vec{U}$
from the causal network.

These causal networks, which are similar in spirit to the Bayesian
networks used to represent and reason about dependences in probability
distributions \cite{Pearl}, will play a significant role in our
definitions.  They are quite similar in spirit to Lewis's
%joe**
%{\em neuron diagrams} \citeyear{Lewis86a}, but
\citeyear{Lewis73a} {\em neuron diagrams}, but
there are significant differences as well.
%%Roughly speaking, neuron diagrams have more expressive power.
%%In the framework we have presented here,
%%a situation would be described a family of neuron diagrams, one for each
%%possible setting of the values of the root nodes in the causal network.
%%Neuron diagrams also have the ability to represent ``inhibitory''
%%signals, which cannot be represented in causal networks.  On the other
%%hand, as we shall see, the structural equations carry all the
%%information we need to do causal reasoning, including all the
%%information about causality and inhibition and counterfactual behavior.
%judea1 rewrite rest of paragraph after "as well though". 9.13
Roughly speaking, neuron diagrams display explicitly
the functional relationships (among variables in $\V$) for
each specific
%joe1: for consistency with earlier notation
%situation $U=u$.
context $\vec{u}$.
%joe+
%The class of functions represented by neuron diagram is
The class of functions represented by neuron diagrams is
limited to those described by
``stimulatory''
%joe1: stimulatory is Hall's word (and presumably Lewis's).  If
%anything, I think it should be "excitatory"
%``excitory''
and ``inhibitory''
binary inputs. Causal networks represent arbitrary functional
relationships, although the exact nature of the functions is
specified in the structural equations and is not encoded
in the diagram.
%joe1
%As we shall see,
The structural equations carry all the
information we need to do causal reasoning, including all the
information about belief, causation, intervention,
and counterfactual behavior.
%continue with next paragraph: " As we shall see"

%%jp5: joe, consider skipping this paragraph
As we shall see, there are many nontrivial decisions to be made when
choosing
%joe1
%the variables to model a particular situation.
the structural model.
%joe**: rest of paragraph new, as is first sentence of next paragraph
\newcomment
One significant decision is the set of variables used.
As we shall see, the events that can be causes and those that can be
caused are expressed in terms of these variables, as are all the
intermediate events.  By way of contrast, in the philosophy literature,
these events can be created on the fly, as it were.  We return to this
point in our examples.  

Once the set of variables is chosen, it must be decided which are
exogenous and which are endogenous.
%joe1: I think there's more to it than this
%Roughly speaking, we can think of the exogenous variables as
%describing the
%background situation, that which we wish to take for granted.
The exogenous variables to some extent encode the background situation,
%joe+
%that which we wish to take for granted.  Other implicit background
which we wish to take for granted.  Other implicit background
assumptions are encoded in the structural equations themselves.
Suppose that we are trying to decide whether a lightning bolt or a
match was the
cause of the forest fire, and we want to take for granted that there is
sufficient oxygen in the air and the wood is dry.
%joe1
%(and the Big Bang occurred).
%joe1: rewrote this material
%then all this is wrapped up into the random variables $u$ (or just
%not modeled at all).
We could model the dryness of the wood by an
exogenous variable $D$ with values $0$ (the wood is wet) and 1 (the wood
is dry).%
\footnote{Of course, in practice, we may want to allow $D$ to have more
values, indicating the degree of dryness of the wood, but that level of
complexity is unnecessary for the points we are trying to make here.}
By making $D$ exogenous, its value is assumed to be given and out of the
control of the modeler.
We could also take the amount of oxygen as an exogenous variable
(for example, there could be a variable $O$ with two values---0, for
insufficient oxygen, and 1, for sufficient oxygen); alternatively, we
could choose not to model oxygen explicitly at all.  For example,
suppose we have, as
before, a random variable $\ML$ for match lit,
and another variable $\WB$ for wood burning,
with values 0 (it's not) and 1 (it is).  The structural equation
%joe8: we're being inconsistent.  Changed f_ to F_ throughout
%$f_{WB}$ would describe the dependence of $\WB$ on $D$
$F_{WB}$ would describe the dependence of $\WB$ on $D$
and $\ML$.  By setting $F_{WB}(1,1) = 1$, we are saying
that the wood will burn if the match is lit and the wood is dry.  Thus,
the equation is
implicitly modeling our assumption that there is sufficient oxygen for
the wood to burn.

%joe25:
We remark that, according to the definition in
Section~\ref{sec:actcaus}, only endogenous variables
can be causes or be caused.  Thus, if no variables
encode the presence of oxygen, or if it is encoded only in an exogenous
variable, then oxygen cannot be a cause of the wood burning.
If we were to explicitly model the amount of oxygen in the air (which
certainly might be relevant if we were analyzing fires on Mount
Everest), then $F_{WB}$ would also take values of $O$ as an argument,
%joe25
and the presence of sufficient oxygen might well be a cause of the wood
burning.%
\footnote{If there are other variables in the model, these would be
arguments to $F_{WB}$ as well;
we have ignored other variables here just to make our point.}
%random
%variables, but in practice it does not matter, since we do not reason
%explicitly about the background conditions.  On the other hand, if we
%do want
%to reason explicitly about, say, how dry the wood is, then there would
%be a random variable corresponding to the wetness of the wood, with (at
%least) two values, {\em dry\/} and {\em wet}.

Besides encoding some of our implicit assumptions, the
structural equations can be viewed as encoding the causal mechanisms at
work.  Changing the underlying causal mechanism can affect what counts
as a cause.
%joe25
%We shall see several
%examples of the importance of the choice of random variables and the
%choice of causal mechanism in Section~\ref{sec:examples}.
%joe19: reworded
%We shall see several
Section~\ref{sec:examples} provides several
examples of the importance of the choice of random variables and the
%choice of causal mechanism in Section~\ref{sec:examples}.
choice of causal mechanism.
%joe19
%Note that we are not claiming that it is
It is not
%joe1
always
straightforward to decide what
%joe1
%random variables should be used to model a given situation, nor that
%joe+
%the ``right'' causal model is in a given situation, nor that it is
the ``right'' causal model is in a given situation, nor is it
%joe1
always
obvious which of two causal models is ``better'' in some sense.
These
%joe1
%are
may be
difficult decisions
%joe1
%that
and
often lie at the heart of determining
actual causality in the real world.
%joe1
Nevertheless, we believe that the tools we provide here should help in
making principled decisions
%jp11 adding
about those choices.

%joe4: cut for UAI
\paragraph{Syntax and Semantics:}
To make the definition of actual
causality precise, it is helpful to have a logic with a formal syntax.
Given a signature $\S = (\U,\V,\R)$, a formula of the form $X = x$, for
$X \in V$ and $x \in \R(X)$, is called a {\em primitive event}.   A {\em
basic causal formula (over $\S$)\/} is one of the form
%$[Y_1 \gets y_1, \ldots, Y_k \gets y_k] \phi(\vec{u})$
%joe20: added comma at end of line
$[Y_1 \gets y_1, \ldots, Y_k \gets y_k] \phi$,
where
%joe10: itemized to make it easier to parse
\begin{itemize}
\item $\phi$ is a Boolean
%where $\phi(\vec{u})$ is a Boolean
%combination of formulas of the form $X(\vec{u}) = x$, where $Y_1,
combination of primitive events,
%joe10: redundant
%formulas of the form $X = x$,
\item $Y_1,
%joe27: I'm not sure where X is coming from
%\ldots, Y_k, X$ are variables in $\V$, with $Y_1, \ldots, Y_k$ are distinct,
%\item $x \in \R(X)$, and
\ldots, Y_k$ are distinct variables in $\V$, and
\item $y_i \in \R(Y_i)$.
\end{itemize}
%and $\vec{u}$ is a vector of values for all the variables in $\U$.
Such a formula is
%joe17
%typically
abbreviated
%as $[\vec{Y} \gets \vec{y}]\phi(\vec{u})$.
as $[\vec{Y} \gets \vec{y}]\phi$.
The special
case where $k=0$
%joe4
%(which is allowed)
is abbreviated as
%$[\true]\phi(\vec{u})$. Intuitively, $[\true]\phi(\vec{u})$ says that
%joe+
%$\phi)$.
$\phi$.
%joe4
%Intuitively, $\phi$ says that $\phi$ holds in the actual world.
%$[Y_1 \gets y_1, \ldots, Y_k \gets y_k] \phi(\vec{u})$ says that
%joe16
Intuitively,
$[Y_1 \gets y_1, \ldots, Y_k \gets y_k] \phi$ says that
%$\phi(\vec{u})$ holds in the counterfactual world that would arise if
$\phi$ holds in the counterfactual world that would arise if
%the endogenous variables are set to $\vec{u}$ and $Y_i$ is set to
%joe+
%$Y_i$ is set to $y_i$, $i = 1,\ldots,k$.
$Y_i$ were set to $y_i$, $i = 1,\ldots,k$.
%joe16
%As we now show, the semantics guarantees that
%this counterfactual world is the one where the values of the variables
%are those in the unique
%solution to the equations obtained after setting the variables
%in $\vec{Y}$ to the values $\vec{y}$.
%joe16: added
A {\em causal formula\/} is a Boolean combination of basic causal
formulas.%
%joe25
\footnote{
If we write $\rightarrow$ for conditional implication, then
a formula such as $[Y\gets y] \phi$ can be written as $Y = y \rightarrow
\phi$: if $Y$ were $y$, then $\phi$ would hold.  We use the present
notation to emphasize the fact that, although we are viewing $Y \gets y$
as a modal operator, we are not giving semantics using the standard
possible worlds approach.}
%joe25: Judea, we could reference your discussion in one of the papers
%with Galles on the connection to Lewis' approach but, to be honest, I'm
%not convinced that that discussion gets at the essence of the issues

%joe16
%A basic causal formula is true or false in a causal model, given a
%context $\vec{u}$.

A causal formula $\psi$ is true or false in a causal model, given a
context.
We write $(M,\vec{u}) \sat \psi$ if
%%the causal formula
$\psi$ is true in
causal model $M$ given context $\vec{u}$.%
%joe25
%\footnote{We remark that in \cite{GallesPearl97,galles:pea98,Hal20},
\footnote{We remark that in \cite{GallesPearl97,Hal20},
the context $\vec{u}$ does not appear on the left-hand side of $\sat$;
%joe+
%rather, it is incorporated in the formula $\psi$ on the right-hand (so
rather, it is incorporated in the formula $\psi$ on the right-hand side (so
that a basic formula becomes $X(\vec{u}) = x$).
%jp13 Adding a pacifying sentence that I owe to the faithful
%readers of my book who invested time in acquiring the
%subscript notation -- a very useful one after all.
Additionally, Pearl \citeyear{pearl:2k} abbreviated
$(M,\vec{u}) \sat [\vec{Y} \gets \vec{y}](X = x)$
%joe12:
% to read:
as
$X_{y}(u)=x$. The presentation here
%seems more natural for our purposes, although they are technically
makes certain things more explicit, although they are technically
equivalent.}
%jp13 The more I work with it, the less I agree with the last sentence.
%I would feel more comfortable if we replaced "more natural"
%with "more explicit".
%joe12: how do you feel about the change above?  By the way, my
%complaint about the notation is the following.  Suppose that there are
%two variables, Y and Z, both of which have domains {0,1}.  What does
%X_0(u) = 1 mean in your notation?  Is it Y that was set to 0, or Z?
%$M \sat [\vec{Y} \gets \vec{y}](X(\vec{u}) = x)$ if in the
$(M,\vec{u}) \sat [\vec{Y} \gets \vec{y}](X = x)$ if
%joe16: moved up from below
the variable $X$ has value $x$
in the
%joe4
%(unique, since we are dealing with recursive equations) solution
%joe20
%(unique, since we are dealing with recursive models) solution
unique (since we are dealing with recursive models) solution
to
%joe8
the equations in
$M_{\vec{Y} \gets \vec{y}}$ in context $\vec{u}$ (that is, the
unique vector
of values for the exogenous variables that simultaneously satisfies all
equations $F^{\vec{Y} \gets \vec{y}}_Z$, $Z \in \V - \vec{Y}$,
with the variables in $\U$ set to $\vec{u}$).
$(M,\vec{u}) \sat
[\vec{Y} \gets \vec{y}]\phi$ for an arbitrary Boolean combination
$\phi$ of formulas of the form $\vec{X} = \vec{x}$ is defined
similarly.
%joe16: added
We extend the definition to arbitrary causal formulas, i.e., Boolean
combinations of basic causal formulas, in the
%joe20
%standard way.
obvious way.

%joe2: cut for abstract
%joe6*: also cut for full paper; I'm not sure it's worth saying
\commentout{
We can now make precise the fact that if $Y$ is not a descendant of $X$
in the causal network, then a change in the value of $X$ can have no
effect on the value of $Y$.  For example, if $(M,\vec{u}) \sat Y = y^*$
(that is, if in the actual world, given context $\vec{u}$, $Y$ has value
$y^*$) then $(M, \vec{u}) \sat [X \gets x](Y = y^*)$.  That is,
if in the actual world, given context $\vec{u}$, $Y$ has value
$y^*$, then $Y$ continues to have value $y^*$ if we set $X$ to $x$.  The
following lemma generalizes this.

\lem\label{indep}
If $M$ is a causal model, $Y$ is not a descendant of $X$ in the
causal network corresponding to $M$, $X_1, \ldots, X_k, X$ are distinct
variables in $\V$, and $(M, \vec{u}) \sat [X_1 \gets x_1, \ldots, X_k
\gets x_k](Y = y^*)$, then $(M, \vec{u}) \sat [X_1 \gets x_1, \ldots, X_k
\gets x_k, X \gets X](Y = y^*)$. \elem
}

Note that the structural equations are deterministic.  We can make
sense out of probabilistic counterfactual statements, even conditional
ones (the probability that
$X$ would be 3 if $Y_1$ were 2, given that $Y$ is in fact 1) in this framework
(see \cite{BalkePearl94}), by putting a probability
%joe6*
%measure on the set of possible values of the exogenous variables,
on the set of possible contexts.
%but this will not be
%necessary for our discussion.
This will not be necessary for our discussion of causality, although it
will play a more significant role in the discussion of explanation.

%joe6*
%\section{Causes and Explanations}\label{sec:actcaus}
%joe16: capitalized for consistency
%\section{The definition of cause}\label{sec:actcaus}
\section{The Definition of Cause}\label{sec:actcaus}

With all this notation in hand, we can
now give
%joe19:
a preliminary version of the
definition of actual cause (``cause'' for
short).
%joe6:
%and explanation.  We start with cause.
%
We want to make
sense out of statements of the form ``event $A$ is
%joe1
%the actual cause
an actual cause of
%joe16: for consistency
%event $B$ (in context $\vec{u}$)''.%
event $\phi$ (in context $\vec{u}$)''.%
%joe10: there was some confusion about this in my talk with the
%philosophers
\footnote{Note that we are using the word ``event'' here in the
standard sense of ``set of possible worlds''
%jp11 adding
%joe12
%(as opposed to a transition between states of affairs);
(as opposed to ``transition between states of affairs'');
essentially we are
identifying events with propositions.}
As we said earlier, the context
%joe19: weakened
%is the background information that has usually been left implicit in
%many other treatments of causality; here we make it explicit.
is the background information.  While this has been left implicit in some
treatments of causality, we find it useful to make it explicit.
%joe6*: added; this would have helped me.
The picture here is that the context (and the structural equations) are
given.
%joe10
Intuitively, they encode the background knowledge.
All the relevant
%joe10
%facts
events
are known.  The only question is picking
out which of them are the causes of $\phi$
%jp7: added
or, alternatively, testing whether a given set of
%joe10
%facts
events
can be considered the cause of $\phi$.%
\footnote{We
%joe7: slight rewrite
%will use past tense and present tense interchangebly
use both past tense and present tense in our examples
(``was the cause'' versus ``is the cause''),
with the usage
depending on whether the scenario implied by  the
%joe+
%context $\vec{u}$) is perceived to have taken place
context $\vec{u}$ is perceived to have taken place
in the past or to persist through the present.}

%joe1: simplified!
%jp5: reintroducing vector causes.
The types of events that we
%joe16
%consider
allow
as actual causes are ones of
the form
$X_1 = x_1 \land \ldots \land X_k = x_k$---that is, conjunctions of
primitive events; we typically abbreviate this as $\vec{X} = \vec{x}$.
%joe2:
%or, even more succinctly, as $x$.
%Only a primitive event of the form $X=x$ can be an actual cause.
The events that can be caused are arbitrary Boolean combinations
%of primitive events.
%jp5
%joe2: shortened
%of primitive events, to be denoted by the symbol $\phi$.
%joe16
%$\phi$
of primitive events.
%joe6*
%joe16: cut this from here
%In fact, in all our examples, the cause is just a primitive event of the
%form $X=x$.
%This is typically the case in all the examples from the
%philosophy literature as well.
%joe16: oops!
%However, as we show in the appendix,
%we gain some important generality by allowing conjunctions of primitive
%events as causes.
We might consider generalizing further to allow
disjunctive
%joe25
causes.
%joe13: weakened
%We do not believe that we lose anything by disallowing disjunctive
We do not believe that we lose much by disallowing disjunctive
causes here.   Since for causality we are
assuming that the structural model and all the relevant facts are known,
the only reasonable definition of ``$A$ or $B$ causes $\phi$'' seems to
be that ``either $A$ causes $\phi$ or $B$ causes $\phi$''.  There are
no truly disjunctive causes once all the relevant facts are known.%
%joe13: added footnote
\footnote{Having said that, see the end of
Example~\ref{xam:arson} for further discussion of this issue.
Disjunctive {\em explanations\/} seem more interesting,
although we cannot handle them well in our framework; 
%joe19
%Section~\ref{sec:explanation}.}
%joe**
%see Part II of this paper.}
these are discussed in Part II.}
%joe8: I think this is important, since so much has been written about it
%joe18: cut; said this earlier
%For similar reasons, there is no need to talk about probabilistic
%causality,
%since the model is assumed
%%joe16
%to be
%deterministic
%(although, as we shall see, probability will play a much more
%important role in explanation;
%%joe10: added
%see also the discussion of ranking
%functions

%jp5: Joe, I am going to erase all our previous writings,
% and write this  section afresh. I will also use some subscript
% notation so that I can think while writing. I will leave
%the decision to you as to which notation you want to use in the
%uai paper.
%%\begin{definition}
\dfn\label{actcaus}
%joe4: I think it actually looks better on one line, and saves space
%(Actual cause)\\
%joe19
%(Actual cause)
(Actual cause; preliminary version)
$\vec{X} = \vec{x}$ is an {\em actual cause of $\phi$ in
$(M, \vec{u})$ \/} if the following
%joe+
%three conditions hold:
three conditions hold.
\newcounter{enumerate}
\begin{description}
%{{\rm AC\arabic{enumerate}.}}{\usecounter{enumerate}
%\setlength{\rightmargin}{\leftmargin}}
%\item\label{ac1}$M, \vec{u} \sat X = x$.
%Kaoru, watch for italics and proper formatting,
%joe2: for consistency
%\item\label{ac1} $X(u)=x, \phi(u) = {\rm true}$
\item[{\rm AC1.}]\label{ac1} $(M,\vec{u}) \sat (\vec{X} = \vec{x}) \land
\phi$.
(That is, both $\vec{X} = \vec{x}$ and $\phi$ are true in the actual
world.)
\item[{\rm AC2.}]\label{ac2}
%joe2: made notation consistent
%There exists a partition $(Z,W)$ of $V$, with $X$ in $Z$, and some
%setting $(x',w')$ of the variables in $(X,W)$, such that
There exists a partition $(\vec{Z},\vec{W})$ of $\V$ with $\vec{X}
\subseteq \vec{Z}$ and some
setting $(\vec{x}',\vec{w}')$ of the variables in $(\vec{X},\vec{W})$
such that
%joe4: put this here, to avoid repetition
%joe11
%if $(M,\vec{u}) \sat \vec{Z} = \vec{z}^*$, then
if $(M,\vec{u}) \sat Z = z^*$ for 
%joe**
all
$Z \in \vec{Z}$, then
%joe25
both of the following conditions hold:
\begin{description}
%joe2: regrouped clauses; it makes for a slightly different story.
%Let's discuss
\item[{\rm (a)}]
%$(M,\vec{u}) [\vec{X} \gets \vec{x};
%\vec{W} \gets \vec{w'}]\phi$ and
$(M,\vec{u}) \sat [\vec{X} \gets \vec{x}',
\vec{W} \gets \vec{w}']\neg \phi$.
%joe6*: put in stronger definition and cut out discussion.  The paper
%seems long enough already.  I suspect you'll be happy with this.  The
%motivation for considering both definitions is no longer as strong.
%jp7: Wondering: why use the strong version of the condition? we never
%had this one before !!!
%I would much prefer if we do not insist on th Z's changing
%- it is enough to insist on Z non changing in AC2(b),
%If AC2(b) holds, then the Z part of AC2(a) holds (we simply
%move those Z's that did not change in AC2(a) to W.
%it seems inelegant to require in a definition conditions
%that can be derived otherwise. If we want the definition to
%be transparent, we should load the reader with minimum
%conditions to think about.
%jp8: removing: (\land_{Z \in \vec{Z}} (Z \ne z^*)))$.
%)$.
%     \begin{itemize}
%     \item
%In words, under the setting $\vec{W} = \vec{w}'$, the
%transition of $\vec{X}$ from
%joe7: We actually did have the stronger condition before; we proved it
%was equivalent to the current condition (I have the proof commented
%out below).  I accept your reasons for sticking with the "weaker"
%condition.  Should we then prove the equivalence of the two conditions
%in an appendix?  (We definitely don't want to clutter up the main
%text with it.)
%jp8. agree, I will just add a remark in footnote 7
In words, changing $(\vec{X},\vec{W})$ from $(\vec{x},\vec{w})$ to
$(\vec{x}',\vec{w}')$ changes
%joe16: added comma
%joe25: changed to period
$\phi$ from true to false.
%jp7: If you object to my suggestion we would need to add here:
%and changes every element of $\vec{Z}$ from $z^*$ to some other value
%     \end{itemize}
%joe11: rewrote, as we discussed
\item[{\rm (b)}]
$(M,\vec{u}) \sat [\vec{X} \gets
%joe**: Judea, please note! I'm allowing any subset of $W$ here; this is
%important to kill an example in your paper with Mark Hopkins.  See the
%appendix, where I show this.
%\vec{x}, \vec{W} \gets \vec{w}', \vec{Z}' \gets \vec{z}^*]\phi$ for
%all subsets $\vec{Z'}$ of $\vec{Z}$.
%In words, setting $\vec{W}$ to $\vec{w}'$ should have no 
\vec{x}, \vec{W'} \gets \vec{w}', \vec{Z}' \gets \vec{z}^*]\phi$ for 
\newcomment
all subsets $\vec{W'}$ of $\vec{W}$ and all subsets $\vec{Z'}$ of $\vec{Z}$. 
In words, setting any subset of variables in $\vec{W}$ to their values
in $\vec{w'}$ should have no 
effect on $\phi$,
as long as $\vec{X}$ is kept at its current value $\vec{x}$,
even if all the variables in an arbitrary subset of $\vec{Z}$ are set to
their original values in the context $\vec{u}$.
%joe11: footnote is now unnecessary
%\footnote{In all of our examples, $\phi$ is a Boolean combination of
%$Z = z^*$ for variables $Z \in \vec{Z}$, so the fact that setting
%$\vec{W}$ to $\vec{w}'$ has no effect on $\phi$ follows from the fact
%that it has no effect on any variable in $\vec{Z}$.}
%     \end{itemize}
\end{description}
\item[{\rm AC3.}] \label{ac3}
$\vec{X}$ is minimal; no subset of $\vec{X}$ satisfies
conditions AC1 and AC2.
Minimality ensures that only those elements of
the conjunction $\vec{X}=\vec{x}$ that are essential for
changing $\phi$ in AC2(a) are
considered part of a cause; inessential elements
are pruned. \bbox
\label{def3.1}  %%changed from {def2.2}
%joe4: moving end defintion marker
\end{description}
%\edfn
\end{definition}
%joe8:
%joe11: cut this; it's distracting and unnecessary here
%For future reference, we say that $\vec{X} = \vec{x}$ is a {\em weak
%cause\/} of $\phi$ in $(M,\vec{u})$ if AC1 and AC2 hold, but not
%necessarily AC3.

%joe19*
Although we have labeled this definition ``preliminary'', it is actually
very close to the final definition.  We discuss the final definition in
Section~\ref{sec:refined}, after we have considered a few examples.

%joe13: added; I couldn't find a better place to put it.  It didn't seem
%to fit in anywhere as a footnote.  If you want to move it, that would
%be fine with me.
%joe19: moved this below:
%Note that we allow $X=x$ to be a cause of itself.  While we do not find
%such trivial causality terribly bothersome, it can be avoided by requiring
%that $\vec{X} = \vec{x} \land \neg \phi$ be consistent for $\vec{X} =
%\vec{x}$ to be a cause of $\phi$.
%joe2: I made a lots of changes in the next paragraph to reflect changes
%in the presentation of the definition.
%joe7: I understand the issue although I'm not sure I understand the
%notion of contrastive explanation.  I agree that this should be at
%best a comment.  Could you write the first draft?

The core of this definition lies in
%joe4
%clauses AC2(a) and AC2(b).
AC2.
%joe2: added
%jp6: revised
Informally, the variables in $\vec{Z}$ should be thought of as
describing the
``active causal
process'' from $\vec{X}$ to $\phi$
%jp8 adding, to make the connection to Lewis more explicit.
%joe8: since we reference lewis:73 and and lewis:86, I suggest we
%combine references and be explicit.  I modified lewis:86 in what I hope
%is an appropriate way.  Please check.
%(also called ``intrinsic process'' in Lewis \cite{lewis:86}).
%joe**: Judea, the only place I could find a discussion of ``intrinsic''
%in the context of ``process'' was in Appendix D of Lewis86a.  Is this
%what you meant?
%(also called ``intrinsic process'' by Lewis \citeyear{Lewis86a}).%
(also called ``intrinsic process'' by Lewis \citeyear[Appendix D]{Lewis86a}).%
\footnote{Recently, Lewis \citeyear{Lewis00} has abandoned attempts
%joe8
%to adequately define ``intrinsic processes''.
to define ``intrinsic process'' formally.
%jp10 I think the reader might be wondering here about:
Pearl's ``causal beam'' \cite[p.~318]{pearl:2k}
is a special kind of active causal process, in which AC2(b) is expected
%to hold for all settings $w'$, not necessarily the one used in (a).}
%jp13 changing last sentence in light of new AC2
%joe12: what's 0 here?  Did you mean Z' = Z*?
%to hold (with $Z'=0$) for all settings $w'$ of $W$, not necessarily
%joe**
%to hold (with $\vec{Z}=\vec{z}^*$) for all settings $w'$ of $W$, not
to hold (with $\vec{Z}=\vec{z}^*$) for all settings $\vec{w}''$ of
$\vec{W}$, not 
necessarily 
%joe**: added next line
%the one used in (a).}
the setting $\vec{w}'$ used in AC2(a).}
These are the variables that
mediate between $\vec{X}$ and $\phi$.
%joe4: cut; this is the wrong place for it.  See below
%jp6:adding a footnote. Lewis and his students have
%struggled  for 25 years to define this concept, I cannot
%miss an opportunity to advertize that we now have it.
%joe3: changed ``the'' to ``a''
%\footnote{Formally, we may define the ``active causal
%process'' from $\vec{X} = \vec{x}$ to phi as the minimal
%jp7: Suggest we restore the footnote in liu of jp6 above.
%joe8: Moved this out of footnote and referred to proof now in the
%appendix.
%\footnote{Formally,
Indeed, we can define an {\em active causal
process\/} from $\vec{X} = \vec{x}$ to $\phi$ as a minimal set
$\vec{Z}$ that satisfies AC2.
%joe7: how about adding:
%jp8:adding "Indeed" and a remark
%joe8: Took out of footnote to emphasize it a bit more and expanded.
We would expect that the variables in an active causal process are all on
a path from a variable in $\vec{X}$ to a variable in $\phi$.  This is
indeed the case.  Moreover, it
can easily be shown that the variables in an active causal process all
change their values when $(\vec{X},\vec{W})$ is set to
$(\vec{x}',\vec{w}')$ as in AC2.
Any variable that does
not change in this transition can be moved to $W$, while
retaining its value in $w'$---the remaining variables
in $\vec{Z}$ will still satisfy AC2.
%joe8: added; Stalnaker was interested in this.
(See the appendix for a formal proof.)
AC2(a) says that
there exists a setting $\vec{x}'$ of $\vec{X}$ that changes $\phi$ to
%jp6: rephrasing
$\neg \phi$, as long as the variables
not involved in the causal process ($\vec{W}$)
take on value $\vec{w}'$.
%jp6: adding, since it is not shown explicitly in AC2(a).
%joe11: cut this sentence from here, since the rest of the paragraph is
%about AC2(a).  Moved it to the next paragraph.
%AC2(b) ensures that that $\vec{X}$ alone is sufficient to
%bring about this change.
AC2(a) is reminiscent of the traditional counterfactual criterion
%joe**
%of Lewis \citeyear{Lewis86a},
of Lewis \citeyear{Lewis73a},
%joe+
%according to which $\phi$ should be false if it were not for
according to which $\phi$ would be false if it were not for
%joe2
%$x$.
$\vec{X}$ being $\vec{x}$.
However, AC2(a) is more permissive than the traditional criterion;
it allows the dependence of $\phi$ on $\vec{X}$ to be tested
%while the variables $\vec{W}$ are held constant at some setting
%$\vec{w}'$.
%jp6: rephrasing, we later refer to "such contingencies"
%and readers familiar with the concept need assurance
%we are not introducing anything new.
under special circumstances
%joe4
% called {\em structural contingencies} \citeyear{pearl:98c,pearl:2k},
%jp7: erasing duplicates
%that Pearl \citeyear{pearl:98c,pearl:2k} calls {\em structural
%contingencies},
in which the variables $\vec{W}$ are held constant at some setting
$\vec{w}'$.
%joe2: cut for abstract
%jp7: slight rearrangement of Hitchcock ahead
This modification of the traditional criterion was proposed
by Pearl \citeyear{pearl:98c,pearl:2k}
%joe6
%(and, more recently, Hitchcock \citeyear{hitchcock:99}),
and was named {\em structural contingency}---an alteration of
the model $M$ that involves the breakdown of some mechanisms
(possibly emerging from external action) but no
change in the context $\vec{u}$.
The need to invoke such
contingencies will be made clear in Example~\ref{xam:arson},
and is further supported by the examples of Hitchcock
\citeyear{hitchcock:99}.

%joe11: added some stuff here
AC2(b),
%joe18
which has no obvious analogue in the literature,
is an attempt to counteract the ``permissiveness'' of AC2(a) with
regard to structural contingencies.  Essentially, it ensures that
$\vec{X}$ alone suffices to bring about the change from $\phi$ to $\neg
%jp13 adding emphasis
\phi$; setting $\vec{W}$ to $\vec{w'}$ merely eliminates
%joe12: X -> \vec{X}
spurious side effects that tend to mask the action of $\vec{X}$.
It captures the fact that setting
$\vec{W}$ to $\vec{w}'$ does not affect the causal process
%by AC2(b), which says that changing $\vec{W}$ from $\vec{w}$
%to $\vec{w}'$ has no effect on $\phi$ or any of the variables in
%$\vec{Z}$ as long as $\vec{X}$ maintains its current value $\vec{x}$.
by requiring that changing 
%joe**
\newcomment
the values of any subset of the variables in
$\vec{W}$ from $\vec{w}$ to $\vec{w'}$ has no
effect on the value of $\phi$.%
\footnote{This version of AC2(b) differs slightly from that in an
earlier version of this paper  \cite{HPearl01a}.  See
Appendix~\ref{app:AC2b} for more discussion of this issue.}
Moreover, although the values in the
variables $\vec{Z}$ involved in the causal process may be perturbed by
the change, the perturbation has no impact on the value of $\phi$.
%joe11:
%In other words,
The upshot of this requirement is that
we are not at liberty to conduct
the counterfactual test of AC2(a) under an arbitrary alteration
of the model.  The alteration considered
%may change only the
%manner in which $X$ affects $\phi$;
%joe18
%must not have an effect of its own on
must not affect
%joe2
%any variable that is not chosen for alteration.
the causal process.
Clearly, if the contingencies considered are limited to
``freezing'' variables at their
actual value
%joe2
%$w' = W(u)$,
%jp7 adding:
(a restriction used by Hitchcock \citeyear{hitchcock:99}),
so that $(M,\vec{u}) \sat \vec{W} = \vec{w}'$,
then AC2(b) is
satisfied automatically.
%joe2: cut; I'm not sure what this means
%and any set $W$ can be employed
%in the counterfactual test of AC2(a).
However, as the examples below show,
genuine causation may sometimes be revealed only
through a broader class of counterfactual tests
in which variables in $\vec{W}$ are set to values
that differ
from their actual values.

%joe18
%joe+
%In \cite{pearl:2k}, a notion of {\em contributory cause\/} is defined
%as well as actual cause.  Roughly speaking, if
Pearl \citeyear{pearl:2k} defines a notion of {\em contributory cause\/}
in addition to actual cause.  Roughly speaking, if
AC2(a) holds only with $\vec{W} = \vec{w}' \ne \vec{w}$,
%joe25
%the $A$ is a contributory cause of $B$; actual
then $\vec{X} = \vec{x}$ is a contributory cause of $\phi$; actual
causality holds only if $\vec{W} = \vec{w}$.
%joe11: added rest of paragraph
Interestingly, in all our examples in Section~\ref{sec:examples},
changing $\vec{W}$ from $\vec{w}$ to $\vec{w}'$ has no impact on the
value of the variables in $\vec{Z}$. That is, $(M,\vec{u}) \sat [\vec{W}
\gets \vec{w}'](Z = z^*)$ for all $Z \in \vec{Z}$.  Thus, to check
AC2(b) in these examples, it suffices to show that
$(M,\vec{u}) \sat [\vec{X} \gets
\vec{x}, \vec{W} \gets \vec{w}']\phi$.  We provide an example in the
appendix to show that there are cases where the variables in $\vec{Z}$
can change value, so the full strength of AC2(b) is necessary.

%joe19*: moved stuff from the UAI paper here plus some more material on
%``strong causality''; please check
We remark that, like the definition here, the causal beam definition
\cite{pearl:2k}
tests for the existence of counterfactual dependency in an auxiliary
model of the world, modified by a select set of structural contingencies.
However, whereas
%joe25: rewrote slightly
%the  beam criterion selects the choice of contingencies
%depends only on the relationship a variable and its parents in the causal
the contingencies selected by the  beam criterion
depend only on the relationship between a variable and its parents in
the causal diagram, the  current definition selects the modifying
contingencies
based on the specific cause and effect pair
%joe25
%that is
being tested. This refinement permits our definition to
avoid certain pitfalls (see Example~\ref{voting})
that are associated with graphical criteria for actual causation.
In addition, the causal beam definition essentially adds another clause
to AC2, placing even more stringent requirements on causality.
Specifically, it requires
\begin{description}
\item[{\rm AC2(c).}]  $(M,\vec{u}) \sat [\vec{X} \gets \vec{x}, \vec{W}
%joe**
%\gets \vec{w}'']\phi$ for all setting $\vec{w}''$ of $\vec{W}$.
\gets \vec{w}'']\phi$ for all settings $\vec{w}''$ of $\vec{W}$.
\end{description}
AC2(c) says that setting $\vec{X}$ to $\vec{x}$ is enough to force
$\phi$ to hold, independent of the setting of $\vec{W}$.%
%joe26 -- jp0 adding footnote
\footnote{Pearl \citeyear{pearl:2k} calls this invariance
{\em sustenance}.}
We say that
$\vec{X} = \vec{x}$ {\em strongly causes\/} $\phi$ if AC2(c) holds in
addition to all the other conditions.  As we shall see, in many of our
examples, causality and strong causality coincide.  In the cases where
they do not coincide, our intuitions suggest that strong causality is
too strong a notion.

%\end{commentout}

%joe6*: now unnecessary
%Note that the proof shows
%that we can take all the pairs $(\vec{x}_Z,\vec{w}_Z)$ in AC2$'$
%to be the pair $(\vec{x}',\vec{w}')$ of AC2.
%This last property further justifies viewing
%$\vec{Z}$ as an ``active process'' mediating
%between $\vec{X}$ and $\phi$.

%joe11: added
AC3 is a minimality condition.
Heckerman and Shachter
\citeyear{HeckShac} have a similar minimality requirement;
%joe19
%there seems to be no analogue
%in the standard definitions in the philosophy literature.
Lewis~\citeyear{Lewis00} mentions the need for minimality as well.
Interestingly, in all the examples we
have considered, AC3 forces the cause to be a single conjunct of the
form $X=x$.
%joe16
%In an earlier draft of this paper, we
%We conjecture that this is in fact a consequence of our
%although we have not been able to prove it.
Although it is far from obvious,
%joe18; Judea, should we also reference your student?
%this has now been shown to be the case
Eiter and Lukasiewicz \citeyear{EL01}
%joe19: added Hopkins
and, independently, Hopkins
\citeyear{Hopkins01}, have shown
that this is in fact a consequence of our definition.
However, it depends crucially on our assumption that the set
%joe+
%$\V$ of endogenous variables is finite; see the Appendix for further
$\V$ of endogenous variables is finite; see the appendix for further
discussion of this issue.
%joe19
As we shall see, it also depends on the fact that we are using causality
rather than strong causality.

%joe11: added next lines as glue
How reasonable are these requirements?
%joe19: moved X being a cause of itself here.
One issue that some might find inappropriate is
that we allow $X=x$ to be a cause of itself.  While we do not find
such trivial causality terribly bothersome, it can be avoided by requiring
that $\vec{X} = \vec{x} \land \neg \phi$ be consistent for $\vec{X} =
\vec{x}$ to be a cause of $\phi$.
More significantly,
%In particular,
is it appropriate to invoke structural changes in the
definition of actual causation?  The following example may help
illustrate why we believe it is.
%joe2: I think this may be premature.
%while at the same time motivating our definition of explanation.

\begin{figure}[htb]
%\input{psfig}
%\centerline{\includegraphicsfigure1m}
\centerline{\includegraphics{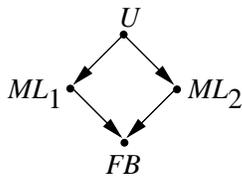}}
%joe8
%\caption{Disjunctive ($M_1$) and conjunctive ($M_2$) scenarios
%of Example~\ref{xam:arson}.} %%example 3.2
\caption{The causal network for $M_1$ and
$M_2$.}
\label{fig1m}
\end{figure}

\xam\label{xam:arson}
%joe1
%If two arsons
Suppose that two arsonists
drop lit matches in different parts of a dry forest, and both
cause trees to start burning.
%joe1: rewrote and expanded material
Consider two scenarios.  In the first, called
%joe25
%``disjunctive,''
the {\em disjunctive scenario},
either match by itself
suffices to burn down the whole forest.  That is, even if only one
match were lit, the forest would burn down.  In the second scenario,
%joe25
%called ``conjunctive,''
called the {\em conjunctive scenario},
both matches are necessary to burn down
the forest; if only one match were lit, the fire would die down
%joe20
before the forest was consumed.
We can describe the essential structure of these two
scenarios using a causal model with four variables:
\begin{itemize}
%joe20: which -> that
\item an exogenous variable $U$ that determines,
among other things, the motivation and state of mind of
the arsonists.  For simplicity, assume that $\R(U) = \{u_{00}, u_{10},
u_{01}, u_{11}\}$; if $U=u_{ij}$, then the first arsonist intends to
start a fire iff $i=1$ and the second arsonist intends to start a fire
iff $j=1$.  In both scenarios $U=u_{11}$.
\item endogenous variables $\ML_1$ and $\ML_2$, each either 0 or 1, where
$\ML_i=0$ if arsonist $i$ doesn't drop the lit match and $\ML_i=1$ if he
does,
for $i =1, 2$.
\item an endogenous variable $\FB$ for forest burns down, with values 0
(it doesn't) and 1 (it does).
\end{itemize}
Both scenarios have the same causal network
(see Figure \ref{fig1m});
they differ only in the equation for $\FB$.
%joe2: shrunk for abstract
%For the disjunctive scenario (Figure \ref{fig1m}(a))
For the disjunctive scenario
we have $F_{FB}(u,1,1) = F_{FB}(u,0,1) =
F_{FB}(u,1,0) = 1$ and $F_{FB}(u,0,0) = 0$ (where $u \in \R(U)$);
%joe2: for abstract
%for the conjunctive scenario (Fig.\ \ref{fig1m}(b))
for the conjunctive scenario
we have $F_{FB}(u,1,1) = 1$ and
$F_{FB}(u,0,0) = F_{FB}(u,1,0) = F_{FB}(u,0,1) = 0$.
%joe6*: I'll need this later.
In general, we expect that the causal
model for reasoning about forest fires would involve many other
variables; in particular, variables for other potential causes
of forest fires such lightning and unattended campfires; here we focus
on that part of the causal model that involves forest fires started by
arsonists.  Since for causality we assume that all the relevant facts
are given, we can assume here that it is known that there were no
unattended campfires and there was no lightning, which makes it safe to
ignore that portion of the causal model.
Denote by $M_1$ and $M_2$ the
%joe6*
(portion of the)
causal models associated with
the disjunctive and conjunctive scenarios, respectively.  The causal
network for the relevant portion of $M_1$ and $M_2$ is described in
Figure~\ref{fig1m}.

Despite the differences in the underlying models, each of $\ML_1=1$
and $\ML_2=1$ is a cause of $\FB=1$
%joe16
%(representing $\phi$)
in both scenarios.  We present the argument for
$\ML_1=1$ here.  To show that $\ML_1=1$ is a cause in $M_1$
let $\vec{Z} =
\{\ML_1,\FB\}$, so $\vec{W} = \{\ML_2\}$. It is easy to see that
the contingency $\ML_2 = 0$ satisfies the two conditions in AC2.
AC2(a) is satisfied because, in the absence of the second
arsonist ($\ML_2 = 0$), the first arsonist is necessary and
sufficient for the fire to occur $(\FB = 1)$.
AC2(b) is satisfied because, if the first match is lit
%joe2
%($X=x$, or $\ML_1=1$)
($\ML_1=1$)
the contingency $\ML_2 = 0$ does not
prevent the fire from burning the forest.
Thus, $\ML_1 = 1$ is a cause of $\FB=1$ in $M_1$.
(Note that we needed to set $\ML_2$ to 0, contrary to
fact, in order to reveal the latent dependence of
$\FB$ on $\ML_1$.
%jp6: we promised justification of "structural contingency"
%so I am adding:
Such a setting constitutes a structural
change in the original model, since it involves the removal
of some structural equations.)

To see that $\ML_1=1$ is also a cause of $\FB=1$ in $M_2$, again let
%joe2
%$\vec{Z} = \{\ML_1,\ML_2,\FB\}$; thus $\vec{W} = \emptyset$.  Since
$\vec{Z} = \{\ML_1,\FB\}$ and $\vec{W} = \{\ML_2\}$.  Since
%joe4: unnecessary
%$M_2,u_{11} \sat [\ML_1 \gets 1, \ML_2 \gets 1](\FB=1)$ and
$(M_2,u_{11}) \sat [\ML_1 \gets 0, \ML_2 \gets 1](\FB=0)$,
AC2(a) is satisfied.
Moreover, since
%joe2
%$\vec{W} = \emptyset$,
the value of $\ML_2$ required for AC2(a) is the same as its current
%value,
%jp13 explicating
value (i.e., $w' = w$),
AC2(b) is satisfied trivially.

%joe2: Moved this discussion into the example, and shortened
%Example \ref{xam:arson}
This example also illustrates the need for
the minimality condition $AC3$.
%joe2: shortened
%in both models. Although
%the conjunction of $\ML_1=1$ and $\ML_2 = 1$ may also
%be tolerated as a ``compound cause'' of $\FB=1$, we cannot
%allow an unchecked proliferation of such factors.
For example, if lighting a match qualifies as the
cause of fire then lighting a match and
sneezing would also pass the tests of AC1 and AC2
and awkwardly qualify as the cause of fire.
Minimality serves here to strip
%joe2: shrunk
%$\vec{X}$ from irrelevant, over-specific details
%and would remove
``sneezing''
and other irrelevant, over-specific details
from the cause.
%$\vec{X} = \vec{x}$.
%joe2: cut; it makes less sense now that our causes can be conjunctions
%\footnote{In the philosophy literature, especially the one following
%the counterfactual tradition of Hume and Lewis, any conjunction
%of causes counted as a cause. It is easy to extend our
%definition to comply with this tradition; we simply define
%a ``composite cause'' as a conjunction of causes that satisfy
%the minimality condition.}

%joe13: added next two paragraphs
It might be argued that allowing disjunctive causes would be useful in this
case to distinguish $M_1$ from $M_2$ as far as causality goes.  A purely
counterfactual definition of causality would make
$\ML_1 = 1 \lor \ML_2 = 1$ a cause of $\FB=1$ in $M_1$ (since, if $\ML_1
= 1 \lor \ML_2 = 1$ were not true, then $\FB=1$ would not be true), but would
make neither $\ML_1 = 1$ nor $\ML_2 = 1$ individually a cause (since,
for example, if $\ML_1 = 1$ were not true in $M_1$, $\FB=1$ would still
be true).  Clearly, our definition does not enforce this intuition.
As is well known
%joe
(and as the examples in Section~\ref{sec:examples}
show) purely counterfactual definitions of causality have other
problems.  We do not have a strong intuition as to the best way to deal
with disjunction in the context of causality, and believe that
disallowing it is reasonably consistent with intuitions.
%jp14. Joe, are we disallowing it? or simply postponing
%the treatment, while alowing future researchers to express
%their intuition, if they desire, by expanding on our definition?.
%joe15: Judea, I think there may be some subtleties here.  I guess we're
%leaving it for future researchers, because at least one of us hasn't
%got the energy right now to think about it!

%joe25: cut this paragraph; we say it again below
%Interestingly, as shown in the companion paper, our
%definition of explanation {\em does\/} distinguish $M_1$ from $M_2$;
%each of $\ML_1 =1$ and $\ML_2 = 1$ is an explanation of $\FB=1$ in $M_1$
%under our definition of explanation, but neither is an explanation of
%$\FB=1$ in $M_2$.  In $M_2$, the explanation of $\FB=1$ is $\ML_1 =1
%\land \ML_2 =1$: both matches being lit are necessary to explain the forest
%burning down.

%joe19*
%joe25
%This example also helps to illustrate the differences betwen causality
%and strong causality.
This example shows that causality and strong causality do not always
coincide.
It is not hard to check that $\ML_1$ and $\ML_2$
are strong causes of $\FB$ in both scenarios.  However, for $\ML_1$ to
be a strong cause of $\FB$ in the conjunctive scenario, we must
include $\ML_2$ in $\vec{Z}$ (so that $\vec{W}$ is empty); if $\ML_2$ is
in $\vec{W}$, then AC2(c) fails.   Thus, with strong causality, it is
no longer the case that we can take $\vec{Z}$ to consist only of
variables on a path between the cause ($\ML_1 = 1$ in this case) and the
effect ($\FB = 1$).

Moreover, suppose that we change the disjunctive scenario slightly by
allowing either arsonist to have guilt feelings and call the fire
department.  If one arsonist calls the fire department, then the forest
is saved, no matter what the other arsonist does.  We can model this by
allowing $\ML_1$ and $\ML_2$  to have a value of 2
%joe25: added
(where $\ML_i = 2$ if arsonist $i$ calls the fire department).
If either is 2,
then $\FB= 0$.  In this situation, it is easy to check
%joe25
that now
neither $\ML_1 =
%joe25
%1$ nor $\ML_2 = 1$ by itself is a cause.
1$ nor $\ML_2 = 1$ by itself is a strong cause of $\FB=1$ in the
disjunctive scenario.
$\ML_1 = 1 \land \ML_2 = 1$ is
a cause, but it seems strange that in the disjunctive scenario,
we should need to take both arsonists dropping a lit match to
%joe25
(strongly)
cause the fire, just
because we
allow for the possibility that an arsonist can call the fire department.
%joe25
Note that this also shows that, in general, strong causes are not always
single conjuncts.
\exam

%joe2: cut for abstract
This is a  good place to illustrate the need for
structural contingencies in the analysis of actual
causation.  The reason we consider $\ML_1=1$ to be a
cause of $\FB=1$ in $M_1$ is that {\em if\/} $\ML_2$ had been 0,
rather than 1, $\FB$ would depend on $\ML_1$. In words,
we imagine a situation in which the second match is not lit,
and we then reason counterfactually that the forest would
not have burned down if it were not for the first match.
\commentout{
But how do we express this formally in a context $u$ in
which $\ML_2=1$ is in
fact true?  To (hypothetically) suppress $\ML_2=1$
in the context created by $u_{11}$, we must use a structural
contingency and imagine that $\ML_2$ is set to 0
by some external intervention (or ``miracle'') that violates
whatever causal laws (or equations) made $\ML_2=1$ in $u_{11}$.
For example, if $u_{11}$ includes the
motivations and conspiratorial plans of the two
%joe1
%arsons,
arsonists,
then $\ML_2$ may be 0 due to
a mechanical failure (say, the match box fell into
a creek) or to arsonist 2 having second thoughts.
We know perfectly well that these changes did not occur,
yet we are committed to contemplating such changes by
the very act of representing our story
in the form of a multi-stage causal model, with each stage
representing an autonomous mechanism.

Recalling that a causal model actually stands, not for one,
but for a whole set of models, one for each
possible setting of the endogenous variables, contemplating
interventional contingencies is
an intrinsic part of every causal thought.
In other words, the autonomy of the mechanisms in
the model means that each mechanism advertises
its  possible breakdown, and these  breakdowns
signal contingencies against which causal expressions
should be evaluated.  It is reasonable, therefore, that we build
such contingencies into the definition of
actual causation.
}%\end{commentout}

Although $\ML_1=1$ is a cause of $\FB=1$ in both the
disjunctive and conjunctive scenarios, the models $M_1$ and $M_2$
%joe6*
%differ in regard to ``explanation''.
differ in regard to explanation, as we shall see in
%joe19
%Section~\ref{sec:explanation}.
Part II of this paper.
In the disjunctive scenario, the lighting
of one of the matches constitutes a reasonable explanation of the forest
burning down; not so in the conjunctive scenario.
Intuitively, we feel that if both matches are needed for
establishing a forest fire, then both $\ML_1=1$ and $\ML_2=1$
together would be required to fully explain the unfortunate
fate of the forest; pointing to just one of these events
would only beg another ``How come?''~question, and would not
stop any serious investigating team from
continuing its search for a more complete answer.

%jp8: I am adding here the remark on "contrastive cause"
%joe8: slight rewrite.  Since we haven't defined explanations formally
%yet, it seems strange to talk about the definition of expblanation
%(I also incorporated your updated version.)
%Finally, one other point concerning the definitions of
%causes and explanations. When we seek a cause or
%an explanation of $\phi$, we often wish to explain, not merely
Finally, a remark concerning a {\em contrastive\/} extension to the
definition of cause. When seeking a cause of $\phi$, we are often not
just interested the occurrence versus nonoccurrence of $\phi$, but also
the manner in which $\phi$ occurred, as opposed to some
alternative way in which $\phi$ could have occurred \cite{hitchcock:96}.
We say, for example,
%joe13 corrected typo
%``$X=x$ caused a fire in June as a opposed to a fire in May.''
``$X=x$ caused a fire in June as opposed to a fire in May.''
%joe8
%In this case,
If we assume that there is only enough wood in the forest for one
forest fire, the two contrasted events, ``fire in May'' and
``fire in June'', exclude but do not complement each other (e.g.,
neither rules out a fire in April.)
Definition~\ref{def3.1} can easily be
extended to accommodate {\em contrastive causation}.  We
define ``$x$ caused $\phi$, as opposed to $\phi'$'',
where $\phi$  and $\phi'$
are incompatible but not exhaustive,
by simply replacing $\neg \phi$
with $\phi'$ in condition AC2(a) of the definition.
%joe8: I would cut this from here; it should be clear anyway.
%The same change applies when we wish to
%define "contrastive explanation" (see Section 5).

Contrast can also be applied to the antecedent, as in
%joe25
%``Susan's, running rather than walking, to music class caused
``Susan's running rather than walking to music class caused
her fall.''
%joe8: I don't understand this.  Why isn't my simpler condition all we
%need?
%jp9 It is  enouph for classifying a causal sentence as true
%or false, but it is not enough for distinguishing STRESS in
%a sentence. We should be able to say that the sentence:
%``Susan's running to MUSIC class caused her fall,''
%^(with stress on music. My proposal permits the
%interpretation of stress in a sentence.
%jp10 connecting next sentence
% and refute the sentence
%joe25: cut: I found this hard to parse
%to be distinguished from the false sentence
%``Susan's running to music rather than history class caused
%her fall.''
%Our definition can accommodate sentences
%of the form ``$X=x$, rather than $Y=y$, caused $\phi$'' by
%insisting that $x'$ in condition AC2 should entail event Y.
%joe**
\newcomment
There are actually two interpretations of this statement.  The first is
that Susan's running is a cause of her falling; moreover, had she walked,
then she would not have fallen.  The second is that, while Susan's
running is a cause of her falling, Susan's walking also would have
caused her to fall, but she did not in fact walk.
%joe**
%We can capture sentences of the form
We can capture both interpretations of 
``$X=x$, rather than
$X=x'$ for some value $x' \ne x$,
%joe**
%caused $\phi$'' 
caused $\phi$ (in context $\vec{u}$ in structure $M$)''.  The first is
%joe**: new 
(1) $X=x$ is a cause of $\phi$ in $(M,\vec{u})$ and
(2) $(M,\vec{u}) \sat [X \gets x'] \neg \phi$;
the second is
%by taking this to mean that
%joe25: added numbers (1) and (2)
%joe**: now using 1', 2'
%(1) $X=x$ caused $\phi$ and
%(2) AC2(b) holds for $X=x'$ and $\phi$.
(1$'$) $X=x$ is a cause of $\phi$ in $(M,\vec{u})$ and
(2$'$) AC2(b) holds for $X=x'$ and $\phi$.
 That is,
the only reason that $X=x'$ is not the cause of $\phi$ is that $X=x'$ is
not in fact what happened in the actual world.%
\footnote{As Christopher Hitchcock [private communication, 2000] has
pointed out, one of the roles of such contrastive statements is to
indicate that $\R(X)$, 
%joe**: added
the set of possible values of $X$, 
should include $x'$.  The sentence does not make 
sense without this assumption.}
%joe9
%$Y=y$
%joe25: cut
%$X=x'$ did not cause $\phi$.
%joe9; added
(More generally, we can make sense of ``$X=x$ rather than $Y=y$ caused
%joe+
%$\phi$.'')
$\phi$''.)
%jp9 a bit ackward to apply this to Suzy. so perhaps we
%should say: ``$X=x$, rather than $X in some set X'$, caused $\phi$''
Contrasting both the antecedent and the consequent components is
straightforward, and allows us to interpret sentences
of the form: ``Susan's running rather than walking to music class
caused her to spend the night in the hospital, as opposed to
her boyfriend's apartment.''

%\end{commentout}

\section{Examples}\label{sec:examples}

In this section we show how our
%joe6: changed back
%joe2
definition of actual causality handles
%definitions of causality and explanation handle
some
examples that have caused problems for other definitions.
%joe2: (for abstract)
%Many of these examples are discussed by Hall~\citeyear{Hall98}.
%joe6: this is the full paper.
%Further examples can be found in the full paper.
%joe10
%joe11: cut
%In Section~\ref{sec:problems}, we consider some examples that cause problems
%for the definition.

\xam\label{xam1} The first example is due to Bennett (and appears in
\cite[pp.~222--223]{sosa:too93}).  Suppose that there was a heavy rain
in April and electrical storms in the following two months; and in June
the lightning took hold.  If it hadn't been for the heavy rain in April,
the forest would have caught fire in May.  The question is whether the
%joe25: I don't think we need the italics here
%April rains caused {\em the} forest fire.  According to  a naive
April rains caused the forest fire.  According to  a naive
counterfactual analysis, they do,
since if it hadn't rained, there wouldn't have
been a forest fire in June.
%joe4: cut for UAI
Bennett says ``That is unacceptable.  A
good enough story of events and of causation might give us reason to
accept some things that seem intuitively to be false, but no theory
should persuade us that delaying a forest's burning for a month (or
indeed a minute) is causing a forest fire.''
%joe2: Lombard comes out of the blue here.
%Lombard agrees, saying
%``It is a good bit of common sense that heavy rains can put out fires,
%they don't start them; it is false to say that the rains caused the
%fire.''

In our framework, as we now show,
it is indeed false to say that the April rains caused
%joe2: I'm not quite sure why this was italicized
%joe25: or here
%{\em the} fire, but they were
the fire, but they were
%joe6*
%part of the reason that there was a fire
a cause of there being a fire
in June, as opposed to May.  This seems to us intuitively right.
%joe2: cut; I'm not sure what this adds.
%(Hall also says
%that the April rains are ``at least in part responsible for the
%presence of the forest in June'' and hence implicitly responsible in
%part  for the fire.)
To capture the situation, it suffices to use a simple model with three
endogenous random variables:
\begin{itemize}
\item $\AS$ for ``April showers'', with two
values---0 standing for did {\em not\/} rain heavily in April and 1
standing for rained heavily in April;
\item $\ES$ for ``electric storms'', with four possible values: $(0,0)$
(no electric storms in either May or June), (1,0) (electric storms in
May but not June), (0,1) (storms in June but not May), and (1,1)
%joe**
%(storms in both April and May);
(storms in both May and June);
\item and $F$ for ``fire'', with three possible values: 0 (no fire at
all), 1 (fire in May), or 2 (fire in June).
\end{itemize}
We do not describe the context explicitly, either here or in the other
examples. Assume its value
$\vec{u}$ is such that it ensures that there is a shower in April, there
are electric storms in both May and June,
there is sufficient oxygen, there are no other potential causes of fire
%joe+
%(like dropped matches), no other inhibitors of fire (alert campers
(such as dropped matches), no other inhibitors of fire (alert campers
setting up a bucket brigade), and so on.  That is, we choose $\vec{u}$
so as to allow us to focus on the issue at hand and to ensure that the
right things happened (there was both fire and rain).

We will not bother writing out the details of the structural
equations---they should be obvious, given the story (at least, for the
context $\vec{u}$); this is also the case for all the other examples in
this section. The causal network is simple: there are edges from
$\AS$ to  $F$ and from $\ES$ to $F$.  It is easy to check
%joe2: I think this is misplaced; also, I'd prefer to include \ES in W,
%since it's not on the causal path from \AS to F
%(using $W=\emptyset; Z=\{\AS,\ES,F\}$) that
%joe+
%that each of the following hold.
that each of the following holds.
\begin{itemize}
%joe2: I'd prefer to call it *a* cause, not the cause
%\item $\AS=1$ is the actual cause of the June fire $(F=2)$,
\item $\AS=1$ is a cause of the June fire $(F=2)$
%joe2: added
(taking $\vec{W}=\{\ES\}$ and $\vec{Z}=\{\AS,F\}$)
but not of fire $(F=2 \vee F=1)$.
%joe**: move this here from below
\newcomment
That is, April showers are not a cause of the fire, but they are a cause of
the June fire.
%\item $E = (1,1)$ is an actual cause of both $F=2$ and $(F=1 \lor F=2)$.
\item $\ES = (1,1)$ is a cause of both $F=2$ and $(F=1 \lor F=2)$.
%joe2: added
Having electric storms in both May and June caused there to be a fire.
\item $\AS=1 \land \ES=(1,1)$ is not a cause of $F=2$, because it
violates the minimality requirement of AC3; each conjunct alone is
a cause of $F=2$.
%joe6*:
Similarly, $\AS=1 \land \ES=(1,1)$ is not a cause of $(F=1 \lor F=2)$.
\end{itemize}
%joe6*: cut from here; moved to the explanation section
%\item $\AS =1 $ is {\em not\/} an explanation of $F=2$, nor is $E=(1,1)$
%(since setting $\AS=0$ results in $F=1$).
%However, $\ES=(1,1)$ is an explanation of $F=1 \lor F=2$ and
%$\AS=1 \land \ES=(1,1)$ is an explanation of $F=2$.
%joe10: added; I think it's worth stressing
%joe**: apparently not true; see Lewis's ``Events''
%The distinction between April showers being a cause of the fire (which
%they are not, according to our analysis) and April showers being a cause
%of a fire in June (which they are) is one that seems not to have been
%made in the discussion of this problem (cf.~\cite{Lewis00});
%nevertheless, it seems to us an important distinction.
%joe4: moved end marker
\exam

%joe1: cut, since I don't think all the results follow
%These results follow directly from the counterfactual definition of
%Lewis \citeyear{Lewis73a};
%Definitions \ref{def3.1} merely contributes
%precision relative to the choice of $\varphi$.

Although we did not describe the context explicitly in
Example~\ref{xam1}, it still played a crucial role.  If we decide that
the presence of oxygen is relevant
%joe6*: rewrote to make it consistent with the later discussion.
%(or, more likely, the presence of
%campers in June who lit a campfire that got out of control),
then we
must take this factor out of the context and introduce it as an explicit
endogenous variables.  Doing so can
%joe6*
%significantly change the
affect the
%joe2
%causality picture, as Example~\ref{xam3} shows.  The next example is
%a simpler variant of Example~\ref{xam3}, and already shows the
%importance of choosing an appropriate granularity in modeling
%the causal process and its structure.
causality picture  (see Example~\ref{xam4}).
%joe19
%as we shall see in Section~\ref{sec:explanation}, it can have an even
%bigger impact on explanations.
The next example
%joe2: cut
%is a simpler variant of Example~\ref{xam3}, and
already shows the
importance of choosing an appropriate granularity in modeling
the causal process and its structure.

\xam\label{xam2} The following story
%joe4
%(again taken from Hall's paper) is
from \cite{Hall98} is
%joe19
%an example of an {\em overdetermined\/} event.
an example of {\em preemption}, where there are two potential causes
of an event, one of which preempts the other.  An adequate definition of
causality must deal with preemption in all of its guises.

\begin{quote}
%jp6: I am shortening the story, just in case we run out
%of page limit due to  my other additions. Similar cuts
%can be apoplied to the other stories from  Hall.
Suzy and Billy both pick up rocks
and throw them at  a bottle.
Suzy's rock gets there first, shattering the
bottle.  Since both throws are perfectly accurate, Billy's would have
shattered the bottle
%joe19
%if Suzy's had not occurred, so the shattering is overdetermined.
had it not been preempted by Suzy's throw.
\end{quote}
%Suzy and Billy, both expert rock-throwers, are engaged in a competition
%to see who can shatter a target bottle first.  They both pick up rocks
%and throw them at the bottle, but Suzy throws hers a split second before
%Billy.  Consequently, Suzy's rock gets there first, shattering the
%bottle.  Since both throws are perfectly accurate, Billy's would have
%shattered the bottle if Suzy's had not occurred, so the shattering is
%overdetermined.  Once the bottle has shattered, however, it cannot do so
%again; thus the shattering of the bottle prevents the process initiated
%by Billy's throw from itself resulting in a shattering.
%\end{quote}

Common sense suggests that Suzy's throw is the cause of the shattering,
but Billy's is not.  This holds in our framework too, but only if we
model the story appropriately.  Consider first a coarse causal
model, with three endogenous variables:
\begin{itemize}
\item $\ST$ for ``Suzy throws'', with values 0 (Suzy does not throw) and
1 (she does);
\item $\BT$ for ``Billy throws'', with values 0 (he doesn't) and
1 (he does);
\item $\BS$ for ``bottle shatters', with values 0 (it doesn't shatter)
and 1 (it does).
%(it shatters as a result of being hit by Suzy's rock), and 2 (it
%shatters as a result of being hit by Billy's rock)
\end{itemize}
Again, we have a simple causal network, with edges from both $\ST$ and
$\BT$ to $\BS$.  In this simple causal network,
%joe18
%$\BT$ and $\BS$ play
$\BT$ and $\ST$ play
absolutely symmetric roles, with $\BS=\ST \vee \BT$;
%joe25
%and there is nothing to distinguish one from the other.
there is nothing to distinguish $\BT$ from $\ST$.
Not surprisingly, both
Billy's throw and Suzy's throw are classified as causes of the
bottle shattering
%joe25
in this model.

The trouble with this model is that it cannot distinguish
the case where both rocks
%joe16: hitting ->
hit the bottle simultaneously (in which case it would be reasonable
to say that both $\ST=1$ and $\BT=1$ are
causes of $\BS=1$) from the case where Suzy's rock
hits first.  The model has to be refined to express this distinction.
%joe19
One way is to invoke a dynamic model \cite[p.~326]{pearl:2k}; this is
discussed below.
A perhaps simpler way to gain expressiveness is to allow $\BS$ to be
three valued, with values 0 (the bottle doesn't shatter), 1
(it shatters as a result of being hit by Suzy's rock), and 2 (it
shatters as a result of being hit by Billy's rock).
We leave it to the reader to check that $\ST = 1$ is a
cause of $\BS = 1$, but $\BT = 1$ is not (if Suzy hadn't thrown but
Billy had, then we would have $\BS = 2$).  Thus, to some extent, this
solves our problem.  But it
%joe1
%bordered on cheating; the answer was
borders on cheating; the answer is almost
programmed into the model by invoking the relation ``as a result of'',
which requires the identification of the actual cause.

\begin{figure}[htb]
\input{psfig}
%\hspace*{2.5in}
\centerline{\includegraphics{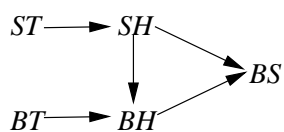}}
\caption{The rock-throwing example.}\label{fig1}
\end{figure}

A more useful choice is to add two new random variables to the model:
\begin{itemize}
\item $\BH$ for ``Billy's rock hits the (intact) bottle'', with values 0
(it doesn't) and 1 (it does); and
\item $\SH$ for ``Suzy's rock hits the bottle'', again with values 0 and
1.
\end{itemize}
With this addition, we can go back to $\BS$ being two-valued.
In this model, we have the causal network shown in Figure \ref{fig1},
with the arrow
$\SH \rightarrow \BH$ being inhibitory; $\BH=\BT \wedge \neg \SH$
%joe6: added
(that is, $\BH=1$ iff $\BT=1$ and $\SH=0$).
Note that, to simplify the presentation, we have omitted the exogenous
variables from the causal network in Figure~\ref{fig1}; we do so in some
of the subsequent figures as well.
%joe10
%joe25
%In addition, we have only given the arrows for the particular context of
In addition, we have given the arrows only for the particular context of
interest, where Suzy throws first.  In a context where Billy throws
first, the arrow would go from $\BH$ to $\SH$ rather than going from
$\SH$ to $\BH$, as it does in the figure.

%joe1: changed all these (otherwise figure may come out at the end of
%the paper if it doesn't fit in right in the text).
%\begin{figure}[h]

%joe2
%Now it is the case that $\ST=1$ is the actual cause of $\BS=1$.
Now it is the case that $\ST=1$ is a cause
%joe6
%and an explanation
of $\BS=1$.
%joe2: I think this is a typo
%To satisfy AC2(b), we choose $\vec{W}=\{\BH\}$ and $w'=0$
%joe4
%To satisfy AC2(b), we choose $\vec{W}=\{\BT\}$ and $w'=0$
To satisfy AC2, we choose $\vec{W}=\{\BT\}$ and $w'=0$
and note that, because $\BT$ is {\em set} to 0, $\BS$ will track
%joe2
%the truth value of $\ST$.  Also note
the setting of $\ST$.  Also note
that $\BT=1$ is not a cause
%joe6
%(or an explanation)
of $\BS=1$; there is no
partition
$\vec{Z} \cup \vec{W}$ that
%would satisfy both AC2(a) and AC2(b).
satisfies AC2.
%jp13 elaborating in view of new AC2
%joe**
%Attempting the symmetric choice $\vec{W}=\{\BT\}$ and $w'=0$
Attempting the symmetric choice $\vec{W}=\{\ST\}$ and $w'=0$
would violate AC2(b) (with $\vec{Z}' = \{\BH\}$), because $\phi$
%joe12
%becomes zero when we set $ST = 0$ and restore Z' to
becomes false when we set $\ST = 0$ and restore $\BH$ to
its current value of 0.

This example illustrates the need for invoking
%joe12: added math mode; slight change in English
%subsets of $\vec{Z}$ in AC2(b) (additional motivations are
%shown in Figure 9). The weaker condition,
subsets of $\vec{Z}$ in AC2(b).  (Additional reasons are
provided in Example~\ref{voting} in the appendix.)
$(M,\vec{u}) \sat [\vec{X} \gets \vec{x}, \vec{W} \gets \vec{w}']\phi$
%joe12
%would have been satisfied by the partition just considered, and
holds if we take $\vec{Z} = \{\BT,\BH\}$ and $\vec{W} = \{\ST,\SH\}$,
and thus without the requirement that AC2(b) hold for all subsets of
$\vec{Z}$, $\BT=1$ would have qualified as a cause of $\BS=1$.
%joe12
%Insisting on $\phi$ remaining unchanged
Insisting that $\phi$ remains unchanged
%joe12:
%under both $Z'= z*$ and $W = w'$ prevents us from choosing contingencies
when both $\vec{W}$ is set to
%joe25: typo
%$\vec{w}'$ and $\vec{Z}'$ is set to $\vec{z}$ (for an arbitrary %subset
$\vec{w}'$ and $\vec{Z}'$ is set to $\vec{z}^*$ (for an arbitrary subset
$\vec{Z}'$ of $\vec{Z}$)
prevents us from choosing contingencies
$\vec{W}$ that interfere with the active causal paths from $\vec{X}$ to
$\phi$.
%joe18: Judea, is this OK?  Please check!
%joe19: cut
%Note that, because of the quantification in Heckerman and Shachter's
%\citeyear{HeckShac} definition, both the variables $\ST$ and $\BT$ cause
%$\BS$.  The subtleties in this example disappear at the level of
%variables.
%joe6
%or EX2.
%\exam

%joe8: added paragraph; I want to stress this point.
%jp13 This paragraph seems to be taken from an old example that is
%no lonnger used. Also, the point is emphasized in the new
%examples. I suggest to delete this paragraph.
%joe12: Hmm ... this paragraph is meant to emphasize the need (stated
%two paragraphs ago) for adding new random variables to the model.
%I'd prefer to keep it in, because I don't think we really made this
%point elsewhere.  Where else do you think we make it?  In any case, if
%we keep it, we'll need different glue, because of the paragraph you inserted.
%I rewrite the glue a little.  If you still want to cut it, perhaps we
%can discuss it by telephone.  (I agree that Lewis makes this point too,
%but that doesn't mean that it isn't true for us.)
This example also emphasizes an important moral.
%There is an important moral here.
If we want to argue in a case of
%joe19
%overdetermination
preemption
that $X=x$ is the cause of $\phi$ rather than $Y=y$,
then there must be a random variable ($\BH$ in this case) that takes on
different values depending on whether $X=x$ or $Y=y$ is the actual
cause.  If the model does not contain such a variable, then it will not
be possible to determine which one is in fact the cause.  This is
certainly consistent with intuition and the way we present evidence.  If
we want to  argue (say, in a court of law) that it was $X$'s shot that
killed $C$ rather than $Y$'s, then we present evidence such as the
%joe25
%bullet entered $C$ from the left side (rather than the right side, which
bullet entering $C$ from the left side (rather than the right side, which
is how it would have entered had $Y$'s shot been the lethal one).
The side from which the shot entered is the relevant random variable in
this case.  Note that the random variable may involve temporal evidence
(if $Y$'s shot had been the lethal one, the death would have occurred
%joe**
%a few seconds later), but it certainly does not have to be.
a few seconds later), but it certainly does not have to.
%%jp9 Joe, I think you should  tame your stress, because this is
%%precisely Lewis' argument for his counterfactual-depencence-
%%chain criterion, which we later criticize. Or, if you are
%%really enthusiastic about this, how about adding:
%joe**
%This is indeed the rationale for Lewis's \citeyear{Lewis86a} criterion of
This is indeed the rationale for Lewis's \citeyear{Lewis73a} criterion of
causation in terms of a counterfactual-dependence chain.
We shall see, however, that our definition goes beyond this
criterion.

It may be argued, of course, that by introducing the
intermediate variables $\SH$ and $\BH$
%joe8: added, because of break in flow by new paragraph
in Hall's story
we have also programmed the desired answer
into the problem; after all, it is the shattering of the bottle,
not $\SH$, which prevents $\BH$.
Pearl \citeyear[Section 10.3.5] {pearl:2k}
analyzes a similar late-preemption problem in
%joe+
%a dynamic structural equation models, where variables are
a dynamic structural equation model, where variables are
time indexed, and shows that the selection of
the first action as an actual cause of the effect
follows from conditions (similar to) AC1--AC3 even without
specifying the owner of the hitting ball.
%joe2: for abstract
%joe25
%A simplified adaptation of this analysis is presented below.
We now present a simplified adaptation of this analysis.

%joe2: cut lots of material for abstract
Let $t_1, t_2$, and $t_3$ stand, respectively, for the time that Suzy
threw her rock, the time that Billy  threw his rock, and
the time that the bottle was found shattered.
Let $H_i$ and $\BS_i$ be variables indicating
%joe2:  unnecessary
%, respectively,
whether the bottle is hit ($H_i$) and 
%joe**
was
shattered ($\BS_i$)
at time $t_i$ (where $i =1,2,3$ and $t_1 < t_2 < t_3$), with values 1
if hit
(respectively, shattered), 0 if not.
%joe6*: rewrote slightly
%$T_3$ is a dummy variable, representing an actor who throws the ball at time $t_3$; it is introduced here to emphasize the time-invariant structure of the
%equations and that the equations in (\ref{eqxxx}) are instantiations of
%the time-invariant equations:
Roughly speaking, if we let $T_i$ be a variable representing ``someone
%joe**
%throws the ball at time $t_i$ and take $\BS_0$ to be vacuously true
%(i.e., always 1),
%joe+
%throws the ball at time $t_i$ and take $\BS_0$ to be vacuously false
throws the ball at time $t_i$'' and take $\BS_0$ to be vacuously false
(i.e., always 0),
then we would expect the following time-invariant equations to hold for
all times $t_i$ (not just $t_1$, $t_2$, and $t_3$):
\begin{eqnarray*}
%joe19*: Judea, these equations aren't recursive!  Hi depends on BSi and
%BSi depends on Hi.  Fixed.
%H_i & = & T_i \wedge \neg \BS_i \\
H_i & = & T_i \wedge \neg \BS_{i-1} \\
\BS_i & = & \BS_{i-1} \vee H_i.
\label{eqxxy}
\end{eqnarray*}
That is, the bottle is hit at time $t_i$ if someone throws the ball at
time $t_i$ and the bottle wasn't already shattered at time $t_i$.
%jp8; adding "similarly"
Similarly, the
bottle is shattered at time $t_i$ either if it was already shattered at
time $t_{i-1}$ or it was hit at time $t_i$.

\begin{figure}[htb]
\input{psfig}
\hspace*{2.5in}\includegraphics{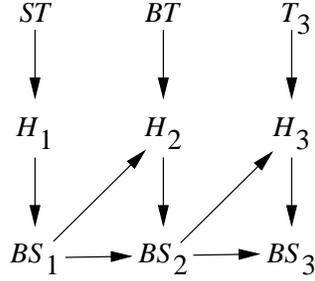}
\caption{Time-invariant rock throwing.}
\label{fig2}
\end{figure}

Since in this case we consider only times $t_1$, $t_2$, and $t_3$, we
get the following structural equations, where we have left in the
variable $T_3$ to bring out the essential invariance:
%joe6
%Assuming that $t_1<t_2<t_3$, and
%Neglecting the
%flight time of the rock, we can write down the structural
%equations that determine the state of the bottle at times
%$t_1, t_2$, and $t_3$.
\begin{eqnarray*}
H_1 & = & \ST \nonumber \\
\BS_1 & = & H_1 \nonumber \\
H_2 & = & \BT \wedge \neg \BS_1 \nonumber \\
\BS_2 & = & \BS_1 \vee H_2 \nonumber \\
H_3 & = & T_3 \wedge \neg \BS_2 \nonumber \\
\BS_3 & = & \BS_2 \vee H_3.
%\label{eqxxx}
\end{eqnarray*}
%joe2: the T_3 comes out of the blue in the equations above.  I think we
%should put some discussion of time-invariant equations below before
%we introduce the specific instance above.  I don't think we'll have
%room for this in the UAI paper (and certainly not in the abstract), so
%we don't have to worry about it now.
The diagram associated with this model is shown in Figure~\ref{fig2}.
In addition to these generic equations, the story also specifies that
the
%joe6:
%contextual conditions
context is such that
\[\ST=1, \BT=1, T_3=0.\]
%joe25
%The following causal network describes the situation.
The causal network in Figure~\ref{fig2} describes the situation.

It is not hard to show that
%joe6
%event $\ST=1$ passes the tests of
%AC2, with $\phi=(\BS_3=1)$ and $w': (\BT =0)$, while
$\ST=1$ is a cause of $\BS_3=1$ (taking $\vec{W} = \{\BT\}$ in AC2 and
$w'=0$).
$\BT=1$ is not a cause of $\BS_3=1$; it fails AC2(b) for every
partition $\vec{Z} \cup \vec{W}$.
To see this, note that to establish counterfactual dependence between
$\BS_3$ and $\BT$, we
%jp13 changing next two sentences to comply with new AC2
%must assign $\BS_2$ to $\vec{Z}$,
%contingency $\BS_1=0$ in $w'$.
%But this contingency results in $H_2=1$ which violates
%condition AC2(b), since an element of $\vec{Z}$ changes.
must assign $H_2$ to $\vec{Z}$,
%joe25
assign
$\BS_1$ to $\vec{W}$, and impose the
contingency $\BS_1=0$.
%joe12
%in $w'$.
But this contingency violates condition AC2(b), since it results
in $\BS_3 = 0$ when we restore $H_2$ to 0 (its
current value).

Two features are worth emphasizing in this example.
First, Suzy's throw is declared a cause of the outcome
event $\BS_3 = 1$ even though her throw did not hasten,
%joe6
%nor delay, nor change any property of that outcome. This can be made
delay, or change any property of that outcome. This can be made
clearer by considering another outcome event, $J_4$ = ``Joe was
unable to  drink his favorite chocolate cocktail from that bottle
on Tuesday night.'' Being a consequence of $\BS_3$, $J_4$ will
also be classified as having been caused by Suzy's throw,
not by Billy's, although $J_4$
%joe2
would have
occurred at precisely the
same time and in the same manner had Suzy not thrown the
ball.  This implies that hastening or delaying the outcome
cannot be taken as the basic principle for deciding actual
causation, a principle advocated by Paul \citeyear{paul:98}.

Second, Suzy's throw is declared a cause of $\BS_3 = 1$ even
though there is no counterfactual dependence chain between the two
(i.e., a chain $A_1 \rightarrow A_2 \rightarrow \ldots \rightarrow A_k$
where each event is counterfactually dependent on its predecessor).
The existence of such a chain  was proposed by
%joe**: changed reference
\newcomment
Lewis \citeyear{Lewis73a} as a necessary criterion for causation
in cases involving preemption.%
\footnote{Lewis \citeyear[Appendix D]{Lewis86a} later amended this
criterion to deal with problematic cases similar to that presented her.}
%joe8:
%Here,
In the actual context,
$\BS_2$ does not depend
(counterfactually) on either $\BS_1$ or on $H_2$;
%jp8: prefer verbal on symbolic explanations
%$\BS_2$ The will remain
%true even if $\BS_1$ turns false, as well as if $H_2$ turns true.
the bottle would be shattered at time $t_2$
even if it were unshattered at time $t_1$
%joe8: added
(since Billy's rock would have hit it),
as well as if
%it were hit (miraculously) at time $t_1$.
%jp13 changing t_1 to t_2:
it were hit (miraculously) at time $t_2$.
%joe**: this issue also pointed out by Lewis in the postscript to
%``Caussation'' 
%The importance of this departure from Lewis's account to
%one based on structural contingencies is
%further emphasized by Hitchcock \citeyear{hitchcock:99}.
\exam

%%jp5: joe: we should consider deleting example 4.3 for the uai paper  2/14}
%joe2: I agree; we won't have the space anyway.
%joe13: moved Example 4.3 to the end

\xam\label{xam4}  Can {\em not} performing an action be (part of) a cause?
Consider the following story, again taken from
%joe19
(an early version of) \cite{Hall98}:

\begin{quote}
%joe+
%Billy having stayed out in the cold too long throwing rocks, contracts a
Billy, having stayed out in the cold too long throwing rocks, contracts a
serious but nonfatal disease.  He is hospitalized
%joe19: modified, to match variables
%on Monday, but
%unfortunately the doctor forgets to administer the needed medication, so
%Billy is still sick on Tuesday.
and treated on Monday, so is fine Tuesday morning.
\end{quote}

%joe**
%Is the doctor's omission to treat Billy on Monday a cause of Billy's
But now suppose the doctor does not treat Billy on Monday.
Is the doctor's omission to treat Billy a cause of Billy's
being sick on Tuesday?  It seems that it should be, and indeed it is
according to our analysis.  Suppose that $\vec{u}$ is the context where,
among other things, Billy is sick on Monday and the situation is such
that the doctor forgets to administer the medication Monday.
(There is much more to the
context $\vec{u}$, as we shall shortly see.)  It seems reasonable that
the model should have two random variables:
\begin{itemize}
%jp6: changing TM to MT, for  consistency
\item $\MT$ for ``Monday treatment'', with values 0 (the doctor does
not treat Billy on Monday) and 1 (he does); and
\item $\BMC$ for ``Billy's medical condition'', with values 0 (recovered)
and 1 (still sick).
%and 2 (dead).  (We don't need the third value
%yet---nothing would change if we made $\BMC$ two-valued---but it will be
%useful for the variant of the story we are about to discuss.)
\end{itemize}
Sure enough, in the obvious causal model, $\MT=0$ is a cause
%joe6
%(and an explanation)
of $\BMC=1$.

This may seem somewhat disconcerting at first.  Suppose there are 100
doctors in the hospital.  Although only one of them was assigned to
%joe+
%Billy (and he forget to give medication), in principle, any of the other
Billy (and he forgot to give medication), in principle, any of the other
99 doctors could have given Billy his medication.  Is the fact that they
didn't give him the medication also part of the cause 
%joe+
%that he was still sick on Tuesday?
of him still being sick on Tuesday?

In the particular model that we have constructed, the other doctors'
failure to give Billy his medication is not a cause, since we have no
%joe+
%random variables to model the other doctor's actions, just as we had no
random variables to model the other doctors' actions, just as we had no
random variable in Example~\ref{xam1} to model the presence of oxygen.
Their lack of action is part of the context.  We factor it out because
(quite reasonably) we want to focus on the actions of Billy's doctor.
If we had included endogenous random variables corresponding to the
other doctors, then they too would be causes of Billy's
being sick on Tuesday.
%jp5 adds the following tw sentences (2/14)
%joe6: cut
%None of these will constitute an
%explanation though, since they stand in conjunctive
%relationship to the outcome (as in $M_2$, Example \ref{xam:arson});
%their overall conjunction is needed to make a genuine explanation
%of why Bill is still sick on Tuesday.

With this background, we
continue with Hall's modification of the original story.

\begin{quote}
Suppose that Monday's doctor is reliable, and administers the medicine
first thing in the morning, so that Billy is fully recovered by Tuesday
afternoon.  Tuesday's doctor is also reliable, and would have treated
Billy if Monday's doctor had failed to \ldots And let us add a twist:
one dose of medication is harmless, but two doses are lethal.
\end{quote}
Is the fact that Tuesday's doctor did {\em not\/}
treat Billy the cause of him being alive (and recovered) on
%joe25: made this change consistently
%Tuesday afternoon?
Wednesday morning?
%jp5: JOE, I SUGGEST TO GO STRAIGHT TO NEXT PARAGRAPH.
%joe2: OK
%That seems reasonable.  After all, if the Tuesday
%doctor had treated Billy, he would have died.
%How about the fact that Billy got the treatment on Monday?  Certainly
%that caused him {\em not\/} to get the treatment on Tuesday.  If that
%is the case, and the fact that he did not get the treatment on Tuesday
%caused him to be alive Wednesday morning, transitivity would suggest
%that
%Monday's treatment should be among the causes of Billy's being alive on
%Tuesday---despite the fact that his disease is non-fatal.
%[[Joe, it is not clear if this is Hall's position or yours, later on
%you knock down transitivity.]]

The causal model for this story is straightforward.
%joe19: as in UAI
There are three random variables: $\MT$ for Monday's treatment (1 if
Billy was treated Monday; 0 otherwise), $\TT$ for Tuesday's treatment (1
if Billy was treated Tuesday; 0 otherwise), and $\BMC$ for Billy's
medical condition
%joe21: minor changes
%(0 if Billy is alive
%Tuesday morning, still alive and well Wednesday morning;
(0 if Billy is fine both Tuesday morning and Wednesday morning;
1  if Billy is
sick Tuesday morning, fine Wednesday morning; 2 if Billy
%joe25
%is sick both Tuesday morning and afternoon; 3  if Bill is fine
%joe+
%is sick both Tuesday and Wednesday morning; 3  if Bill is fine
is sick both Tuesday and Wednesday morning; 3  if Billy is fine
Tuesday morning and dead Wednesday morning).
We can then describe Billy's condition as a function of the four
possible combinations of treatment/nontreatment on Monday and Tuesday.

\begin{figure}[htb]
\input{psfig}
%\hspace*{2.65in}\includegraphics{fig3.eps}}
\centerline{\includegraphics{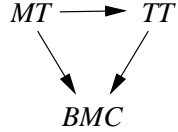}}
\caption{Billy's medical condition.}
\label{fig3}
\end{figure}

%joe: In the causal net
In the causal network
corresponding to this causal model, shown in
Figure~\ref{fig3}, there is an edge
from $\MT$ to $\TT$, since whether the Tuesday treatment occurs depends on
%joe5: fixed glitch
whether the Monday treatment occurs, and
%joe19
%there is an edge from both $\MT$
edges from both $\MT$
and $\TT$ to $\BMC$, since Billy's medical condition depends on both
treatments.

%joe2
%In this causal model, it is true that $\MT=1$ is an actual cause of
In this causal model, it is true that $\MT=1$ is a cause
%joe6
%(and an explanation)
of
$\BMC=0$, as we would expect---because Billy is treated Monday, he is not
treated on Tuesday morning, and thus recovers Wednesday morning.
%jp6: important point to add, now that my dynamic solution for
%Suzy's bottle is ommitted.
%joe4:
%\footnote{Note that Lewis' revised criterion of
%counterfactual-dependence-chain \cite{lewis:86} would fail in this
%joe8:  Again, updated Lewis ref
%\footnote{Lewis' \citeyear{lewis:86} revised criterion of
%joe**: cut, because, as Hitchcock points out, this is an artifact of
%our choice of variables.  It  wouldn't hold if we replaced BMC by separate
%variables for Tuesday's state of health and Wednesday's state of health.
%\footnote{Lewis's \citeyear{Lewis86a} revised criterion of
%counterfactual-dependence-chain also fails in this
%example; $\BMC$ does not depend on either $\MT$ or $\TT$
%in the context given.}
$\MT=1$ is also a cause
%joe6
%(and an explanation)
of $\TT=0$, as we would
expect, and $\TT=0$ is a cause
of Billy's being alive ($\BMC=0
\lor \BMC=1 \lor \BMC=2$).  However, $\MT=1$ is {\em not\/} a cause
of Billy's being alive.  It fails condition AC2(a): setting
$\MT=0$ still leads to Billy's being alive (with $W=\emptyset$).
%joe6: moved footnote into text
%\footnote{
%joe12: rewrote slightly
%We are not at liberty to choose $W=\{\TT\}$, for then
%AC2(b) would be violated under $\TT=1$
%%jp13 adding
%(with $Z' = 0).
Note that it would not help to take $\vec{W}= \{\TT\}$.  For if $\TT=0$,
then Billy is alive no matter what $\MT$ is, while if $\TT=1$, then Billy is
dead when $\MT$ has its original value, so AC2(b) is violated (with
$\vec{Z}' = \emptyset$).
%joe6
%Similarly, $\MT=1$ is not an explanation of Billy's being alive.

This shows that
%our definition of actual causality is not transitive.
causality is not transitive, according to our definitions.
Although $\MT=1$ is a cause of $\TT=0$ and $\TT=0$ is a
cause of $\BMC=0 \lor \BMC=1 \lor \BMC=2$, $\MT=1$ is not a cause
of $\BMC=0 \lor \BMC=1 \lor \BMC=2$.
%joe6
%and similarly for explanation.
%Nor are the notions closed
Nor is causality closed
under {\em right weakening}:  $\MT=1$ is a cause of $\BMC=0$,
which logically implies $\BMC=0 \lor \BMC=1 \lor \BMC=2$, which is not
caused by $\MT=1$.%
%joe8: added
\footnote{Lewis \citeyear{Lewis00} implicitly assumes
right weakening in his defense of transitivity.  For example, he says
``\ldots it is because of Black's move that Red's victory is caused one
way rather than another.  That means, I submit, that in each of these
cases, Black's move did indeed cause Red's victory.  Transitivity
succeeds.''  But there is a critical (and, to us, unjustifiable) leap in
this reasoning.  As we already saw in Example~\ref{xam1}, the fact that
April rains cause a fire in June does {\em not\/} mean that they cause
the fire.}
%The same goes for explanation.

Hall \citeyear{Hall01,Hall98}
discusses the issue of transitivity of causality, and suggests that
there is a tension between the desideratum that causality be transitive
and the desideratum that we allow causality due to the failure of some
event to occur.  He goes on to suggest that there are actually two
concepts of causation: one corresponding to counterfactual dependence
and the other corresponding to ``production'', whereby $A$ causes $B$ if
$A$ helped to produce $B$.
%joe10: As Chris pointed out, this was backwards.
%Causation by dependence is transitive; causation by production is not.
Causation by production is transitive; causation by dependence is not.

Our definition certainly has some features of both counterfactual
dependence and of production---AC2(a) captures some of the intuition of
counterfactual dependence (if $A$ hadn't happened then $B$ wouldn't have
%joe12
%happened under $W=w'$) and AC2(b) captures some of the
happened if $\vec{W}=\vec{w}'$) and AC2(b) captures some of the
features of production ($A$ forced $B$ to happen,
%jp11 $W=w=w'$ seems  unclear, changing to:
%again under $W=w$ and under $W=w'$).
%joe12: seems more consistent with the current definition
even if $\vec{W} = \vec{w}'$).
%joe2: Judea, there's a slight inconsistency here.
%Your earlier definition of sustenance required it to hold
%for all changes to $W$, not just w'
Nevertheless,
%joe4: cut for UAI
%the core of our definition lies in {\em sustenance\/}---%
%the ability of $A$ to protect $B$ from changing as $W$ is set to $w'$.
%Moreover,
we do not require two separate notions
%joe4:
to deal with these concerns.
%joe4
%our two criteria are expressed in the standard language of
%counterfactuals.

%jp11, adding for  background:
Moreover, whereas Hall attributes the failure of transitivity
to a distinction between presence and absence of events,
%joe12
%our definition shows that
according to our definition,
the requirement of
transitivity causes problems whether or not we allow causality due to
the failure of some event to occur.  It is easy enough to construct a
story whose causal model has precisely the same formal structure as that
above, except that $\TT=0$ now means that the treatment was given and
$\TT=1$ means it wasn't.  (Billy starts a course of treatment on Monday
which, if discontinued once started, is fatal \ldots)  Again, we don't
get transitivity, but now it is because an event occurred (the treatment
was given), not because it failed to occur.

%joe8: added paragraph
Lewis \citeyear{Lewis86a,Lewis00} insists that causality is transitive,
%joe10
%partly to deal with some difficulties in his definition.
partly to be able to deal with
%joe19
%what is called in the literature {\em
%joe12
%early preemption} \cite{Lewis86a}.  An example of  early preemption
preemption \cite{Lewis86a}.
%joe19
%An example of preemption
%(taken from \cite{hitchcock:99}) is a scenario where an
%assassin-in-training, who is an excellent shot, fires and kills the
%victim.  However, his supervisor (an equally skilled shot) is present on
%the mission in case the trainee loses his nerve.  We would like to call
%the trainee the cause of the victim's death, even though if the trainee
%hadn't shot, the victim would have died anyway.
As Hitchcock \citeyear{hitchcock:99} points out,
our account handles
%%joe12
%%early
%joe25: typo
%with
the standard examples of preemption
without needing to invoke
transitivity,
%we simply take $W$ to be such that the supervisor does not shoot.
%Of course, we
%could build transitivity into our definition of causality (by taking the
%transitive closure of our current definition).  However,
%%joe10
%%as our examples show,
%since our definition can deal with the difficulties
%encountered by Lewis's definition without
%invoking transitivity and
which,
as Lewis's own examples show,
leads to counterintuitive conclusions.
\exam

\xam\label{xam5} This example considers the problem of what Hall calls
{\em double
prevention}.  Again, the story is taken from Hall \citeyear{Hall98}:

\begin{quote}
Suzy and Billy have grown up, just in time to get involved in World War
III.  Suzy is piloting a bomber on a mission to blow up an enemy target,
and Billy is piloting a fighter as her lone escort.  Along comes an
%joe**: changed Lucifer to Enemy throughout, as in most recent version
%of Hall's paper 
%enemy fighter plane, piloted by Lucifer.  Sharp-eyed Billy spots
enemy fighter plane, piloted by Enemy.  Sharp-eyed Billy spots
Enemy, zooms in, pulls the trigger, and Enemy's plane goes down in
flames.  Suzy's mission is undisturbed, and the bombing takes place as
planned.
\end{quote}

Does Billy deserve part of the cause for the success of the
mission? After all, if he hadn't pulled the trigger, Enemy would have
eluded him and shot down Suzy.  Intuitively, it seems that the answer is
yes, and the obvious causal model gives us this.
Suppose we have the following random variables:
\begin{itemize}
\item $\BPT$ for ``Billy pulls trigger'', with values 0 (he doesn't) and
1 (he does);
\item $\LE$ for ``Enemy eludes Billy'', with values 0 (he doesn't) and
1 (he does);
\item $\LSS$ for ``Enemy shoots Suzy'', with values 0 (he doesn't) and
1 (he does);
\item $S\ST$ for ``Suzy shoots target'', with values 0 (she doesn't) and
1 (she does);
\item $\TD$ for ``target destroyed'', with values 0 (it isn't) and
1 (it is).
\end{itemize}
The causal network corresponding to this model is just
%joe**: try \arc instead?
%$$\BPT \longrightarrow \LE \longrightarrow \LSS \longrightarrow \SST
%\longrightarrow \TD.$$
$$\BPT \arc \LE \arc \LSS \arc \SST
\arc \TD.$$

%joe6*: do we really want this addition of SPS?  I now think it might be
%a distraction (although I don't feel strongly about it).  I suppose
%it's only a paragraph.
In this model, $\BPT=1$ is a cause
%and an explanation
of $\TD=1$.  
%joe**
Of course, $\SST=1$ is a cause of $\TD=1$ as well.
%This is all right,
%as far as it goes, but it seems to suggest that Suzy plays no role.
%This becomes particularly clear when we observe that 
It may be somewhat disconcerting to observe that 
$\BPT=1$ is also a
cause of $\SST=1$. 
%joe**
%Billy's pulling the trigger causes (or, perhaps
%%better, results in, Suzy shooting the target.
%better, enables) Suzy to shoot the target.
%\footnote{It is also true that $\SST=1$ is a cause of $\TD=1$, as, for
%that 
%matter, are $\LE=0$ and $\LSS=0$, but this does not affect the
%unreasonableness of $\BPT=1$ being a cause of $\SST=1$.}
%The problem with this causal model is that it makes
It seems strange to think of Billy being a cause of Suzy doing something
she was planning to do all along. 
Part of the problem is that, according to our definition (and 
all other definitions of causality that we are aware of), if $A$
enables $B$, then $A$ is a cause of $B$.  Arguably another
part of the problem with $\BPT=1$ being a cause of $\SST=1$ and $\TD=1$
is that it 
seems to leave Suzy out of the picture altogether.  
%joe**
%This may seem to make Suzy seem like an
%automaton.  We would use the same causal model to describe the
%situation 
%where Suzy's plane is actually an unmanned (unwomaned?) plane
%pre-programmed to shoot at the target if it is not shot down.  Under
%%those circumstances, it seems perfectly reasonable to view $\BPT=1$ as
%those circumstances, it may seem more reasonable to view $\BPT=1$ as
%a cause of both $\SST=1$ and $\TD=1$.  
We can bring Suzy more into the picture by having 
a random variable corresponding to Suzy's
plan or intention.  Suppose that we add a random variable $\SPS$ for
``Suzy plans
to shoot the target'', with values 0 (she doesn't) and 1 (she does).
Assuming that Suzy shoots if she plans to, we
then get the following causal network, where now $\SST$ depends on both
$\LSS$ and $\SPS$:

\begin{figure}[htb]
\input{psfig}
\hspace*{1.75in}\includegraphics{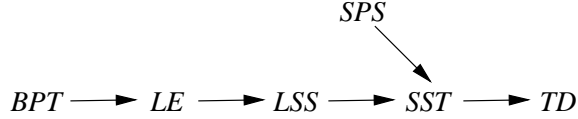}
\caption{Blowing up the target.}
\label{fig4}
\end{figure}

%%In this case, it is easy to check that $\BPT=1 \land \SPT=1$ is the cause
%%of $\TD=1$.
In this case, it is easy to check
%joe2: cut;
%(with $W=\emptyset$)
that each of
%joe10: corrected typo: SPT -> SPS
$\BPT=1 $ and $\SPS=1$ is a cause of $\TD=1$.
%joe2: added
%joe6:
%However, neither alone constitutes an explanation; their conjunction
%$\BPT=1 \land \SPT=1$ is an explanation of $\TD=1$.

\begin{figure}[htb]
\input{psfig}
\hspace*{1.75in}\includegraphics{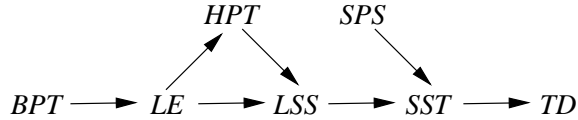}
\caption{Blowing up the target (refined version).}
\label{fig5}
\end{figure}
Hall suggests that further complications arise if we add a second
fighter plane escorting Suzy, piloted by Hillary.  Billy still shoots
down Enemy, but if he hadn't, Hillary would have.
%judea1 revises treatment of Hillary    9.15
The natural way of dealing with this is to add just one more variable
$\HPT$ representing Hillary's pulling the trigger iff $\LE=1$
(see Figure \ref{fig5}), but then,
using the naive counterfactual criterion, one might conclude
that the target will be destroyed $(\TD = 1)$ regardless of Billy's
action, and $\BPT=1$ would lose its
``actual cause'' status (of $\TD=1$).
Fortunately, our definition goes beyond this naive
criterion and classifies $\BPT=1$ as
a cause of $\TD=1$, as expected.
%joe**: again, an artefact of our choice of model.
%\footnote{Note that Lewis's revised criterion of
%counterfactual-dependence-chain \cite{Lewis86a} also fails in this
%example; $\LSS$ does not depend on either $\HPT$ or $\LE$
%in the context given.}
This can be seen by noting that the partition
%joe6: moved SPS into W
%$Z=\{\BPT, \LE, \LSS, \SST, \TD, \SPS\}, W =\{\HPT\}$
$\vec{Z}=\{\BPT, \LE, \LSS, \SST, \TD\}, \vec{W} =\{\HPT,\SPS\}$
satisfies conditions AC1--AC3
%joe6
%(with $\HPT=0$ for $w'$).
(with $w'$ such that $\HPT=0$ and $\SPS=1$).
The intuition rests, again, on structural contingencies;
although Billy's action seems superfluous under ideal
conditions, it becomes essential under a
%joe+
%contingency in which Hillary would fail her mission
contingency in which Hillary would fail in her mission
to shoot Enemy. This contingency is represented by setting
$\HPT$ to $0$ (in testing AC2(a)),
irrespective of $\LE$.
%%The naive way of dealing with this is to add just one more variable
%%$\HPT$ representing Hillary's pulling the trigger, but then we get that
%%each of $\BPT=1$ and $\HPT=1$ are the actual causes or $\TD=1$ (or $\BPT=1
%%\land \SPS=1$ and $\HPT=1 \land \SPS=1$, if we add $\SPS$ to the picture).
%%The problem, just as
%%in the rock throwing story in Example~\ref{xam2}, is that Billy and
%%Hillary play symmetric roles in this causal model.  The solution is
%%again to add another two random variables, $\BSH$ (Billy's shot hit
%%Enemy) and $HSH$ (Hillary's shot his Enemy) to make explicit the
%%asymmetry.  Since the details are just like those in Example~\ref{xam2},
%%we omit them here.  All the other variants of this story considered by
%%Hall are equally easy to deal with in our framework.
\exam

%joe25: moved this example to the next section
%\xam\label{shadow}
%Consider an example originally due to
%McDermott \citeyear{mcdermott95}, and also considered  by Lewis
%\citeyear{Lewis00} and Hitchcock \citeyear{hitchcock:99}.
%A ball is caught by a fielder...

%joe19*: Added
\section{A More Refined Definition}\label{sec:refined}
We labeled our definition ``preliminary'', suggesting that there are
some situations it cannot deal with.  The following example illustrates
the problem.

\xam
%joe13: this is Example 4.3, rewritten
\label{xam3}
Consider Example~\ref{xam2}, where both Suzy and Billy throw a rock at a
bottle, but Suzy's hits first.  Now suppose that there is a noise which
%joe+
%causes Suzy to delay her throw slightly, but still before Billy's.
causes Suzy to delay her throw slightly, but that she still throws
before Billy. 
Suppose that we model this situation using the approach described in
%joe**: add three variables (H_{1.5} too)
\newcomment
%Figure~\ref{fig2}, adding two extra variables, $N$ (where $N=0$ if
%there is no noise and $N=1$ if there is a noise) and $\BS_{1.5}$ (where
%$BS_{1.5} = 1$ if the bottle is shattered at time $t_{1.5}$, where $t_1
%< t_{1.5} < t_2$, and $\BS_{1.5} = 0$ otherwise).  In the actual
Figure~\ref{fig2}, adding three extra variables, $N$ (where $N=0$ if there
is no noise and $N=1$ if there is a noise), $H_{1.5}$ (which is 1 if the
bottle is hit at time $t_{1.5}$, where $t_1 < t_{1.5} < t_2$, and 0
otherwise) and $\BS_{1.5}$ (which is 1 if the bottle is  shattered at
time $t_{1.5}$ and 0 otherwise).  In the actual
situation, there is a noise and the bottle shatters at $t_{1.5}$, so
%joe**
%$N=1$ and $\BS_{1.5} = 1$.  Just as in Example~\ref{xam2}, we can show
$N=1$, $H_{1.5} = 1$, and $\BS_{1.5} = 1$.  Just as in
Example~\ref{xam2}, we can show 
that Suzy's throw is a cause of the bottle shattering and Billy's throw
is not.  Not surprisingly, $N=1$ is a cause of $\BS_{1.5} = 1$ (without
the noise, the bottle would have shattered at time 1).  Somewhat
disconcertingly though, $N=1$ is also a cause of the bottle shattering.
That is, $N=1$ is a cause of $\BS_3 = 1$.

This seems unreasonable.
Intuitively, the bottle would have shattered whether or not there had
been a noise.  However, this intuition is actually not correct in our
causal model.  Consider the contingency where Suzy's throw
hits the bottle.  If $N=1$ and $\BS_1 = 0$, then the bottle does
%joe**
%shatter at time $1.5$.  Given this, it easily follows that, according to
shatter at $t_{1.5}$.  Given this, it easily follows that, according to
our definition, $N=1$ is a cause of $\BS_3 = 1$.%
%joe29: moved footnote (again) from end of previous paragraph, and ended
%example here
%joe16: moved footnote again
\footnote{We thank Chris Hitchcock for bringing this example to our
attention.}
\exam

%joe19*: rewrote and got rid of ranking functions.
The problem here is caused by what might be considered an extremely
unreasonable scenario:  If $N=1$ and $\BS_1 = 0$, the bottle does not
shatter despite being hit by 
Suzy's rock.  Do we want to consider such scenarios?  That is up to
the modeler.  Intuitively, if we allow such scenarios, then the noise
ought to be a cause; if not, then it shouldn't.  

%joe29: expanded; I think it's worth taking this a bit more slowly
%Capturing this intuition in the formal framework is straightforward.  
It is easy to modify our preliminary definition so as to be able to
capture this intuition.  We take an {\em extended causal model\/} to now
be a tuple $(\S,\F,\E)$, where $(\S,\F)$ is a causal model, and $\E$ is
a set of {\em allowable settings\/} for the endogenous variables.  That is, if
the endogenous variables are $X_1, \ldots, X_n$, then $(x_1, \ldots,
x_n) \in \E$ if $X_1 = x_1, \ldots, X_n = x_n$ is an allowable setting.
%We simply have a set of allowable
%settings for the exogenous variables; all the setting considered in
%AC2(a) and (b) must be in the set of allowable settings.
%For this example, if we disallow settings where $\BS_1 = 0 \land H_1 =
%1$, we are back to the original setting, and the noise is not a cause.%
We say that a setting of a subset of the endogenous variables is
allowable if it can be extended to a setting in $\E$.
We then slightly modify clauses 
AC2(a) and (b) in the definition of causality to restrict to allowable
settings.  In the special case where $\E$ consists of all settings, this
definition reduces to the definition we gave in Section~\ref{sec:actcaus}.
We can deal with Example~\ref{xam3} in extended causal models
by disallowing settings where $\BS_1 = 0 \land H_1 = 1$.  This
essentially puts us back in the original setting.
The following example further illustrates the need to be able to deal
with ``unreasonable'' settings.
\xam
\label{Larry}
 Fred has his finger severed by a machine at the
factory ($\FS = 1$).  Fortunately, Fred is covered by a health plan.
He is rushed to the hospital, where his finger is sewn back on.
A month later, the finger is fully functional ($\FF=1$).  In this
story, we would not want to say that $\FS=1$ is a cause of $\FF=1$ and,
indeed, according to our definition, it is not, since $\FF=1$ whether or
not $\FS=1$
%jp13 adding
(in all contingencies satisfying AC2(b)).

However, suppose we introduce a new element to the story, representing a
nonactual structural contingency:  Larry the Loanshark may be waiting
outside the factory with the intention of cutting off Fred's
finger, as a warning to him to repay his loan quickly.  Let $\LL$
represent whether or not Larry is waiting and let $\LC$ represent whether
%joe+
%Larry cuts of the Fred's finger.  If Larry cuts off Fred's finger, he
Larry cuts off Fred's finger.  If Larry cuts off Fred's finger, he
will throw it away, so Fred will not be able to get it sewn back on.
In the actual situation,
$\LL=\LC=0$;  Larry is not waiting and Larry does not cut off Fred's
%jp11 changing
%finger.  However, there seems to be no harm in adding this fanciful
%into
finger.  So, intuitively, there seems to be no harm in adding this fanciful
element to the story.  Or is there?  Suppose that, if Fred's finger is
cut off in the factory, then Larry will not be able to cut off the
finger himself (since Fred will be rushed off to the hospital).  Now
$\FS=1$ becomes a cause of $\FF=1$.  For in the structural contingency
where $\LL=1$, if $\FS=0$ then $\FF=0$ (Larry will cut off Fred's
finger and throw it away, so it will not become functional again).
Moreover, if $\FS=1$, then $\LC=0$ and $\FF=1$, just as in the actual
situation.%
%joe16: moved footnote
\footnote{We thank Eric Hiddleston for bringing
%this issue and
this example to our attention. 
%joe**
The example is actually a variant of one originally due to Kvart
\citeyear{Kvart91}, although Kvart's example did not include Larry the
%joe+
%Loandshark  
Loanshark  
and was intended to show a violation of transitivity.}

%joe19*:
If we really want to view Larry's cutting off Fred's finger as totally
fanciful, then we simply disallow all settings where $\LL = 1$.  On the
%joe25
%other hand, if we want to take Larry seriously, then it seems more
%reasonable to take seriously the possibility that Larry might be out
%there waiting for Fred,
other hand, if having fingers cut off in a way that they cannot be put
on again is rather commonplace,
then it seems more reasonable to view the
accident as a cause of Fred's finger being functional a month after the
accident.
\exam

%joe29
In extended models, it is also straightforward to deal with problems
of causation by omission.
%joe**: kill next line
%by 
\newcomment
\xam
Hall and Paul \citeyear{HallP03} give an example due to Sarah McGrath 
suggesting that there may be a difference between causation by omission
and causation by commission:
\begin{quote}
Suppose Suzy goes away on vacation,
leaving her favorite plant in the hands of Billy, who has promised to
water it.  Billy fails to do so.  The plant dies---but would not have,
had Billy watered it. \ldots Billy's failure to water the plant caused
its death.  But Vladimir Putin also failed to water Suzy's plant.  And,
%joe+
%had he done so, it would not have died.  Why do we there also not count
had he done so, it would not have died.  Why do we also not count
his omission as a cause of the plant's death?
\end{quote}
Billy is clearly a cause in the obvious structural model.  So is Vladimir
Putin, if we do not disallow any settings and include Putin watering the
plant as one of the endogenous variables.  However,
if we simply disallow the setting where Vladimir Putin waters the plant.
then Billy's failure to water the plants is a cause, and Putin's failure
%joe+
%is not.  We could equally well get this result by not take Putin's
is not.  We could equally well get this result by not taking Putin's
watering the plant as one of the endogenous variables in the model.
(Indeed, we suspect that most people modeling the problem would not
include this as a random variable.)

Are we giving ourselves too much flexibility here?  We believe not.
%joe**
%it is up to a modeler to defend her choice of model.  A model which does
It is up to a modeler to defend her choice of model.  A model which does
not allow us to consider Putin watering the plant can be defended in the
obvious way: that is a scenario too ridiculous to consider.  On the
other hand, if Suzy's sister Maggie (who has a key to the house) also
came by to check up on things, then it does not seem so unreasonable for
Suzy to get slightly annoyed at Maggie for not watering the plant, even
if she was not supposed to be the one responsible for it.  Intuitively,
it seems reasonable not to disallow the setting where Maggie waters the
plant. 
\exam

%joe19*
%joe29
%Allowing settings to be excluded plays a more significant role in our
Considering only allowable settings plays a more significant role in our
framework than just that of allowing us to ignore fanciful scenarios.
As the following example shows, it helps clarify the relationship
between various models of a story.

\xam\label{xam6}  This example concerns what Hall calls the
distinction between causation and determination.  Again, we quote Hall
%joe**
%\citeyear{Hall98}:
\citeyear{Hall01}:

\begin{quote}
%joe**: rewrote example, as in Journal of Philosophy version
%You are standing at a switch in the railroad tracks.  Here comes the
%train: If you flip the switch, you'll send the train down the left-hand
%track; if you leave it where it is, the train will follow the right-hand
%track.  Either way, the train will arrive at the same point, since the
%tracks reconverge up ahead.  Your action is not among the causes of this
%arrival; it merely helps to determine how the arrival is brought about
%(namely, {\em via} the left-hand track, or {\em via} the right-hand
%track).
The engineer is standing by a switch in the railroad tracks.  A train
approaches in the distance.  She flips the switch, so that the train
travels down the right-hand track, instead of the left.  Since the
tracks reconverge up ahead, the train arrives at its destination all
the same \ldots
\end{quote}

Again, our causal model gets this right.  Suppose we have three random
variables:
\begin{itemize}
%joe+
%\item $F$ for ``flip'', with values 0 (you don't flip the switch) and 1
%(you do);
\item $F$ for ``flip'', with values 0 (the engineer doesn't flip the
switch) and 1 
(she does);
\item $T$ for ``track'', with values 0 (the train goes on the left-hand
track) and 1 (it goes on the right-hand track); and
\item $A$ for ``arrival'', with values 0 (the train does not arrive at
the point of reconvergence) and 1 (it does).
\end{itemize}
Now it is easy to see that flipping the switch ($F=1$)
%joe25
%does cause
causes the
train to go down the left-hand track ($T=0$), but does not cause it to
arrive ($A=1$), thanks to AC2(a)---whether or not the switch is flipped,
the train arrives.
%joe1: made your extension part of the example
%\exam

\begin{figure}[htb]
\input{psfig}
%\hspace*{2.5in}\includegraphics{fig6.eps}
\centerline{\includegraphics{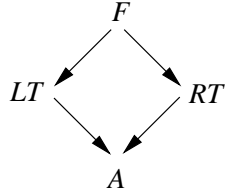}}
\caption{Flipping the switch.}
\label{fig6}
\end{figure}

%judea1 add an extension to switching causation  9.15
However, our proposal goes one step beyond
this simple picture. Suppose that we
%joe16
%decide to
model
the tracks using {\em two} variables:
\begin{itemize}
%joe1: typo (corrected below too)
%\item $\LT$ for ``left-rack'', with values 1 (the train goes on the
\item $\LT$ for ``left-track'', with values 1 (the train goes on the
left-hand track) and 0 (it does not go on the left-hand track); and
\item $\RT$ for ``right-track'', with values 1 (the train goes on the
right-hand track) and 0 (it does not go on the right-hand track).
\end{itemize}
The resulting causal diagram is shown in Figure \ref{fig6};
%joe2: cut for abstract
it is
isomorphic to a class of problems 
%joe+
that
Pearl \citeyear{pearl:2k} calls
``switching causation''.
%joe19*: cut; we now have better ways of dealing with this.
%\footnote{Pearl \citeyear{pearl:2k} used a switch between two light sources,
%so as to avoid questions of $\RT$ and $\LT$ influencing each other.}
%In
%our story, we might as well imagine that more than one train
%%joe1
%%may arrive, so both $RT$ and $LT$ may be true.}
%may arrive, so that $\RT$ and $\LT$ can simultaneously be true.}
%joe19*:
It seems reasonable to disallow settings where $\RT = \LT = 1$; a train
cannot go down more than one track.  If we do not disallow any other
settings, then,
lo and behold, this representation classifies $F=1$ as a
%joe+
%cause of $A$.  At first sight, may seem
cause of $A$.  At first sight, this may seem
counterintuitive:
Can a change in representation turn a non-cause into a cause?

It can and it should!
%joe1
%We shall next argue that
The change to a two-variable model is not merely syntactic, but
represents a profound change in the story.
The two-variable model depicts the tracks as two independent
mechanisms, thus allowing one track to be set (by action or
mishap) to false (or true) without affecting the other.
Specifically, this permits the disastrous mishap of
flipping the switch while the left track is malfunctioning.
%joe19*: added
%joe**: to be consistent with the rewritten story
%More formally, it allows a setting where $F= 1$ and $\LT = 0$.
More formally, it allows a setting where $F= 1$ and $\RT = 0$.
Such abnormal
%joe19: for consistency
%eventualities
settings
are imaginable and expressible in the
two-variable model, but not in the one-variable model.
%joe19*:
%joe**
%Of course, if we disallow settings where $F = 1$ and $\LT = 0$, or where
%$F = 0$ and $\RT = 0$, then we are essentially back at the earlier model.
Of course, if we disallow settings where $F = 1$ and $\RT = 0$, or where
$F = 0$ and $\LT = 0$, then we are essentially back at the earlier model.
The potential for such
%joe19
%eventualities
settings
is precisely
%joe+
%what renders $F=1$ a cause of the $A$ in the model
what renders $F=1$ a cause of $A$ in the model
of Figure \ref{fig6}.%
\footnote{This can be seen by noting that condition
AC2 is satisfied by the partition $\vec{Z}=\{F, \LT, A\},
%joe**
%\vec{W}=\{\RT\}$, and choosing $w'$ as the setting $\RT=0$. The event
%$\RT=0$ conflicts with $F=0$ under normal situations.}
\vec{W}=\{\LT\}$, and choosing $w'$ as the setting $\LT=0$.}

%joe6*: rewrote use "cause" instead of "explanation".
Is flipping the switch a
%joe6
%valid explanation of the train arrival?
legitimate cause of the train's arrival?
Not in ideal situations, where all mechanisms work
as specified. But this is not what
%joe6
%causal explanations
causality
(and causal modeling) are all about.
%joe6*
%Causal explanations earn their
Causal models earn their
value in abnormal circumstances, created by structural
contingencies, such as the possibility of a malfunctioning
track. It is this possibility that
should enter our mind whenever we decide to designate each track
as a separate mechanism (i.e., equation) in the model
and, keeping this contingency in mind, it should not be too
odd to name the switch position a cause of the train
arrival (or non-arrival).
\exam

%joe19%
Example~\ref{xam6} gives some insight into the process of model construction.
While  there is no way of proving that a given model is the ``right''
model, it is clearly important for a model to have enough random
variables to express what the modeler considers to be all reasonable
situations.  On the other hand, by allowing for the possibility of
restricting the set of possible settings in the definition of causality,
we do not penalize the modeler for inadvertently having too many possible
settings.

%joe19*: added

\xam\label{major}
%joe**: reordered Morgana-Merlin and Sergeant-Major
\newcomment
%The next pair of examples are discussed in Lewis's recent paper
%\citeyear{Lewis00}; they are examples of what Lewis calls ``trumping''.
The next pair of examples were introduced by
Schaffer~\citeyear{Schaffer01} under the name {\em trumping preemption}.
To quote Schaffer:
\begin{quote}
%joe**: to match Schaffer
%Suppose that the laws of magic state that what will happen at midnight
%must match the first spell cast on the previous day. The first spell of the
%day, as it happens, is Merlin's prince-to frog spell in the morning.
%Morgana casts another prince-to-frog spell in the evening.  At midnight
%the prince turns into a frog.  Either spell would have done the  job,
%had it been the only spell of the day; but Merlin's spell was first, so
%it was his spell that caused the transmogrification.  Merlin's spell
%trumped Morgana's.
Imagine that it is a law of magic that the first spell cast on a given
day match the enchantment that midnight.  Suppose that at noon Merlin
casts a spell (the first that day) to turn the prince into a frog, that
at 6:00 PM Morgana casts a spell (the only other that day) to turn the
prince into a frog, and that at midnight the prince becomes a frog.
\end{quote}
%joe**
\newcomment
Clearly Merlin is a cause of the enchantment.  What about Morgana?
There is an intuition that Merlin should be the only cause, since his
spell ``trumps'' Morgana's.  Can this be captured in a causal model?

A coarse-grained  model for this story has three variables:
\begin{itemize}
\item $\Mer$, with values $0$ (Merlin did not cast a spell), $1$ (Merlin
cast a prince-to-frog spell in the morning), and 2 (Merlin cast a
prince-to-frog spell in the evening);%
\footnote{The variable could take on more values, allowing for other
spells that Merlin could cast and other times he could cast them, but
this would not affect the analysis.}
\item $\Mor$, with values 0, 1, 2, with interpretations similar to those
%joe+
for $\Mer$;
\item $F$, the outcome, with values 0 (prince) or 1 (frog).
\end{itemize}
In this model, with the obvious structural equations, both Merlin's
spell and Morgana's spell are the causes of the transmogrification.
(We do need to specify what happens if both Merlin and Morgana cast a
spell at the same time.  The choice does not affect the analysis.)
%joe26: jp0 change
%joe**
%This result clashes with the story's intuition, where Merlin's
%spell is deemed to be the cause, because our coarse-grained
%model fails to capture the inner workings of the laws of magic.
The problem, of course, is that the model does not capture how Merlin's
spell trumps Morgana's; Merlin and Morgana are being treated completely
symmetrically.  
In particular, the model fails to represent the temporal precedence
%joe**
%requirement ``must match the first spell cast''.
requirement that ``the first spell on a given day match the enchantment
that midnight''.

%joe25
%To get a model where Merlin is a cause, we can use a model similar in
%joe**
%To make Merlin's spell a cause, we can use a model similar in
%spirit to that used in the rock-throwing example.  We need to addition
To prevent Morgana's spell from being a cause, we can use a model similar in
spirit to that used in the rock-throwing example.  We need two additional
variables, $\MerE$ (for Merlin's spell effective) and $\MorE$ (for
Morgana's spell effective).  The picture is very similar to
Figure~\ref{fig1}, with $\MerE$ and $\MorE$ replacing $\SH$ and $\BH$:
\begin{figure}[htb]
\input{psfig}
%\begin{verbatim}
%
%  Mer     Mor
%   |       |
%  MerE -- MorE
%    \      /
%      \  /
%        O
%
%\end{verbatim}
\centerline{\includegraphics{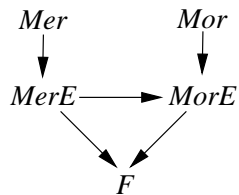}}
\caption{Merlin and Morgana.}\label{figMerlin}
\end{figure}

\noindent
%joe**
(Again, we are not specifying what happens if Merlin and Morgana throw
at the same time, because it is irrelevant to the analysis.)
In this model Morgana's spell is not a cause; it fails AC2(b).
%joe26: jp0 suggestion, but I couldn't figure out what you were saying here.
%What setting have we disqualified?  The model is completely symmetric.
%because it disqualifies the symmetric contingency
%where Morgana is the only one to cast a spell.
%joe**: I disagree with this now.  I think the right way to think about
%AC2(b) is that it's insisting that things happen the way they actually
%did, without worry about how reasonable that is.  For example, Billy's
%throw in the the cause because in fact his rock did not hit the
%bottle.  It's irrelevant that it would have hit the bottle had Suzy not
%thrown; the conditions considered in AC2(b) aren't necessarily
%reasonable (and dont have to be).
%but only
%because we allow a setting where Morgana is the only one to cast a
%spell, but Morgana's spell is not effective.
%joe26: jp0 change
%Again, it is up to the modeler to decide if this is reasonable.%
%joe**
%Again, it is up to the modeler to ensure that the structural
This again emphasizes the point that causality is relative to a model.
It is up to the modeler to ensure that the structural 
equations properly represent the dynamics in the story.
%joe26: didn't make the change in footnote, because I couldn't
%understand previous change.
%joe**: cut this, in line with my previous comment (perhaps that's what
%your comment was that I didn't understand).
%\footnote{Note that the analogous setting in the rock-throwing example
%would be where Billy throws but does not hit the bottle.  While this
%seems
%%joe25
%like
%a reasonable eventuality to consider, note that if we assume that
%Billy never misses, and so decide to exclude this eventuality, then
%Billy's throw is back to being a cause in the model in Figure~\ref{fig1}.}

%joe**
\newcomment
%The first is actually due to Bas van Fraassen.
The second example of trumping preemption is actually due to Bas van
Fraassen. Quoting Schaffer again:
\begin{quote}
%joe**: using Schaffer's version
%The Sergeant and the Major are shouting orders at the
%soldiers.   The soldiers know that in the case of conflict, they must
%obey the superior officer.  But, as it happens, there is no conflict.
%Sergeant and Major simultaneously shout ``Advance!''; the solders hear
%them both; the soldiers advance.  Their advancing is redundantly caused:
%if the Sergeant had shouted ``Advance!'' and the Major had been silent,
%or if the Major had shouted ``Advance!'' and the Sergeant had been
%silent, the soldiers would still have advanced.  But the redundancy is
%asymmetrical: since the soldiers obey the superior officer, they advance
%because the Major orders them to, not because the Sergeant does.  The
%major preempts the Sergeant in causing them to advance.  The major {\em
%trumps\/} the sergeant.
Imagine that \ldots the major and the sergeant stand before the
corporal, both shout ``Charge!'' at the same time, and the corporal
decides to charge.  
\end{quote}
Schaffer (and Lewis~\citeyear{Lewis00}) claim that, because orders from
higher-ranking soldiers trump those of lower-ranking soldiers, this is
again a case of trumping preemption:~the major is a cause of the charge;
the sergeant is not.

\begin{figure}[htb]
\input{psfig}
%\begin{verbatim}
%
%   M    S
%    \  /
%     A
%
%\end{verbatim}
\centerline{\includegraphics{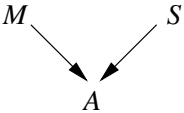}}
\caption{A simple model of the sergeant and the major}
\label{sergeant1}
\end{figure}

Our intuition does not completely agree with that of Schaffer and Lewis
in this example. 
In what seems to us the most obvious model of the story, both the
sergeant's order and the major's order are the causes of the advance.
Consider the model described in Figure~\ref{sergeant1}.
Assume for definiteness that the sergeant and the major can
each either order
an advance, order a retreat, or do nothing.  Thus, $M$ and $S$ can each
take three values, 1, $-1$, or 0, depending on what they do.  $A$
describes what the solders do; as the story suggests, $A = M$ if $M \ne
0$; otherwise $A = S$.
In the actual context, $M = S = A = 1$.
In this model, it is easy to see that both $M = 1$ and $S= 1$ are causes
of $A=1$, although $M = 1$ is a strong cause of $A=1$, while $S=1$ is
not.

\begin{figure}[htb]
\input{psfig}
%\begin{verbatim}
%   M    S
%   |\   |
%   |  \ |
%   |   SE
%    \ /
%     A
%\end{verbatim}
\centerline{\includegraphics{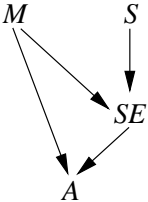}}
\caption{A model of the sergeant and the major that captures trumping.}
\label{sergeant2}
\end{figure}

%joe**
%It is possible to get a model of the story that perhaps comes closer to
%Lewis's intuitions by  explicitly capturing the fact that, if the
%Major actually issues an order, then the Sergeant is ignored.  
Of course, it is possible to get a model of the story that arguably
captures trumping by modeling the fact that if the major actually issues an
order, then the sergeant is ignored.   
To do this, we add a new variable $\SE$ that captures the sergeant's
``effective'' order.  If the major does not issue any orders (i.e., if
$M = 0$), then $\SE = S$.  If the major does issue an order,
then $\SE = 0$; the sergeant's order is effectively blocked.  In this model,
illustrated in Figure~\ref{sergeant2}, $A = M$ if $M \ne 0$; otherwise,
$A = \SE$.

In this model, the major does cause the corporal to advance, but the
sergeant does not.  For suppose we want to argue that $S=1$ causes $A =
1$.  The obvious thing to do is to take $\vec{W} = \{M\}$ and $\vec{Z} =
\{S, \SE, A\}$.  However, this choice does not satisfy AC2(b), since if
%joe25: more typos
%$M = 0$, $\SE = 0$ (it's original value), and $\SE = 1$, then $A = 0$,
$M = 0$, $\SE = 0$ (its original value), and $S = 1$, then $A = 0$,
not 1.  We leave it to the reader to check that it does not help to put
$\SE$ into $\vec{W}$.
The key point is that this
%joe25
more refined
model allows a setting where $M = 0$, $S =
1$, and $A = 0$ (because $\SE = 0$).  That is, despite the sergeant
issuing an order to attack and the major being silent, the corporal does
nothing (intuitively, because of some perceived ``interference'' from the
major, despite the major being silent).  
%joe**: cut this
%It seems that, to capture
%Lewis's intuition, we need to allow such settings (at least, in our
%framework). 

%joe**
\newcomment
Schaffer \citeyear[p.~175--176]{Schaffer01} seems to want to disallow
this model, or at least to allow other mechanisms for trumping.  
There may well be other mechanisms for trumping.  We believe that if
they are spelled out carefully, it should also be possible to capture
them using an appropriate causal model.  However, we cannot speak about
trumping preemption in our framework without being explicit as to how
the trumping takes place.%
\footnote{Of course, we could extend the framework to allow epistemic
%joe+
%considerations, using a standard possible worlds framework, where the
considerations, using a standard possible-worlds framework, where the
``worlds'' are causal models.  An agent could then be uncertain about how the
trumping takes place, while still knowing that the major is the cause of
charge, not the sergeant.  Nevertheless, in each of the possible worlds,
the trumping mechanism would still have to be specified.}

%joe26: insert-3.  I like this insert!
It is important to note that the diversity of answers
%joe**
%in this example does not reflect undisciplined freedom to
in these examples does not reflect undisciplined freedom to
tinker with the model so as to get the desired answer. Quite
the contrary; it reflects
an ambiguity in the original specification of the story,
which our definition helps disambiguate.
%joe**
%Each of the two models considered reflects a legitimate
Each of the models considered reflects a legitimate
interpretation of the story in terms of a distinct
%joe**
%model of the soldiers' attention-focusing strategy.
%For example, Figure~\ref{sergeant1} describes the soldiers' strategy as
%joe+: next line was accidentally erased.
model of the corporal's attention-focusing strategy.
For example, Figure~\ref{sergeant1} describes the corporal's strategy as
a single 
input-output
mechanism, with no intermediate steps. Figure~\ref{sergeant2} refines that
model into a two-step process
%joe**
%where soldiers first determine whether the major is silent
where the corporal first determines whether the major is silent
or speaking and, in the latter case, follow the major's
command. Naturally, the major should be deemed the cause of
advancing (in our scenario) given this strategy.
We can also imagine a completely different strategy
where the sergeant, not the major, will be deemed the cause
%joe**
%of advancing. If soldiers first determine whether or not
of advancing. If the corporal first determines whether or not
there is conflict between the two commanders and then, in
case of no conflict, pays full attention to the sergeant
(perhaps because his dialect is clearer, or his posture less
%joe+
%intimidating) it would make perfect sense then to say that the sergeant
intimidating), it would make perfect sense then to say that the sergeant
was the cause of advancing. Structural-equation models
provide a language for formally representing
these fine but important distinctions, and our definition
translates these distinctions into different
classifications of actual causes.
\exam

%joe25: moved example here; did some slight rewriting
\xam\label{shadow}
Consider an example originally due to
McDermott \citeyear{mcdermott95}, and also considered  by 
%joe**: added
Collins \citeyear{Collins00},
Lewis \citeyear{Lewis00}, and Hitchcock \citeyear{hitchcock:99}.
A ball is caught by a fielder.  A little further along its path there is
a solid wall and, beyond that, a window.  Does the fielder's catch cause
%joe+
%the window to remain unbroken?  As Lewis \citeyear{Lewis00} says
the window to remain unbroken?  As Lewis \citeyear{Lewis00} says,
\begin{quote}
We are ambivalent.  We can think: Yes---the fielder and the wall between
them prevented the window from being broken, but the wall had nothing to
do with it, since the ball never reached the wall; so it must have been
the fielder.  Or instead we can think:  No---the wall kept the window
safe regardless of what the fielder did or didn't do.
\end{quote}

Lewis argues that our ambivalence in this case ought to be respected,
and both solutions should be allowed.  We can give this ambivalence
formal expression in our framework.  If we make both the wall and the
%joe+
%fielder endogenous variables then,
fielder endogenous variables, then
%joe25: rewrote
%joe**
%then the wall is a cause of window being safe, under the
the fielder's catch is a cause of the window being safe, under the
assumption that the fielder not catching the ball and the wall not being
there is considered a reasonable scenario.  
%joe**: added
\newcomment
Note that if we also have a variable for whether the ball hit the wall,
then the presence of the wall is not a cause for the window's being safe
in this case; the analysis is essentially the same as that of the
Suzy-Billy rock-throwing example in Figure~\ref{fig1}.%
\footnote{We thank Chris Hitchcock for making this point.}
%joe**: cut
%Under these circumstances,
%the fielder's catch is also a cause of the window being safe.
%either one can be the cause.
%Note, however, by making them
%%joe10
%%part of the context,
%endogenous,
%%nwe allow for the
%possibility both that the fielder did not catch the ball (which seems
%reasonable) and that the wall miraculously did not stop the ball if the
%fielder did not catch it.
%This seems perhaps less reasonable, but  not totally surprising.
%Our framework has the resources to show how seriously this possibility
%should be taken, but letting the modeler decide whether to make the
%wall 
%endogenous and take
%By making the wall endogenous, we automatically proclaim the
%process of erecting (or maintaining) the wall subject to
%possible contingencies, since such contingencies are intrinsic
%to the semantics of endogenous variables.
On the other hand, if we take it for granted the wall's presence (either
by making the wall an exogenous variable, not including it in the model,
or not allowing situations where it doesn't block the ball if the
fielder doesn't catch it),  then the fielder's catch is not a cause
of the window being safe.
It would remain safe no matter what the fielder did, in any structural
contingency.

This example again stresses the importance of the choice of model, and
thinking through what we want to vary and what we want to keep fixed.
(Much the same point is made by Hitchcock \citeyear{hitchcock:99}.)
\exam

%joe25: added discussion of Yablo
%joe+
%This is perhaps a good place to compare our approach to that of Yablo
This is perhaps a good place to compare our approach with that of Yablo
\citeyear{Yablo02}.  The approaches have some surface similarities.
They both refine the standard notion of counterfactual dependence.  We
consider counterfactual dependence under some (possibly counterfactual)
contingency.  Yablo considers counterfactual dependence under the
%joe26: jp0 change
%assumption that some feature of the actual world remains fixed.
assumption that some feature of (or events in) the actual world remains
fixed.
The problem is, as Yablo himself shows, that for any $\vec{X} = \vec{x}$
and $\phi$ that actually happens, we can find some feature of the world
that we can hold fixed such that $\phi$ depends on $\vec{X} = \vec{x}$.
Take $\psi$ to be the formula $\vec{X} = \vec{x} \dimp \phi$.  If
$\vec{X} = \vec{x}$ and $\phi$ are both true in the actual situation,
then so is $\psi$.  Moreover, under the assumption that $\psi$ holds,
$\phi$ depends counterfactually on $\vec{X} = \vec{x}$.  In the closest
world to the actual world where $\vec{X} = \vec{x} \land \psi$ holds,
$\phi$ must hold, while in the closest world to the actual world where
$\vec{X} \ne \vec{x} \land \psi$ holds, $\neg \phi$ must hold.
%joe26: jp0 change
%Yablo tries to put ``naturalness'' conditions on what can be held fixed.
%With these conditions, a more refined notion of causation is that
To counteract such difficulties, Yablo imposes a
requirement of ``naturalness'' on what can be held fixed.  With
these requirement, a more refined notion of causation is that
$\vec{X} = \vec{x}$ is a cause of $\phi$ if
there is some $\psi$ true in the actual world that can be held fixed so
as to make $\phi$ counterfactually depend on $\vec{X} = \vec{x}$, and
no other ``more natural'' $\psi'$ can be
%joe+
%found such that make the dependence ``artificial''.  While Yablo does
found that makes the dependence ``artificial''.  While Yablo does
give some objective criteria for naturalness, much of the judgment is
subjective, and it is not clear how to model it formally.
%joe26: jp0 change
In other words, it is not clear what relationships
among variables and events must be encoded in the
model in order to formally decide whether one event is
``more natural'' than another, or whether no other
``more natural'' event can be contrived.
The analogous decisions in our formulation are
managed by condition AC2(b), which distinguishes unambiguously
between admissible and inadmissible contingencies.
%joe**
In addition, it restricts the form of contingencies;
only contingencies of the form $\vec{W} = \vec{w}$ are allowed, and not,
for example, contingencies such as $X=Y$.

%joe19: end{commentout}

%joe19: cut for this paper

\section{Discussion\label{sec:discussion}}

%joe4: incorporated your conclusion
We have presented a formal representation of
causal knowledge and a principled way of
determining actual causes
%joe19
%and explanations
from such knowledge.
We have shown that the counterfactual
approach to causation, in the tradition of Hume and Lewis,
need not be abandoned; the language of
counterfactuals, once supported with structural semantics,
can yield a plausible and elegant account of actual causation that
resolves major difficulties in the traditional account.
%joe10: added; truth in advertising
%joe25: no more discussion of problematic aspects. Should we add some?
%While,
%%joe18
%as we have seen,
%our account also has its problematic aspects, we are optimistic
%that they can be resolved along the lines we have sketched.

The essential principles of our account include
\begin{itemize}
%joe1y
%\item the use of structural equation semantics of counterfactuals;
\item
%joe16
%the use of structural equation semantics of counterfactuals
using structural equations to model causal mechanisms and counterfactuals;
\item
%joe16
%the use of
using
uniform counterfactual notation to encode
and distinguish facts, actions, outcomes, processes, and
contingencies;
%joe4
%\item Deployment of structural contingencies
%to uncovering genuine causal dependencies.
\item using structural contingencies
%joe26: jp0 change.  It's the first time we say ``latent''.  Should we
%define it, or perhaps use the word ``hidden''?
%to uncover causal dependencies;
to uncover latent counterfactual dependencies;
\item careful screening of these contingencies
to avoid tampering with the causal processes to be uncovered.
%jp7 adding two more bullets
%jp8 fixing in light of joe7
%joe19: cut for this paper
%\item treating a simple explanation as a proposition that conveys
%new knowledge and that, once believed, would constitute
%a cause (of the explanandum)
%joe7: I didn't really understand this bullet
%joe8: actually, it's the following bullet I didn't understand
%jp8 remove. it is already subsumed by previous bullet
%\item focusing the knowledge conveyed by each explanatory
%sentence on the support of a specific causal claim.
\end{itemize}

%joe4 added
%joe26: jp0 change
%Our approach also stresses the importance of careful modeling.
%%joe10: added; this may be controversial.  Let's discuss and perhaps rewrite
%In particular, it shows that the choice of random variables can have a
%significant effect on the causality relation.  This perhaps can be
%viewed as as significant deficiency in the approach.  We prefer to think
%that it shows that, in the end, causality is a human construct.
%Thinking of a situation in terms of causality can be very useful and is
%something that humans often do.  As a result, it is perhaps not
%surprising that we are good at choosing the appropriate random variables
%for doing so.
Our approach also stresses the importance of careful modeling.
In particular, it shows that the choice of model granularity
can have a significant effect on the causality relation.
This perhaps can be
%joe+
%viewed as as deficiency in the approach.  We prefer to think
viewed as a deficiency in the approach.  We prefer to think
that it shows that the internal structures of the processes
assumed to underlie causal stories play a crucial role in our judgment of
actual causation, and that it is important therefore to properly
cast such stories in a language that represents those
structures explicitly. Our approach is built on just such a language.

As the examples have shown, much depends on choosing the ``right'' set
of variables with which to model a situation, which ones to make
exogenous, and which to make endogenous.  While the examples have
suggested some heuristics for making appropriate choices,
we do not have a general theory for how to
make these choices.  We view this as an important direction for future
research.  
%joe**
\newcomment
(See \cite{Hitchcock03} for some preliminary discussion of the issue of
finding ``good'' models.)

%joe**
While we do feel that it should be possible to delineate good guidelines
\newcomment
for constructing appropriate models,  ultimately, the choice of model is
a subjective one.  The choice of which variables to focus on and which
to ignore (that is, the choice of exogenous and endogenous variables)
and the decision as to which contingencies to take seriously (that is,
which settings to take as allowable) is subjective, and depends to some
extent on what the model is being used for.  (This issue arises
frequently in discussions of causality and the law \cite{HH85}.)  
By way of contrast, most of the work in the
philosophy literature seems to implicitly assume that, in any given
situation, there is one correct answer as to whether $A$ is a cause of
$B$.   Rather than starting with a model, there are assumed to be events
in the world; new events can be created to some extent as needed,
leading to issues like ``fragility'' of events and how fine-grained
events should be (see, for example, \cite{Lewis00,paul:01}).

%Of course, as we mentioned before, we cannot prove that our definition
of causality 
is ``right''.  However, the fact that it deals so well with the many
difficult examples in the literature does provide some support for the
reasonableness of the definition.  Further support is provided by the
ease with which it can be extended to define other notions, such as
explanation (see Part II of this paper) and responsibility and blame 
\cite{ChocklerH03}.

%joe8:
\appendix
%joe10: one technical issue became the voting example
%\section{Appendix: three technical issues}
%joe11: two became ``some''
%\section{Appendix: two technical issues}
%In this appendix, we consider two technical issues related to the
\section{Appendix: Some Technical Issues}
In this appendix, we consider some technical issues related to the
definition of causality.

\subsection{The active causal process}
We first show that,
without loss of generality, the variables in the set
$\vec{Z}$ in condition AC2 of the definition of causality can all be
%joe**
%taken to be on a path from a path from a variable in $\vec{X}$ to one
taken to be on a path from a variable in $\vec{X}$ to a
variable in $\phi$.  In fact, they can, without loss of generality, be
assumed to change value when $\vec{X}$ is set to $\vec{x}'$ and
$\vec{W}$ is set to $\vec{w}'$.  More formally, consider
the following strengthening of AC2:
\begin{description}
\item[AC2$'$.]
There exists a partition $(\vec{Z},\vec{W})$ of $\V$ with $\vec{X}
\subseteq \vec{Z}$
and some setting $(\vec{x}',\vec{w}')$
of the variables in $(\vec{X},\vec{W})$
such that,
%joe10
%if $(M,\vec{u}) \sat \vec{Z} = \vec{z}^*$ then
if $(M,\vec{u}) \sat Z = z^*$ for $Z \in \vec{Z}$, then
\begin{description}
\item[{\rm (a)}]
$(M,\vec{u}) \sat [\vec{X} \gets \vec{x}',
%joe12: changing according to your preference
%\vec{W} \gets \vec{w}'](\neg \phi \land \land_{Z \in \vec{Z}}(Z \ne z^*))$
\vec{W} \gets \vec{w}'](\neg \phi \land Z \ne z^*)$
for all $Z \in \vec{Z}$;
\item[{\rm (b)}]
$(M,\vec{u}) \sat [\vec{X} \gets
\vec{x}, \vec{W} \gets \vec{w}', \vec{Z}' \gets \vec{z}^*]\phi$ for all
subsets $\vec{Z'}$ of $\vec{Z}$.
%joe10
%(\vec{Z} = \vec{z}^*))$.
\end{description}
\end{description}

As we now show,
we could have replaced AC2 by AC2$'$; it would not have affected
the notion of causality.  Say that $\vec{X}=\vec{x}$ is an {\em actual
cause$'$\/} of $\phi$ if AC1,  AC2$'$, and AC3 hold.

%joe25
%\pro\label{cause'} $\vec{X} = \vec{x}$ is a cause of $\phi$ iff
%$\vec{X}=\vec{x}$ is a cause$'$ of $\phi$.
\pro\label{cause'} $\vec{X} = \vec{x}$ is an actual cause of $\phi$ iff
$\vec{X}=\vec{x}$ is an actual cause$'$ of $\phi$.
\epro
%joe20
%\prf Define $\vec{X} = \vec{x}$ to be a {\em weak cause\/} (\respc {\em
%weak cause$'$\/}) of $\phi$ if AC1 and AC2 (\respc AC1 and AC2$'$) hold,
\prf
%joe25: proof rewritten a bit; it wasn't quite right before.
The ``if'' direction is immediate, since AC2$'$ clearly implies AC2.
For the ``only if''
direction, suppose that $\vec{X} = \vec{x}$ is a cause of
$\phi$.  Let $(\vec{Z},\vec{W})$ be the partition of $\V$ and
$(\vec{x}',\vec{w}')$ the setting of the variables in
$(\vec{X},\vec{W})$ guaranteed to exist by AC2.  Let $\vec{Z}' \subseteq
\vec{Z}$ consist of
%joe25: can't assume this; it may not be true (that's why a rewrote the
%proof, to take this case into account)
%the variables $\vec{X}$ together with the
variables $Z \in \vec{Z}$ such that
$(M,\vec{u}) \sat [\vec{X} \gets \vec{x}',
\vec{W} \gets \vec{w}'](Z \ne z^*)$.  Let $\vec{W}' = \V - \vec{Z}'$.
Notice that $\vec{W}'$ is a superset of $\vec{W}$.
%joe25
Moreover, a priori, $\vec{W}'$ may contain some variables in $\vec{X}$,
although we shall show that this is not the case.
Let
$\vec{w}''$ be a setting of the variables in $\vec{W}$ that agrees with
$\vec{w}'$ on the variables in $\vec{W}$ and for $Z \in \vec{Z} \inter
\vec{W}'$, sets $Z$ to $z^*$ (its original value).  Note that if there
is a variable $V \in \vec{X} \inter \vec{W}'$, then the setting of $V$ is
the same in $\vec{x}'$, $\vec{x}$, and $\vec{w}''$.
Thus, even if $\vec{X}$ and $\vec{W}'$ have a
nonempty intersection, the models $M_{\vec{X} \gets \vec{x}', \vec{W}
\gets \vec{w}'}$ and $M_{\vec{X} \gets \vec{x}', \vec{W}' \gets
\vec{w}''}$ are well defined.  Since $Z = z^*$ in
the unique solution to the equations in $M_{\vec{X} \gets \vec{x}', \vec{W}
\gets \vec{w}'}$  and the equations
in $M_{\vec{X} \gets \vec{x}, \vec{W} \gets \vec{w}'}$, it follows that
(a) the equations in
$M_{\vec{X} \gets \vec{x}', \vec{W}' \gets \vec{w}''}$ and
$M_{\vec{X} \gets \vec{x}', \vec{W} \gets \vec{w}'}$ have the same
solutions and
(b) the equations in
$M_{\vec{X} \gets \vec{x}, \vec{W}' \gets \vec{w}''}$ and
$M_{\vec{X} \gets \vec{x}, \vec{W} \gets \vec{w}'}$ have the same
solutions.
Thus,
$(M,\vec{u}) \sat [\vec{X} \gets \vec{x}', \vec{W}' \gets
\vec{w}''](\neg \phi \land (Z \ne z^*))$ for all $Z \in \vec{Z}'$
and $(M,\vec{u}) \sat [\vec{X} \gets
\vec{x}, \vec{W} \gets \vec{w}'](\phi \land (Z = z^*))$ for all $Z \in
\vec{Z}'$.  That is, AC2$'$ (and hence AC2) holds for the pair
$(\vec{Z}',\vec{W}')$.  It follows that $\vec{W}' \inter \vec{X} =
\emptyset$, for otherwise $\vec{X} = \vec{x}$ is not a cause of $\phi$:
it violates AC3.   Thus, $\vec{Z}' \supseteq \vec{X}$, and
$\vec{X} = \vec{x}$ is a cause$'$ of $\phi$, as desired.
\eprf

Proposition~\ref{cause'} shows that, without loss of generality, the
variables in
$\vec{Z}$ can be taken to be ``active'' in the causal
process, in that they change value when the variables in
$\vec{X}$ do.  This means that each variable in $\vec{Z}$ must be a
descendant of some variable in $\vec{X}$ in the causal graph.
The next result shows that, without loss of generality, we can also
assume that the variables in $\vec{Z}$ are on a path from a variable in
$\vec{X}$ to a variable that appears in $\phi$.  Recall that we defined
an active causal process to consist of a minimal set $\vec{Z}$ that
satisfies AC2.

\pro\label{cause''}  All the variables in an active causal process
corresponding to a cause $\vec{X} = \vec{x}$ for $\phi$ in $(M,\vec{u})$
must be on a path from some variable in $\vec{X}$ to a variable in
$\phi$ in the causal network corresponding to $M$.
\epro
\prf Suppose that $\vec{Z}$ is an active causal process,
$(\vec{Z},\vec{W})$ is the partition satisfying AC2 using the setting
$(\vec{x}',\vec{w}')$.
By Proposition~\ref{cause'}, all the variables in $\vec{Z}$
must be descendants of a variable in $\vec{X}$.  Suppose that some
variable $Z \in \vec{Z}$ is not on a path from a variable in $\vec{X}$
to a variable in $\phi$.  That means there is no path from $Z$ to a
variable in $\phi$.  It follows that there is no path from $Z$ to
a variable $Z' \in \vec{Z}$
that is on a path from a variable in $\vec{X}$ to a variable in $\phi$.
%joe25
%That means that changing the value of $Z$
Thus, changing the value of $Z$
cannot affect the value of $\phi$ nor of any variable $Z' \in
\vec{Z}$.  Let $\vec{Z}' = \vec{Z} - \{Z\}$ and $\vec{W}' = \vec{W}
\union \{Z\}$.  Extend $\vec{w}'$ to $\vec{w}''$ by assigning $Z$ to its
original value $z^*$ in context $(M,\vec{u})$.  It is now immediate from
the preceding observations that $(\vec{Z}',\vec{W}')$ is a partition
satisfying AC2 using the setting $(\vec{x}',\vec{w}'')$.  This
contradicts the minimality of $\vec{Z}$. \eprf

%joe11: added the voting example here (moved it from the problematic examples
%section) and rewrote it somewhat
\subsection{A closer look at AC2(b)}\label{app:AC2b}
%joe19: rewrote slightly
Clause AC2(b) in the definition of causality
is complicated by the need to check that $\phi$ remains true if
%joe**
\newcomment
%$\vec{X}$ is set to $\vec{x}$, $\vec{W}$ is set to $\vec{w}'$, and
$\vec{X}$ is set to $\vec{x}$, any subset of the variables in $\vec{W}$
is set to $\vec{w}'$, and 
%joe25: slight rewording
%any subset $\vec{Z}'$ of the variables in $\vec{Z}$
all the variables in an arbitrary subset $\vec{Z}'$ of $\vec{Z}$
are set to their
original values $\vec{z}^*$ (that is, the values they had in the
original context, where $\vec{X} = \vec{x}$ and $\vec{W} = \vec{w}$).
This check would be simplified considerably if, for each variable $z\in
\vec{Z}$ 
%joe**
\newcomment
%we have $Z = z^*$ when $\vec{X} = \vec{x}$ and $\vec{W} = 
and each subset $\vec{W}'$ of $\vec{W}$, 
we have that $Z = z^*$ when $\vec{X} = \vec{x}$ and $\vec{W}' = 
%joe+
%\vec{w}'$; that is, if we require that in AC2(b) that $(M,u) \sat
\vec{w}'$; that is, if we require in AC2(b) that $(M,u) \sat
%joe**
%[\vec{X} \gets \vec{x}, \vec{W} \gets w']Z = z^*$ for all variables $Z
%joe+
%[\vec{X} \gets \vec{x}, \vec{W}' \gets w']Z = z^*$ for all variables $Z
[\vec{X} \gets \vec{x}, \vec{W}' \gets w'](Z = z^*)$ for all variables $Z
\in \vec{Z}$
%joe**
and all subsets $\vec{W}'$ of $\vec{W}$.  
(Note that this requirement would imply the current requirement.)
This stronger requirement holds in all the examples we have considered so far.
However, the following example shows that it does not hold in general.

\xam\label{voting}  Imagine that a vote takes place.  For simplicity,
two people vote.  The measure is passed if at least one of them votes in
favor.  In fact, both of them vote in favor, and the measure passes.
This version of the story is almost identical to
%joe25
the disjunctive scenario in
Example~\ref{xam:arson}.  If we use $V_1$ and $V_2$ to denote how the
voters vote ($V_i = 0$ if voter $i$ votes against and $V_i = 1$ if she
votes in favor) and $P$ to denote whether the measure passes ($P=1$ if
it passes, $P=0$ if it doesn't), then in the context where $V_1 = V_2 =
1$, it is easy to see that each of $V_1 = 1$ and $V_2 = 1$ is a cause of
$P=1$.  However, suppose we now assume that there is a voting machine
that tabulates the votes.  Let $M$ represent the total number of
votes recorded by the machine.  Clearly $M = V_1 + V_2$ and $P=1$ iff $M
\ge 1$.  The following causal network represents this more refined
version of the story.
\begin{figure}[htb]
\input{psfig}
%\hspace*{.75in}
\centerline{\includegraphics{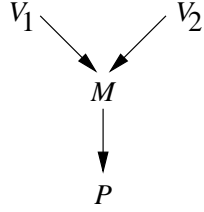}}
\caption{An example showing the need for AC2(b).}
\label{fig-app}
\end{figure}
In this more refined scenario,
%$V_1=0$ and $V_2=0$ are still both causes of $P=1$.
%jp13 I believe it should be:
$V_1=1$ and $V_2=1$ are still both causes of $P=1$.
%Consider $V_1=1$.  Take $\vec{Z} =\{V_1, M, P\}$ and $\vec{W} = {V_2}$.
%Much like the simpler version of the story, if $V_1 = V_2 = 0$, then $P
%= 0$, satisfying AC2(a).  Changing $V_1$ to 1 (its original value) and
%keeping $V_2 = 0$ results in $P=1$, but $M$ is 1, not 2 as it is if
%$V_1 = V_2 = 1$.  Nevertheless, $P=1$ regardless of whether $M=1$ or
%$M=2$, so AC2(b) is satisfied and $V_1 = 1$ is a cause of $P=1$.
%However, if we had insisted in AC2(b) that $(M,u) \sat [\vec{X} \gets
%\vec{x}, \vec{W} \gets w']Z = z^*$ for all variables $Z \in \vec{Z}$
%(which in this case means that $M$ would have to retain its original
%value of 2 when $V_1=1$ and $V_2=0$), then
%neither $V_1 = 1$ nor $V_2 = 1$ would be a cause of $P=1$ (although with
%the obvious modification of the definition, $V_1 = 1 \land V_2 = 1$
%would be a cause of $P=1$.)
%\exam
%jp13. Joe. I will rewrite the explanation in accordance
%with our original motivation of AC2, feel free to take and
%use whatever you like from it.
Consider $V_1=1$.  Take $\vec{Z} =\{V_1, M, P\}$ and $\vec{W} = {V_2}$.
Much like the simpler version of the story, if we choose the
%joe12: I basically like it; I made a few minor changes
%contingency w': $V_2 = 0$, then $P$ is counterfactually
contingency $V_2 = 0$, then $P$ is counterfactually
dependent on $V_1$, so AC2(a) holds.  To check
%joe25
%if this contingency satisfies AC2(b), we set $V_1$ to 1 (their original
%value) and check that setting $V_2$ to 0 does not change the value of
%$P$. This is indeed the case. Although $M$ becomes 1, not 2 as it is
%when $V_1 = V_2 = 1$, nevertheless, $P=1$ continues to hold,
%so AC2(b) is satisfied and $V_1 = 1$ is a cause of $P=1$.
that this contingency satisfies AC2(b), note that setting $V_1$ to 1 and
$V_2$ to 0 results in $P = 1$, even if we also set $M$ to 2 (its
current value).
However, if we had insisted in AC2(b) that $(M,u) \sat [\vec{X} \gets
%joe+
%\vec{x}, \vec{W} \gets w']Z = z^*$ for all variables $Z \in \vec{Z}$
\vec{x}, \vec{W} \gets w'](Z = z^*)$ for all variables $Z \in \vec{Z}$
(which in this case means that $M$ would have to retain its original
value of 2 when $V_1=1$ and $V_2=0$), then
neither $V_1 = 1$ nor $V_2 = 1$ would be a cause of $P=1$ (although
%joe25: no need to further modify the definition
%with the obvious modification of the definition,
$V_1 = 1 \land V_2 = 1$
would be a cause of $P=1$).
Since, in general, one can always imagine that a change in
one variable produces some feeble change in another,
we cannot insist on
%joe12
%$Z$ remaining absolutely constant;
%instead, we require merely that changes in $Z$ should not affect
the variables in $\vec{Z}$ remaining constant;
instead, we require merely that changes in $\vec{Z}$ not affect
$\phi$.
\exam

%joe18: added; should we move this example earlier?
We remark that this example is not handled correctly by Pearl's causal
beam definition.  According to the causal beam definition, there is no
cause for $P=1$!
%joe19
%More generally, the causal beam definition handles
%only causality between primitive events.
%joe+
%It can be shown if $X=x$
It can be shown that if $X=x$
is an actual (or contributory) cause of $Y=y$ according to the causal
beam definition given in \cite{pearl:2k}, then it is an actual cause
according to the definition here.  As Example~\ref{voting} shows, the
converse is not necessarily true.

%joe**: new material
\newcomment
Another complicating factor in AC2(b) is that the requirement must hold
for all subsets $\vec{W}'$ of $\vec{W}$.  In a preliminary version of
%joe+
%this paper, we required only that AC2(b) hold for $\vec{W}$.  That is,
this paper \cite{HPearl01a}, we required only that AC2(b) hold for $\vec{W}$.  That is,
the condition we had was 
\begin{description}
\item[AC2(b$'$).] $(M,\vec{u}) \sat [\vec{X} \gets
\vec{x}, \vec{W} \gets \vec{w}', \vec{Z}' \gets \vec{z}^*]\phi$ for 
all subsets $\vec{Z'}$ of $\vec{Z}$.   
\end{description}
However, as Hopkins and Pearl
\citeyear{HopkinsP02}  pointed out, AC2(b$'$) is too
permissive. To use their example, suppose that a prisoner dies 
either if $A$ loads $B$'s gun and $B$ shoots, or if $C$ loads and shoots
his gun.  Taking $D$ to represent the prisoner's death and making the
obvious assumptions about the meaning of the variables, we have that
$D=1$ iff $(A=1 \land B=1) \lor (C=1)$.  Suppose that in the actual
context $u$, $A$ loads $B$'s gun, $B$ does not shoot, but $C$ does load
and shoot his gun, so that the prisoner dies.  Clearly $C=1$ is a cause
of $D=1$.  We would not want to say that $A=1$ is a cause of $D=1$,
given that $B$ did not shoot (i.e., given that $B=0$).  However,
with AC2(b$'$), $A=1$ is a cause of $D=1$.  For we can take
$\vec{W} = \{B,C\}$ and consider the contingency where $B=1$ and $C=0$.
It is easy to check that AC2(a) and AC2(b$'$) hold for this contingency,
so under the old definition, $A=1$ was a cause of $D=1$.  However,
AC2(b) fails in this case, for $(M,u) \sat [A \gets 1, C \gets 0]D=0$.

%joe19*: new subsection
\subsection{Causality with infinitely many variables}\label{s:infinite}
Throughout this paper, we have assumed that $\V$, the set of exogenous
variables, is finite.  Our definition (in particular, the minimality
clause AC3) has to be modified if we drop this assumption.  To see why,
consider the following example:

\xam\label{xam:infinite}
Suppose that $\V = \{X_0, X_1, X_2, \ldots, Y\}$.  Further assume that the
structural equations are such that $Y = 1$ iff infinitely many of the
$X_i$'s are 1; otherwise $Y = 0$.  Suppose that in the actual context,
all of the $X_i$'s are 1 and, of course, so is $Y$.  What is the cause
of $Y=1$?

According to our current definitions, it is actually not hard to check
that there is no event which is the cause of $Y=1$.  For suppose that
$\land_{i \in I} X_i = 1$ is a cause of $Y=1$, for some subset $I$ of
the natural numbers.  If $I$ is finite, then to satisfy
AC2(a), we must take $\vec{W}$ to be a cofinite subset of the $X_i$'s
(that is, $\vec{W}$ must include all but finitely many of the $X_i$'s).
But then if we set all but finitely many of the $X_i$'s in $\vec{W}$ to
0 (as we must to satisfy AC2(a) if $I$ is finite), AC2(b) fails.  On the
other hand, if $I$ is infinite and there exists a partition
$(\vec{Z},\vec{W})$ such that AC2(a) and (b) hold, then if $I'$ is the
result of removing the smallest element from $I$, it is easy to see that
$\land_{i \in I'} X_i = 1$ also satisfies AC2(a) and (b), so AC3 fails.
\exam

Example~\ref{xam:infinite} shows that the definition of causality must
be modified if $\V$ is infinite.  It seems that the minimality condition
AC3 should be modified. Here is a suggested modification:
\begin{description}
\item[{\rm AC3$'$.}]
If any strict subset $\vec{X}'$ of $\vec{X}$ satisfies
conditions AC1 and AC2, then there is a strict subset $\vec{X}''$ of
$\vec{X}'$ that also satisfies AC1 and AC2.
\end{description}
It is easy to see that AC3 and AC3$'$ agree if $\V$ is finite.
Roughly speaking, AC3$'$ says that if there is a minimal conjunction
that satisfies AC1 and AC2, then it is a cause.  If there is no minimal
one (because there is an infinite descending sequence), then any
conjunction along the sequence qualifies as a cause.

If we use AC3$'$ instead of AC3, then in Example~\ref{xam:infinite},
$\land_{i \in I} X_i= 1$ is a cause of $Y = 1$ as long as $I$ is
infinite.  Note that it is no longer the case that we can restrict to a
single conjunct if $\V$ is infinite.

We do not have sufficient experience with this definition to be
confident that it is indeed just what we want, but it seems like a
reasonable choice.

\subsection{Causality in nonrecursive models}
We conclude by considering how the definition of causality
%joe19
%(and hence explanation)
can be modified to deal with nonrecursive models.
In nonrecursive models, there may be more than one solution to an
equation in a given context, or there may be none.
%joe16: added
In particular, that means that a context no longer necessarily
determines the values of the endogenous variables.  Earlier, we
identified a primitive event such as $X=s$ with the basic causal formula
$[\, ](X=x)$, that is, with the special case of a formula of the form
$[Y_1 \gets y_1, \ldots, Y_k \gets y_k]\phi$ with $k=0$.
%joe25: expanded
%Now we no longer want to do this,
$(M,\vec{u}) \sat [\, ](X=x)$ if $X=x$ in all solutions to the equations
where $\vec{U} = \vec{u}$.  It seems reasonable to identify $[\, ](X=x)$
with $X=x$ if there is a unique solution to these equations.  But it is
not so reasonable if there may be several solutions, or no solution.
What we really want to do is to be able to say that
$X=x$ under a particular setting of the variables.
Thus, we now take the truth of a primitive event such as
%joe+
%such as $X = x$ relative not just to a context,
$X = x$ relative not just to a context,
but to a complete description $(\vec{u},\vec{v})$ of the values of both
the exogenous and the endogenous variables.  That is,
$(M,\vec{u},\vec{v}) \sat X=x$ if $X$ has value $x$ in $\vec{v}$.  Since
the truth of $X=x$ depends on just $\vec{v}$, not $\vec{u}$, we
sometimes write $(M,\vec{v}) \sat X=x$.  We extend this definition to
Boolean combinations of primitive events in the standard way.
%joe16
%We then define $(M,\vec{u},\vec{v}) \sat [\vec{Y} \gets \vec{y}](X=x)$
%if in all solutions to the equations in
%$M_{\vec{Y} \gets \vec{y}}$ in context $\vec{u}$, the variable $X$ has
%value $x$.
We then define $(M,\vec{u},\vec{v}) \sat [\vec{Y} \gets \vec{y}]\phi$
if $(M,\vec{v}') \sat \phi$ for all solutions $(\vec{u},\vec{v}')$ to
the equations in $M_{\vec{Y} \gets \vec{y}}$.
Since the truth of $[\vec{Y} \gets \vec{y}](X=x)$ depends
only on the context $\vec{u}$ and not on $\vec{v}$, we typically
write $(M,\vec{u}) \sat [\vec{Y} \gets \vec{y}](X=x)$.

The formula $\<\vec{Y} \gets \vec{y}\>(X=x)$ is the dual of
$[\vec{Y} \gets \vec{y}](X=x)$; that is, it is an abbreviation of
$\neg [\vec{Y} \gets \vec{y}](X \ne x)$.  It is easy to check that
$(M,\vec{u},\vec{v}) \sat \<\vec{Y} \gets \vec{y}\>(X=x)$ if in some
solution to the equations in
$M_{\vec{Y} \gets \vec{y}}$ in context $\vec{u}$, the variable $X$ has
value $x$.  For recursive models, it is immediate that
$[\vec{Y} \gets \vec{y}](X = x)$ is equivalent to
$\<\vec{Y} \gets \vec{y}\>(X = x)$, since all equations have exactly one
solution.

With these definitions in hand, it is easy to state our definition of
causality for arbitrary models.  Note it is now taken with respect to a
tuple $(M,\vec{u},\vec{v})$, since we need the values of the exogenous
variables to define the actual world.

%jp13 Joe, I see that you left the old AC2 in this definition
% is it on purpose?
%joe12: no; it was a mistake.  I corrected it.
\dfn
$\vec{X} = \vec{x}$ is an {\em actual cause of $\phi$ in
$(M, \vec{u},\vec{v})$\/} if the following
%joe+
%three conditions hold;
three conditions hold.
\begin{description}
\item[{\rm AC1.}] $(M,\vec{v}) \sat (\vec{X} =
\vec{x}) \land \phi$.
\item[{\rm AC2.}]
There exists a partition $(\vec{Z},\vec{W})$ of $\V$ with $\vec{X}
\subseteq \vec{Z}$ and some
setting $(\vec{x}',\vec{w}')$ of the variables in $(\vec{X},\vec{W})$
such that
if $(M,\vec{u},\vec{v}) \sat \vec{Z} = \vec{z}^*$, then
\begin{description}
\item[{\rm (a)}]
$(M,\vec{u}) \sat \<\vec{X} \gets \vec{x}',
\vec{W} \gets \vec{w}'\>\neg \phi$.
\item[{\rm (b)}]
$(M,\vec{u}) \sat [\vec{X} \gets
%jp11 correcting w into w', to read
%\vec{x}, \vec{W} \gets \vec{w}'](\phi \land
%(\vec{Z} = \vec{z}^*))$.
%joe12: correction
\vec{x}, \vec{W} \gets \vec{w}', \vec{Z}' \gets \vec{z}^*]\phi$ for all
subsets $\vec{Z'}$ of $\vec{Z}$.
(Note that in part (a) we require that the value of $\phi$ change only
in some solution to the equations, while in (b), we require that it stay
true in {\em all\/} solutions.)
\end{description}
\item[{\rm AC3.}]
$\vec{X}$ is minimal; no subset of $\vec{X}$ satisfies
conditions AC1 and AC2. \eprf
\end{description}
\end{definition}
%joe25:
%There is a similar generalization of the definition of explanation.

%joe25: corrected typos
%While these seem like the most natural generalizations
%of our definition of causality and explanation to deal with nonrecursive
%models, we have not examined examples to verify that these definitions
%joe29
%While this seem like the most natural generalization
While this seems like the most natural generalization
of the definition of causality  to deal with nonrecursive
models, we  have not examined examples to verify that this definition
gives the expected result, partly because all the standard
examples are most naturally modeled using recursive models.

\subsection*{Acknowledgments} We thank Christopher Hitchcock for many
useful 
%joe**
%joe+
%discussion,
discussions,
for his numerous comments on earlier versions of the paper,
%joe13: added next line
and for pointing out Example~\ref{xam3};
%joe**: added Mark, James, and Spohn
%joe+
%Mark Hopkins, James Park, Wolfgang Spohn, Zoltan Szabo for
Mark Hopkins, James Park, Wolfgang Spohn, and Zoltan Szabo for
stimulating discussions; and Eric Hiddleston for pointing out
Example~\ref{Larry}.

\bibliographystyle{chicago}
\bibliography{z,joe,refs,expl} %%,book}
\end{document}